\documentclass[twocolumn]{svjour3}

\usepackage[latin1]{inputenc}
\usepackage{times}

\usepackage{psfrag, amsmath, amssymb, amscd, amsfonts, latexsym, epsf, graphicx}
\usepackage{natbib}
\usepackage{hyperref}

\usepackage[switch]{lineno}

\usepackage[labelformat=simple]{subcaption}

\def\bbbr{{\mathbb R}} 
\def\bbbz{{\mathbb Z}}

\newcommand{\argmin}{\operatorname{argmin}}
\newcommand{\orth}{\bot}
\newcommand{\diag}{\operatorname{diag}}

\def\norm{\scriptsize\mbox{norm}}

\def\oned{\scriptsize\mbox{1D}}
\def\dc{\scriptsize\mbox{DC}}
\def\ideal{\scriptsize\mbox{ideal}}

\journalname{arXiv preprint}

\begin{document}

\title{\bf Modelling and analysis of the 8 filters from the "master key filters hypothesis" for depthwise-separable deep networks in relation to idealized receptive fields based on scale-space theory%
\thanks{The support from the Swedish Research Council 
              (contract 2022-02969) is gratefully acknowledged. }}

\titlerunning{Modelling of the 8 filters from "master key filters
  hypothesis" in relation to idealized receptive fields based on
  scale-space theory}
\authorrunning{Tony Lindeberg, Zahra Babaiee and Peyman M. Kiasari}
          
\author{Tony Lindeberg$^1$, Zahra Babaiee$^2$ and Peyman M. Kiasari$^2$}

\institute{$^1$KTH Royal Institute of Technology,
  Computational Brain Science Lab,
  Department of Computational Science and Technology,
  SE-100 44 Stockholm, Sweden.
  $^2$TU Wien, Informatics, AT-1040 Vienna, Austria.}

\date{Received: date / Accepted: date}

\maketitle

\begin{abstract}
  This paper presents the results of analysing and modelling a set of
  8 ``master key filters'', which have been extracted by applying
  a clustering approach to the receptive fields learned in
  depthwise-separable deep networks based on the ConvNeXt
  architecture.
  
  For this purpose, we first compute spatial spread measures in
  terms of weighted mean values and weighted variances of the
  absolute values of the learned filters, which support the working
  hypotheses that: (i)~the learned filters can be modelled by separable
  filtering operations over the spatial domain, and that (ii)~the
  spatial offsets of the those learned filters that are non-centered
  are rather close to half a grid unit. Then, we model the clustered
  ``master key filters'' in terms of difference operators applied to
  a spatial smoothing operation in terms of the discrete analogue of
  the Gaussian kernel, and demonstrate that the resulting idealized
  models of the receptive fields show good qualitative similarity to
  the learned filters.

  This modelling is performed in two
  different ways: (i)~using possibly different values of the scale
  parameters in the coordinate directions for each filter, and (ii)~using
  the same value of the scale parameter in both coordinate
  directions.
  Then, we perform the actual model fitting by either (i)~requiring
  spatial spread measures in terms of spatial variances of the
  absolute values of the receptive fields to be equal, or
  (ii)~minimizing the discrete $l_1$- or $l_2$-norms between
  the idealized receptive field models and the learned filters.

  Complementary experimental results then demonstrate that the idealized
  models of receptive fields have very good predictive properties for
  replacing the learned filters by idealized filters in
  depthwise-separable deep networks, thus showing that the learned
  filters in depthwise-separable deep networks can be well
  approximated by discrete scale-space filters.
  Notably, we show that, for a reduced version
  of the ConvNeXt architecture, using a set of only 8 discrete scale-space
  filters leads to almost as good accuracy as for the receptive fields
  trained from scratch on the ImageNet dataset.
  
\end{abstract}

\section{Introduction}

In computer vision operations, specifically such operations formulated in terms of deep networks applied to image data, a main computational function is performed in terms of the receptive fields, that integrate the image information over non-infinitesimal support regions over the image domain. A both theoretical and practical problem does hence concern how to choose the models to use for such receptive fields in computer vision algorithms.

In the area of scale-space theory, the problem of choosing appropriate
models for receptive fields has been addressed from a normative
theoretical viewpoint, by formulating assumptions, referred to as
scale-space axioms, that reflect desirable properties of a vision
system under selected families of symmetry transformations. By
axiomatic derivations, it has in this respect been shown that Gaussian
kernels and Gaussian derivatives constitute a canonical family of
linear filters to be used in the first layer of the image operations,
see Iijima (\citeyear{Iij62}), Koenderink (\citeyear{Koe84}),
Koenderink and van Doorn (\citeyear{KoeDoo92-PAMI}),
Lindeberg (\citeyear{Lin96-ScSp}, \citeyear{Lin10-JMIV}, \citeyear{Lin21-Heliyon}) and
Weickert {\em et al.\/}\ (\citeyear{WeiIshImi99-JMIV}).
The idealized models for spatial receptive fields obtained in this way have also been shown to agree qualitatively quite well with biological receptive fields registered by neurophysiological recordings in the retina, the lateral geniculate nucleus (LGN) and the primary visual cortex (V1) of higher mammals, see
Lindeberg (\citeyear{Lin13-BICY}, \citeyear{Lin21-Heliyon}).

In the area of deep learning, the problem of choosing the receptive fields is on the other hand addressed from a pure learning perspective, where the resulting receptive fields then arise as the result of optimizing some suitably selected loss function for a specific deep learning architecture. Hybrid approaches to this problem have also been formulated, where the use of Gaussian derivative operators has been extended to higher layers in deep networks by
Jacobsen {\em et al.\/} (\citeyear{JacGemLouSme16-CVPR}),
Lindeberg (\citeyear{Lin21-SSVM}, \citeyear{Lin22-JMIV}),
Pintea {\em et al.\/} (\citeyear{PinTomGoeLooGem21-IP}),
Sangalli {\em et al.\/} (\citeyear{SanBluVelAng22-BMVC}),
Penaud-Polge {\em et al.\/} (\citeyear{PenVelAng22-ICIP}),
Smets {\em et al.\/}  (\citeyear{SmePorBekDui23-JMIV}),
Gavilima-Pilataxi and Ibarra-Fiallo (\citeyear{GavIva23-ICPRS}) and
Perzanowski and Lindeberg (\citeyear{PerLin25-JMIV}).
The use of Gaussian derivative operators in the higher layers of such
hybrid networks has, however, not previously been theoretically motivated by
necessity, as the use of Gaussian derivatives in the first layer of a
vision system has been formally established, based on the
axiomatically derived scale-space theory by
Iijima (\citeyear{Iij62}), Koenderink (\citeyear{Koe84}),
Koenderink and van Doorn (\citeyear{KoeDoo92-PAMI}),
Lindeberg (\citeyear{Lin96-ScSp}, \citeyear{Lin10-JMIV}, \citeyear{Lin21-Heliyon}) and
Weickert {\em et al.\/}\ (\citeyear{WeiIshImi99-JMIV}).

\begin{figure*}[hbtp]
  \begin{center}
    \setlength{\tabcolsep}{-4pt}
    \begin{tabular}{cccccccc}
      {\em Filter~1} & {\em Filter~2} & {\em Filter~3} & {\em Filter~4} &
      {\em Filter~5} & {\em Filter~6} & {\em Filter~7} & {\em Filter~8} \\
      \includegraphics[width=0.13\textwidth]{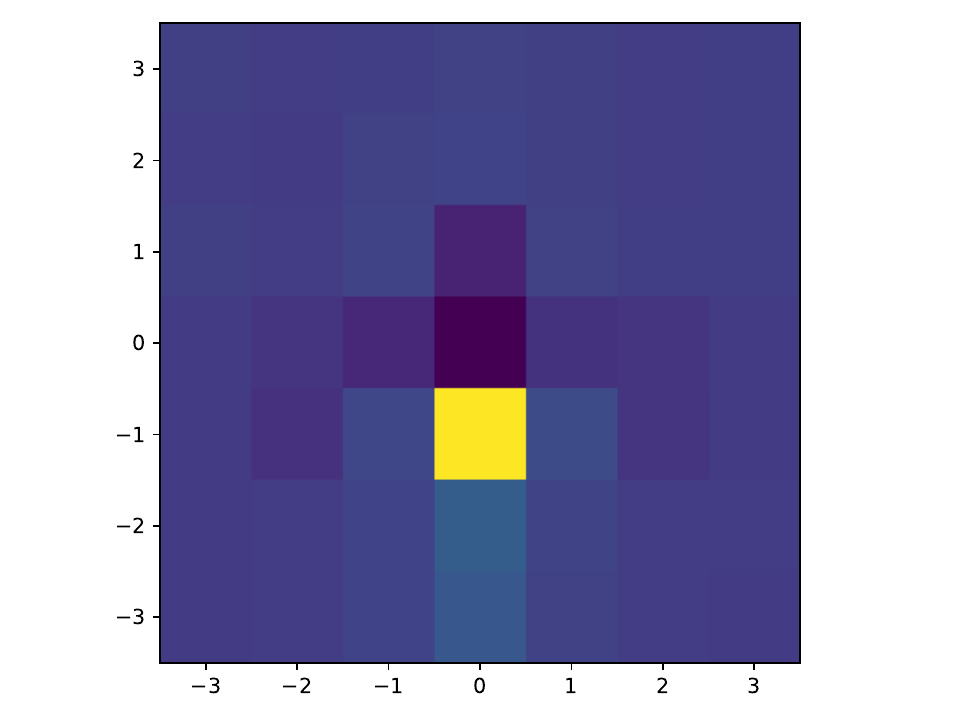}
        & \includegraphics[width=0.13\textwidth]{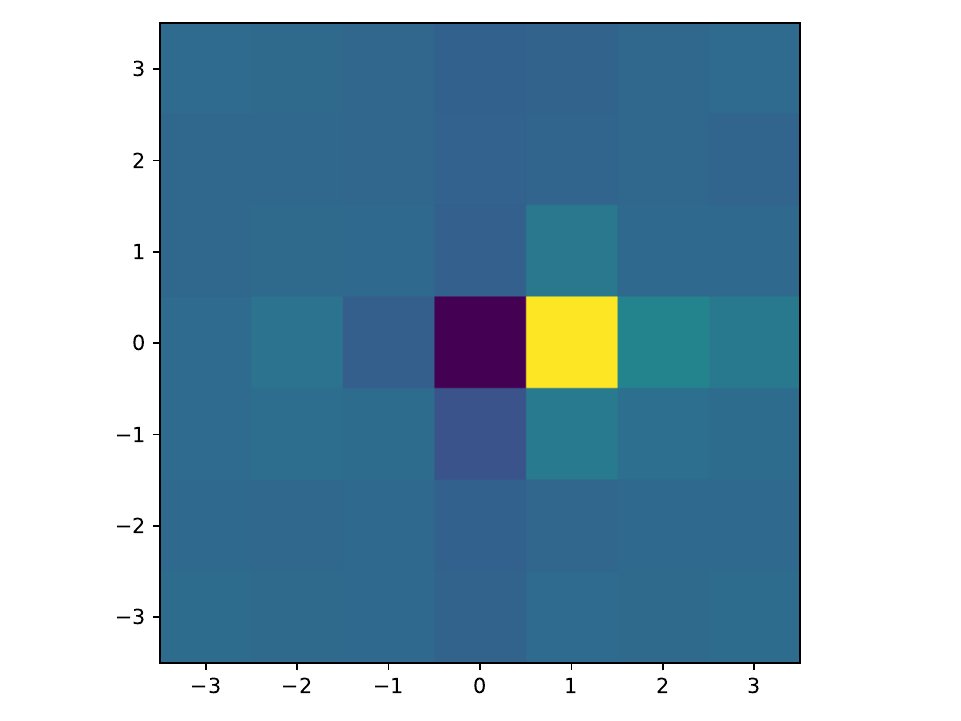}
        & \includegraphics[width=0.13\textwidth]{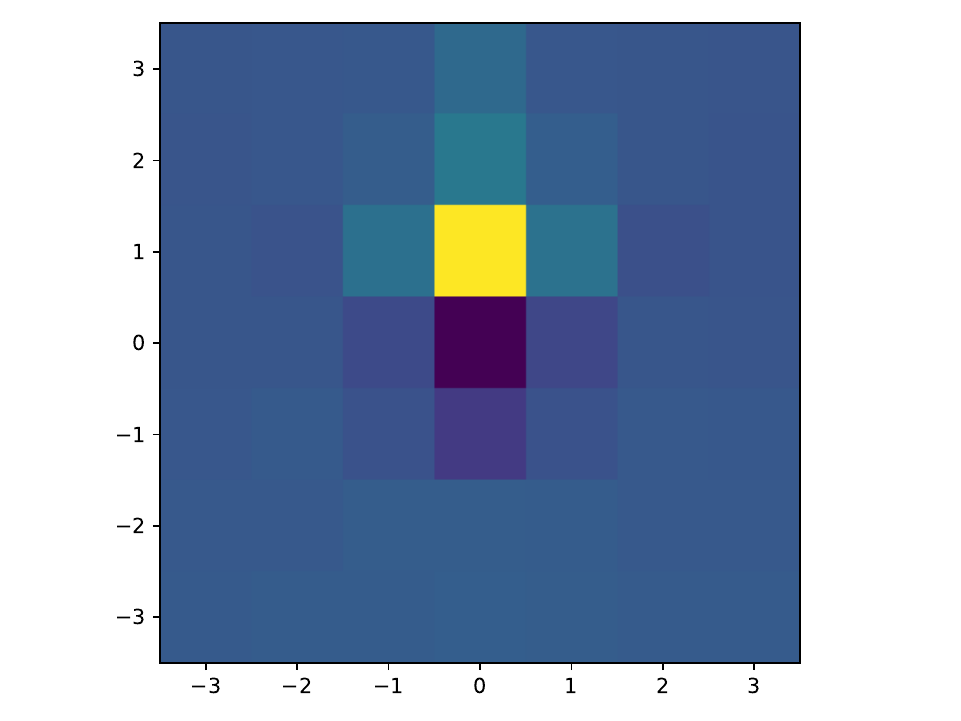}
        & \includegraphics[width=0.13\textwidth]{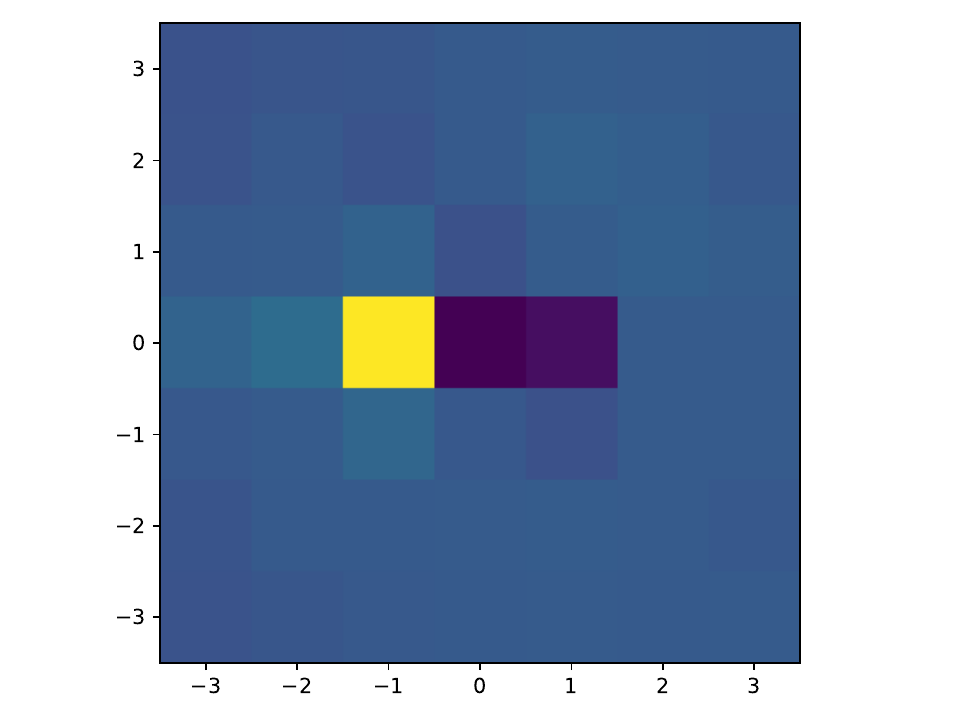} 
        & \includegraphics[width=0.13\textwidth]{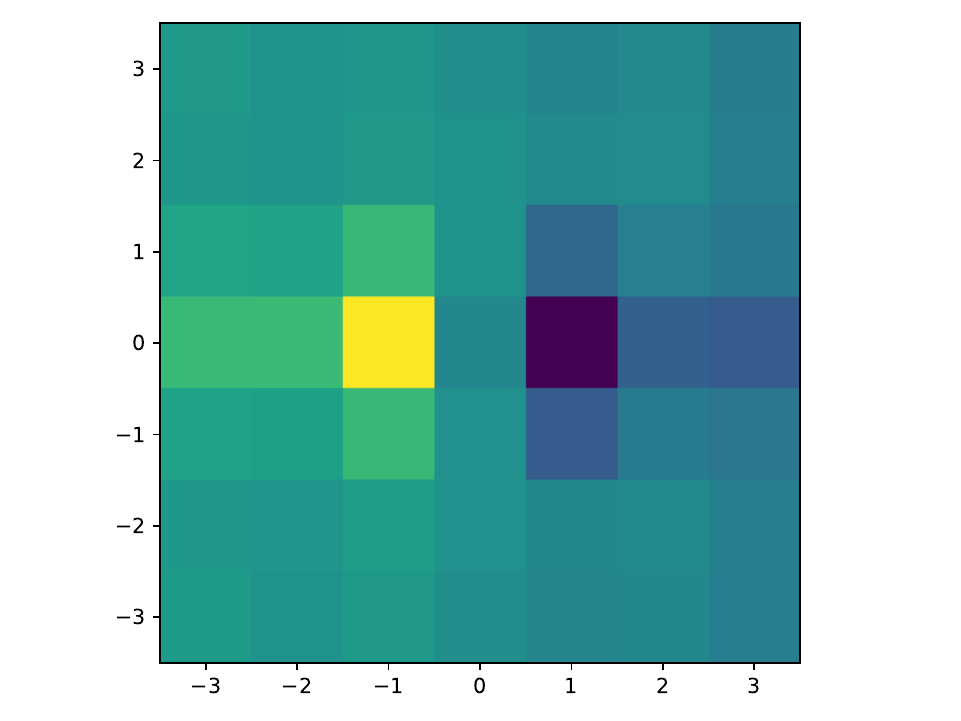}
        & \includegraphics[width=0.13\textwidth]{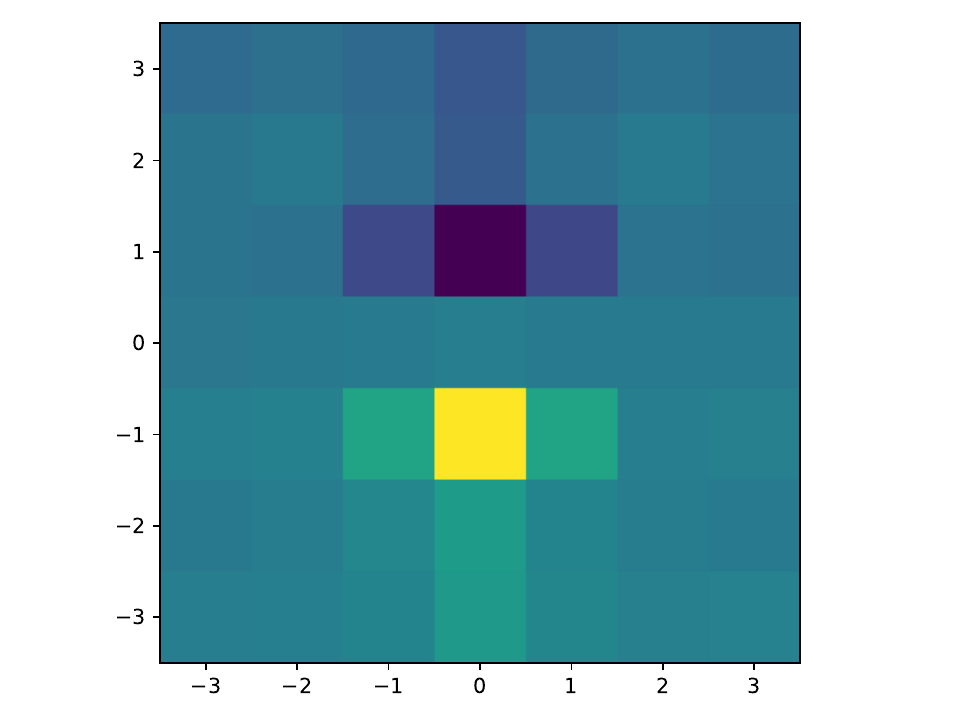}
        & \includegraphics[width=0.13\textwidth]{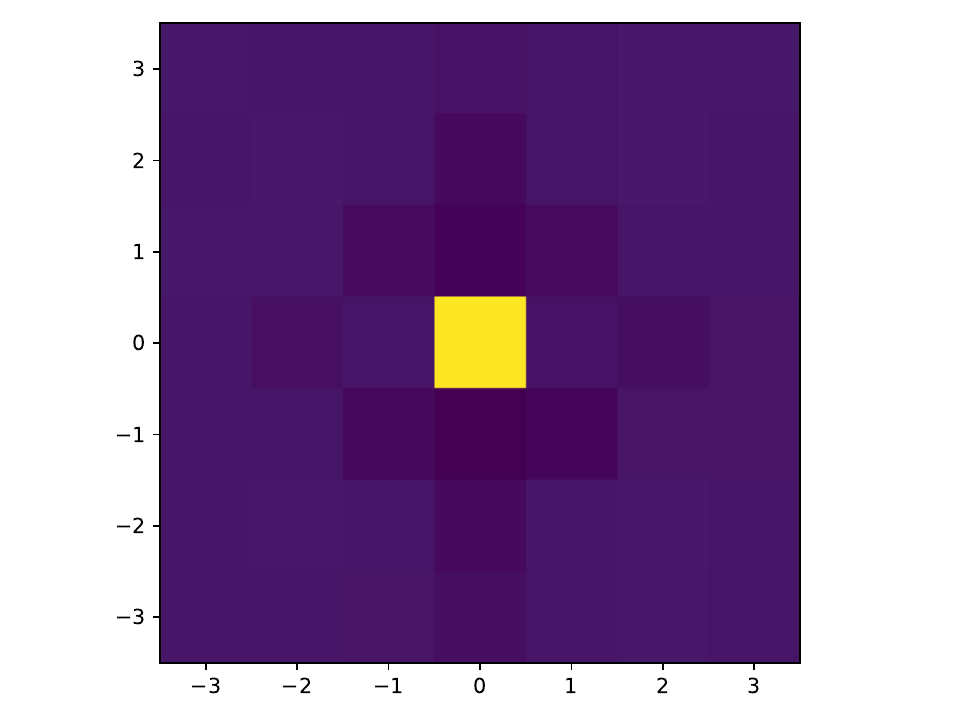}
        & \includegraphics[width=0.13\textwidth]{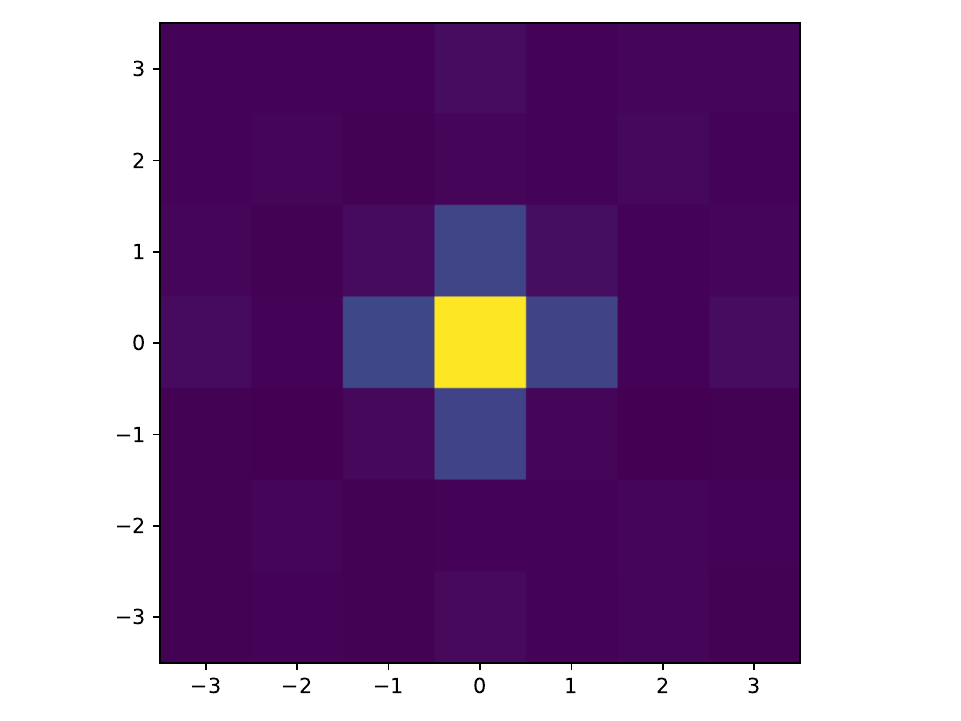}
    \end{tabular}
  \end{center}
  \caption{The original set of 8 ``master key filters'', obtained by applying a clustering technique to
    the receptive fields learned from depthwise-separable deep networks,
    as extracted by Babaiee {\em et al.\/}
    (\citeyear{BabKiaRusGro25-AAAI-master}), ``through greedy search
    on the ConvNeXt V2 Tiny model'' developed by
    Woo {\em et al.\/} (\citeyear{WooDebHuCheLiuKweXi23-CVPR}),
    in turn based on the regular ConvNeXt model developed by
    Liu {\em et al.\/} (\citeyear{LiuMaoWuFeiDarXie22-CVPR}).
   (Horizontal axes: horizontal filter indices $m \in [-3, 3]$.
    Vertical axes: vertical filter indices $n \in [-3, 3]$.)}
  \label{fig-8-extract-filters}
\end{figure*}

\begin{figure*}[hbtp]
  \begin{center}
    \setlength{\tabcolsep}{-4pt}
    \begin{tabular}{ccccccccc}
      &
      {\em Filter~1} & {\em Filter~2} & {\em Filter~3} & {\em Filter~4} &
      {\em Filter~5} & {\em Filter~6} & {\em Filter~7} & {\em Filter~8} \\
      {\em Orig} $\quad$
        & \includegraphics[width=0.13\textwidth]{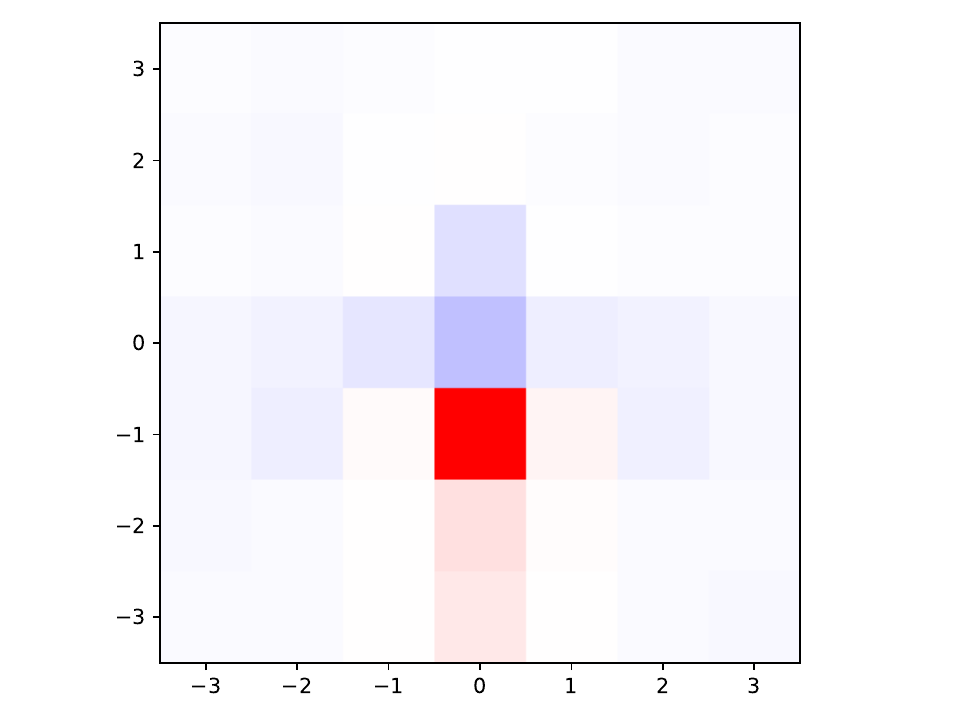}
        & \includegraphics[width=0.13\textwidth]{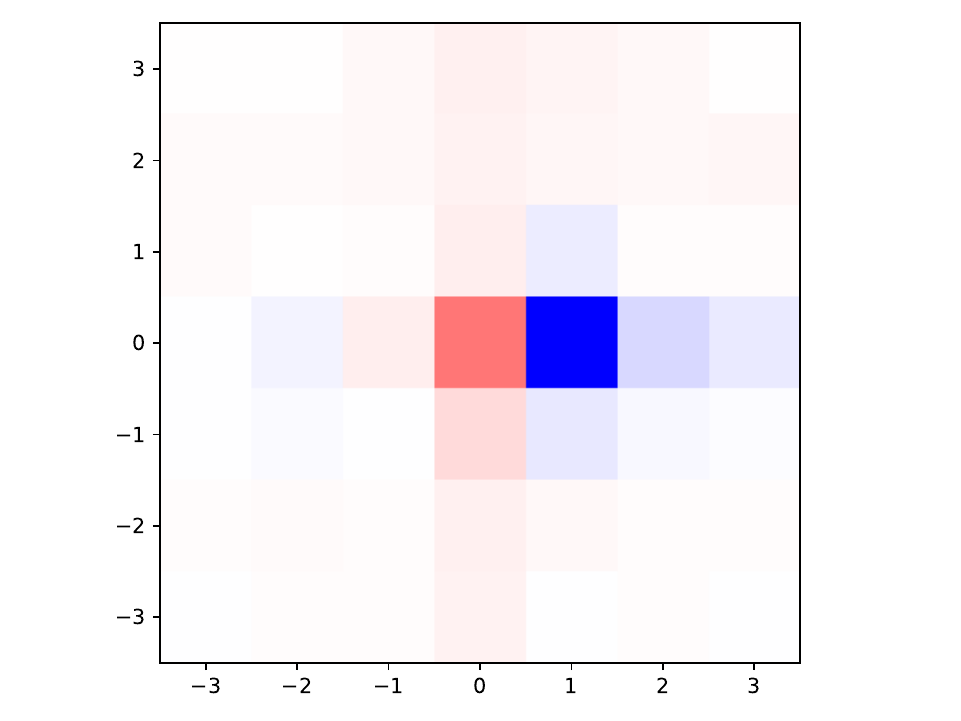}
        & \includegraphics[width=0.13\textwidth]{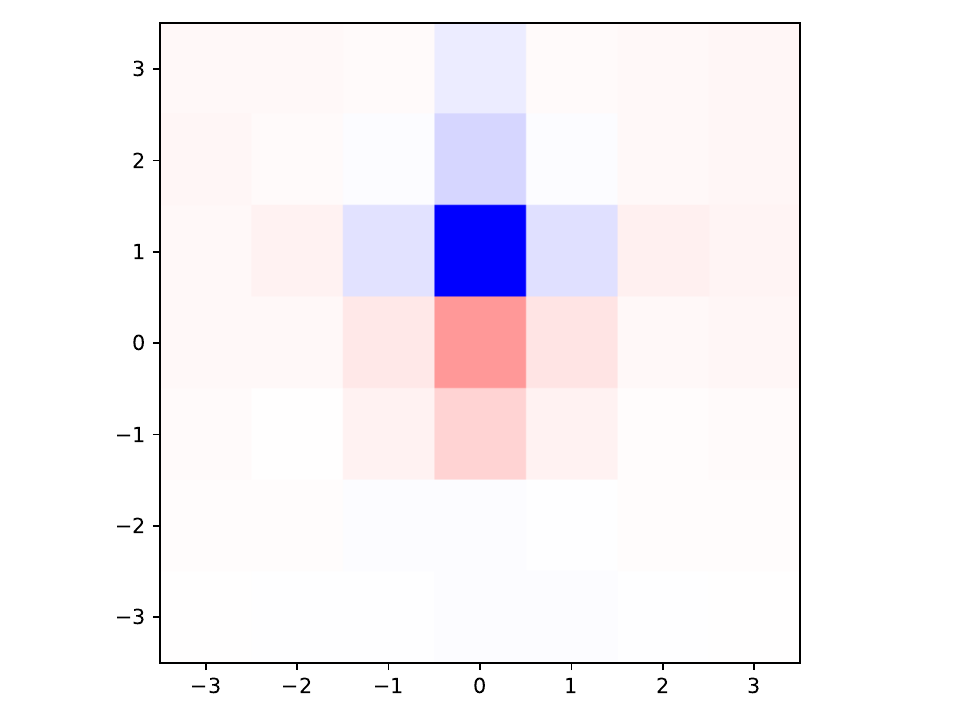}
        & \includegraphics[width=0.13\textwidth]{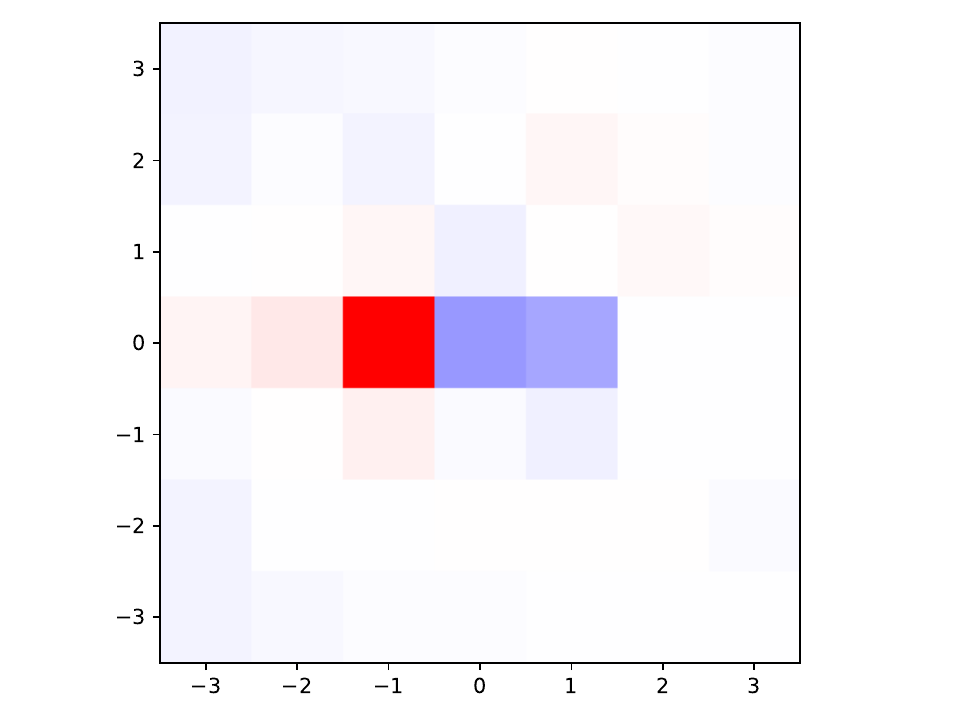} 
        & \includegraphics[width=0.13\textwidth]{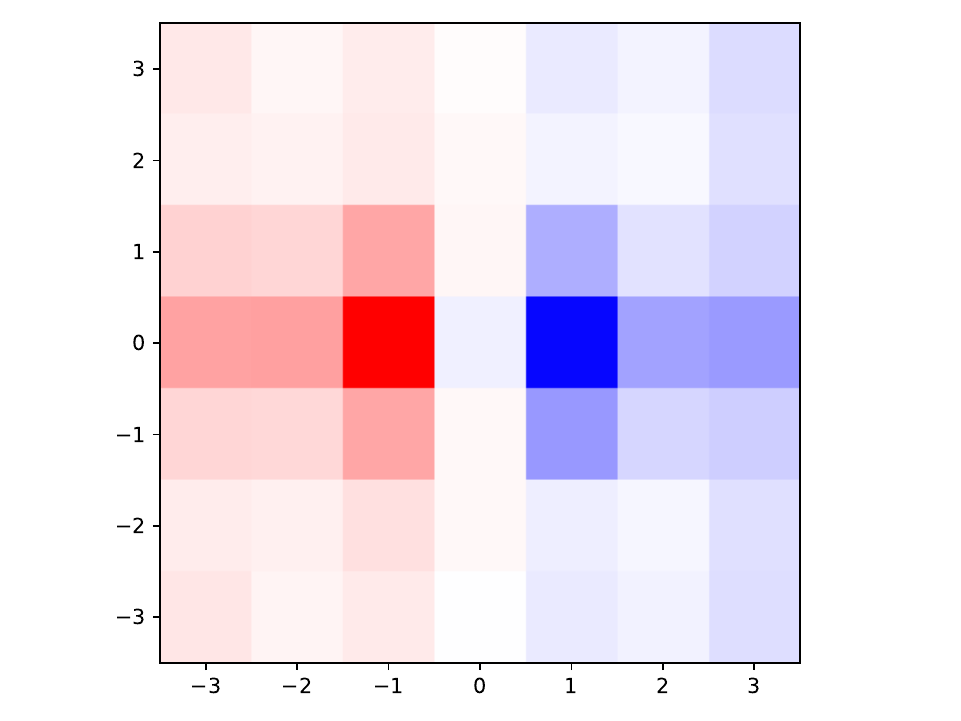}
        & \includegraphics[width=0.13\textwidth]{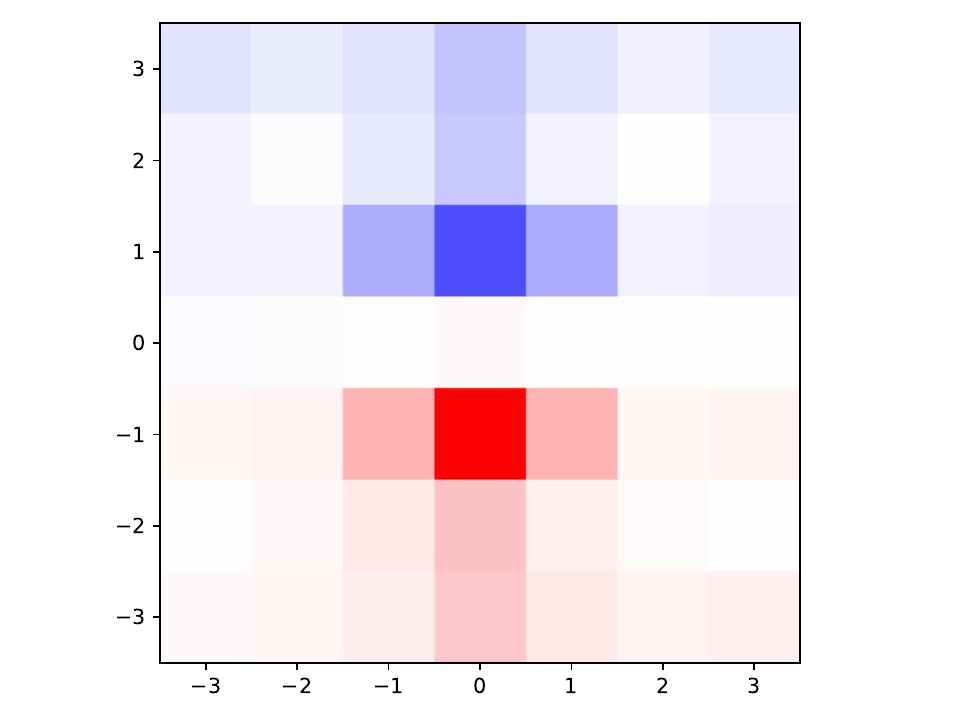}
        & \includegraphics[width=0.13\textwidth]{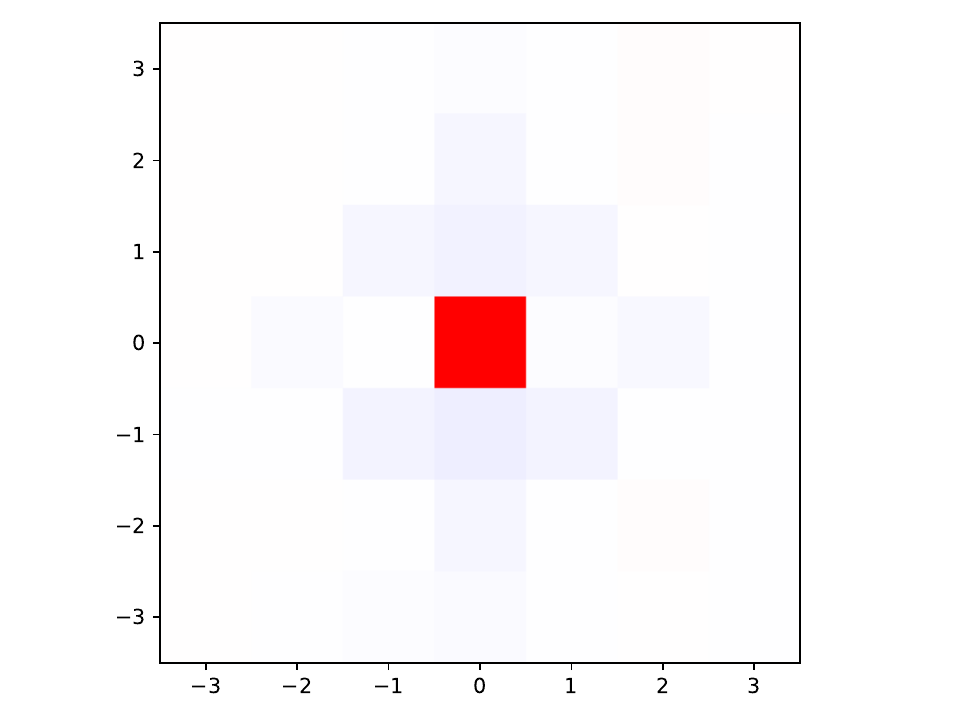}
        & \includegraphics[width=0.13\textwidth]{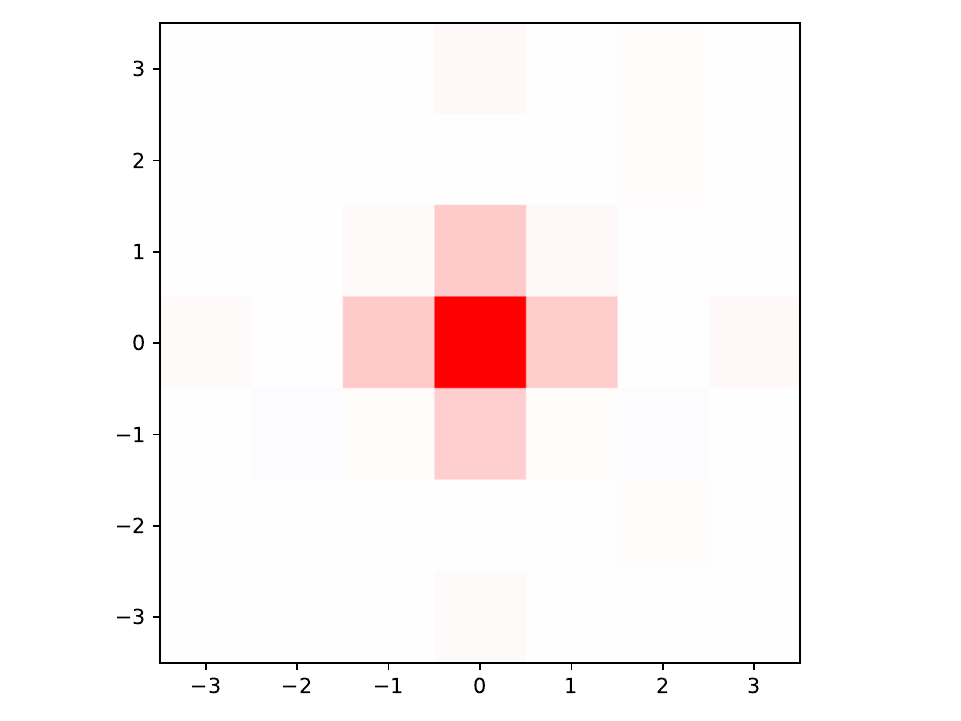} \\
      {\em A} $\quad$
        & \includegraphics[width=0.13\textwidth]{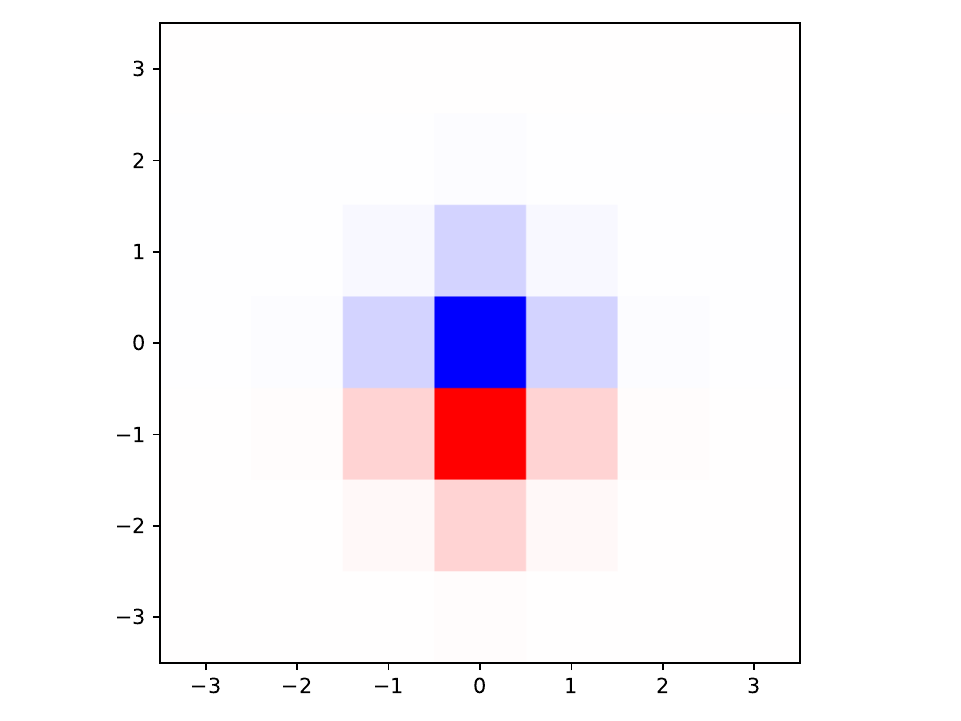}
        & \includegraphics[width=0.13\textwidth]{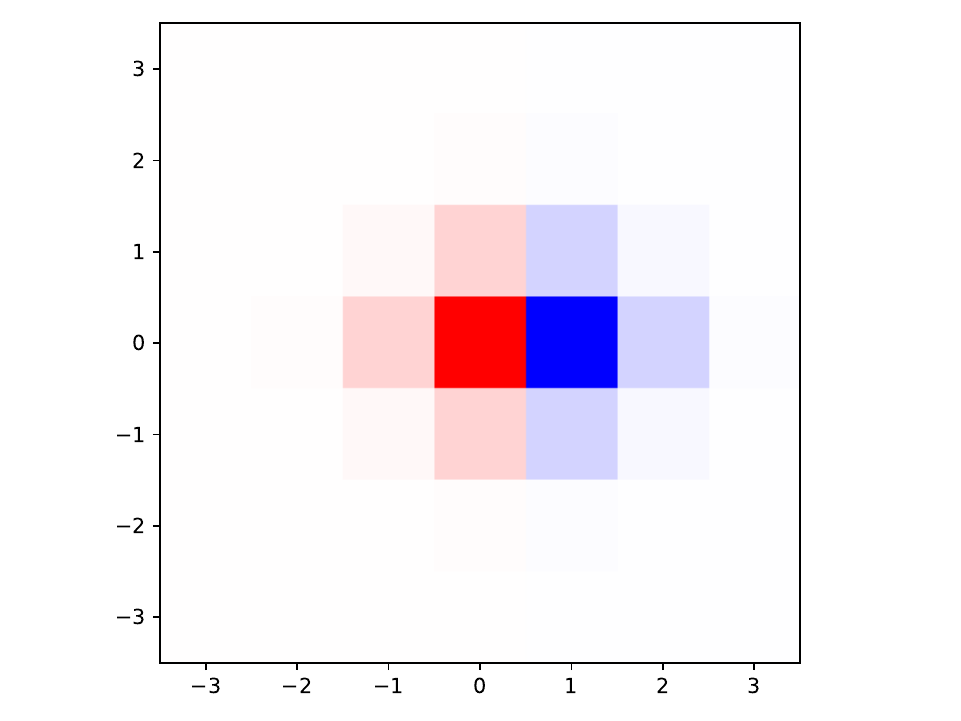}
        & \includegraphics[width=0.13\textwidth]{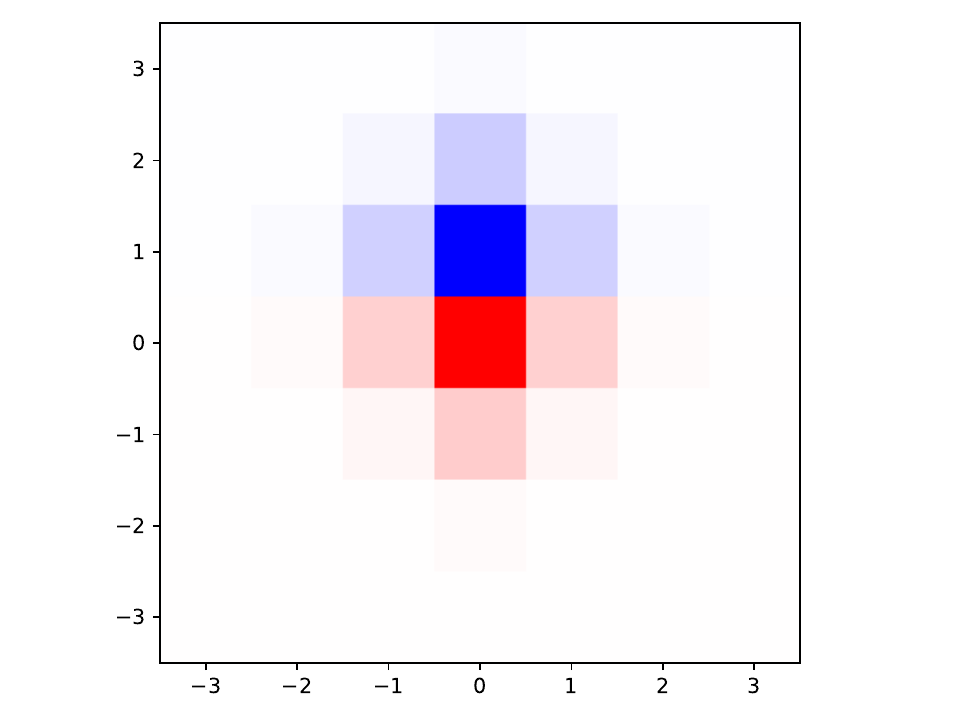}
        & \includegraphics[width=0.13\textwidth]{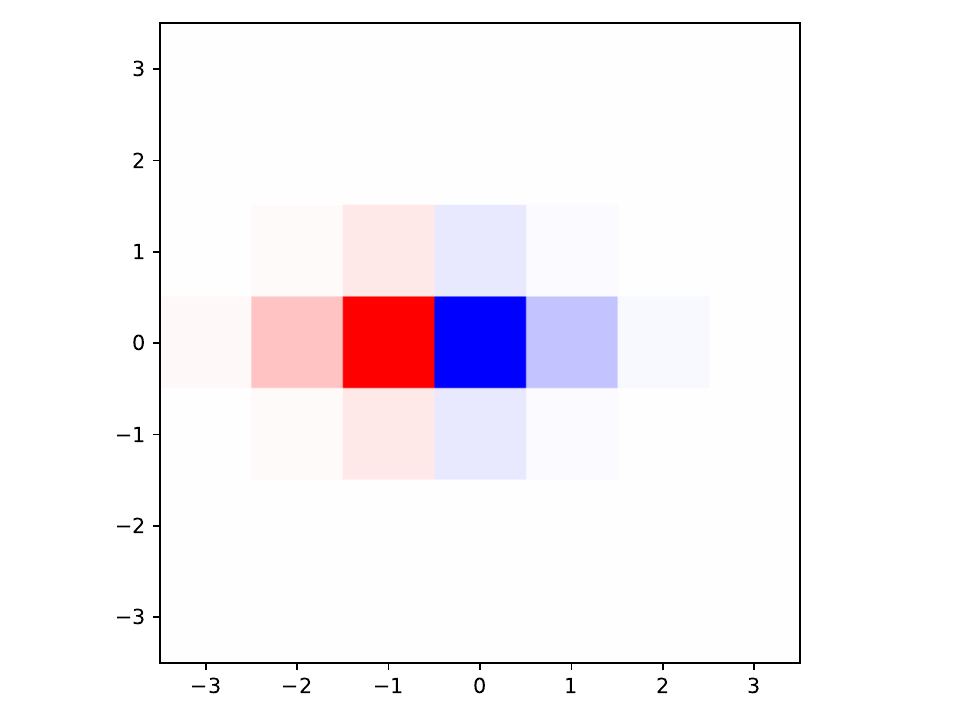} 
        & \includegraphics[width=0.13\textwidth]{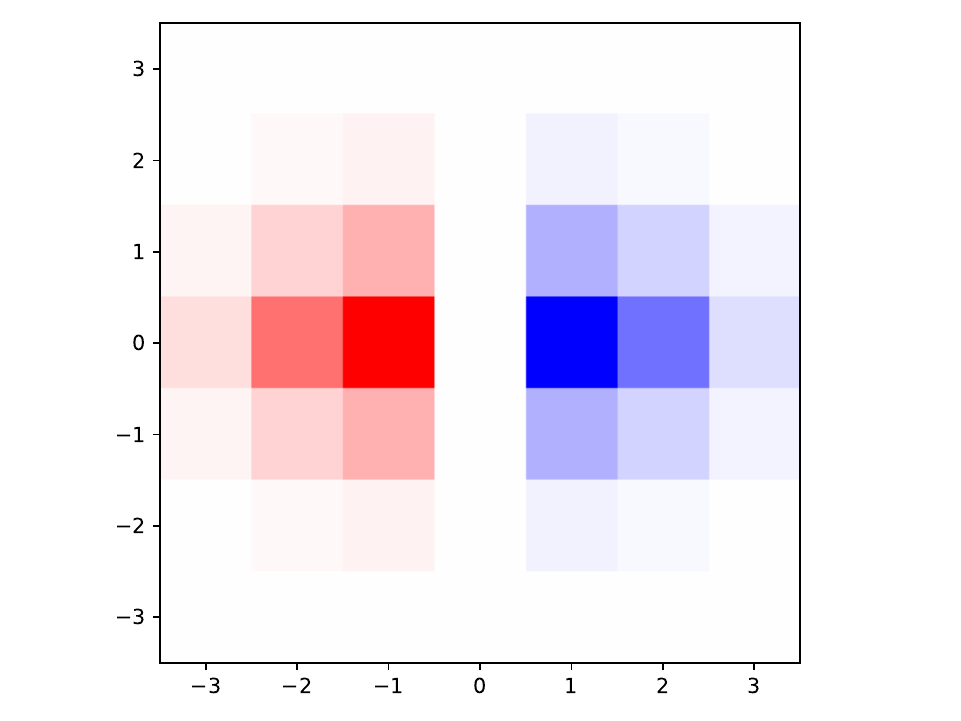}
        & \includegraphics[width=0.13\textwidth]{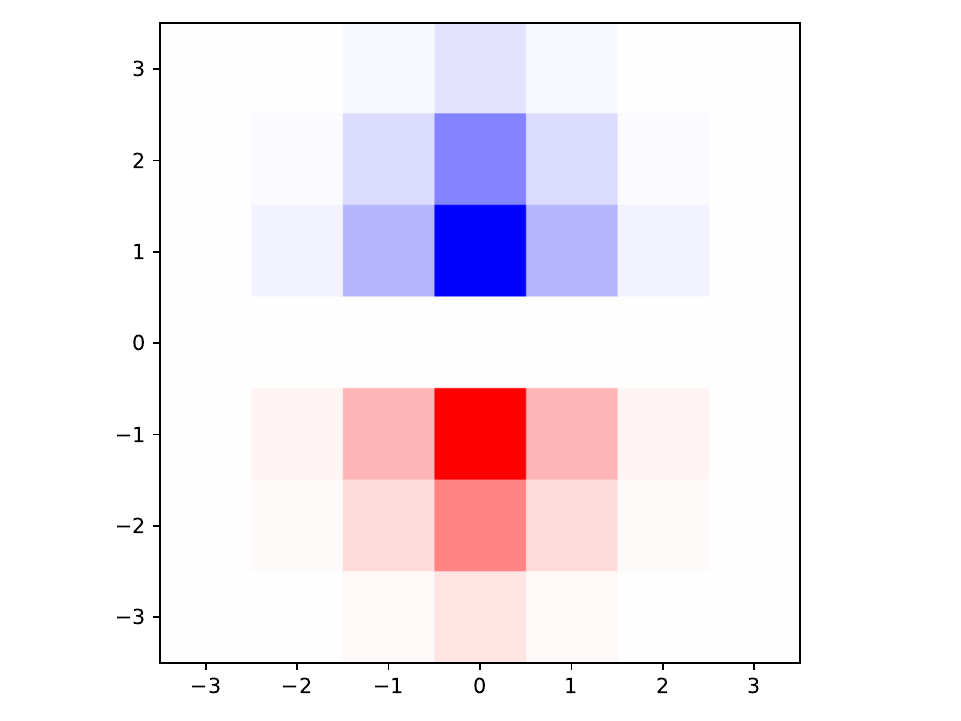}
        & \includegraphics[width=0.13\textwidth]{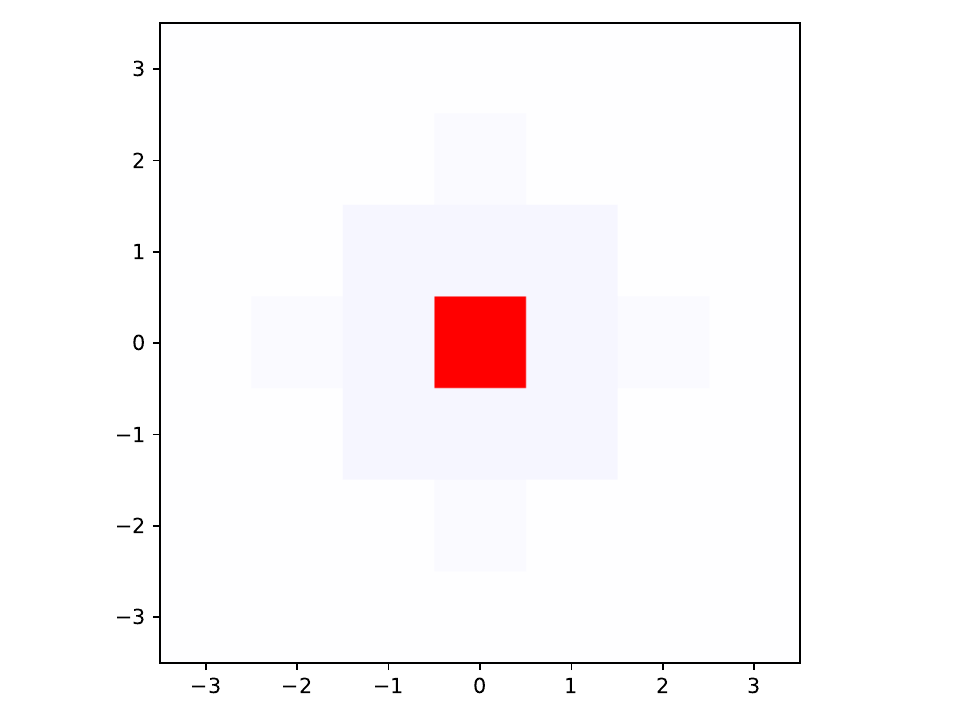}
        & \includegraphics[width=0.13\textwidth]{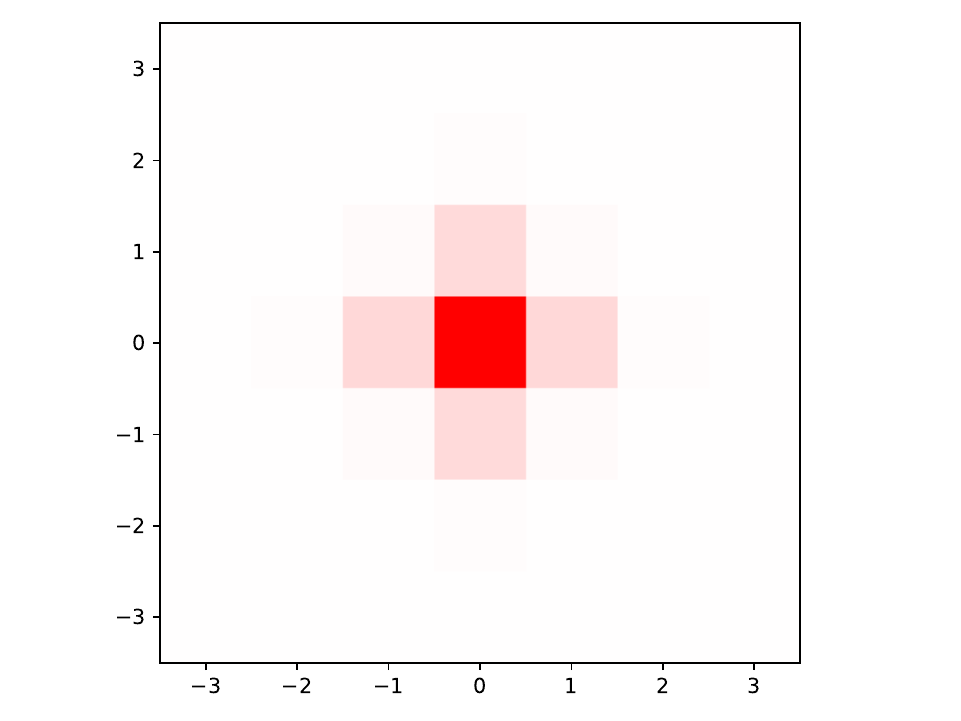} \\
      {\em B} $\quad$
       & \includegraphics[width=0.13\textwidth]{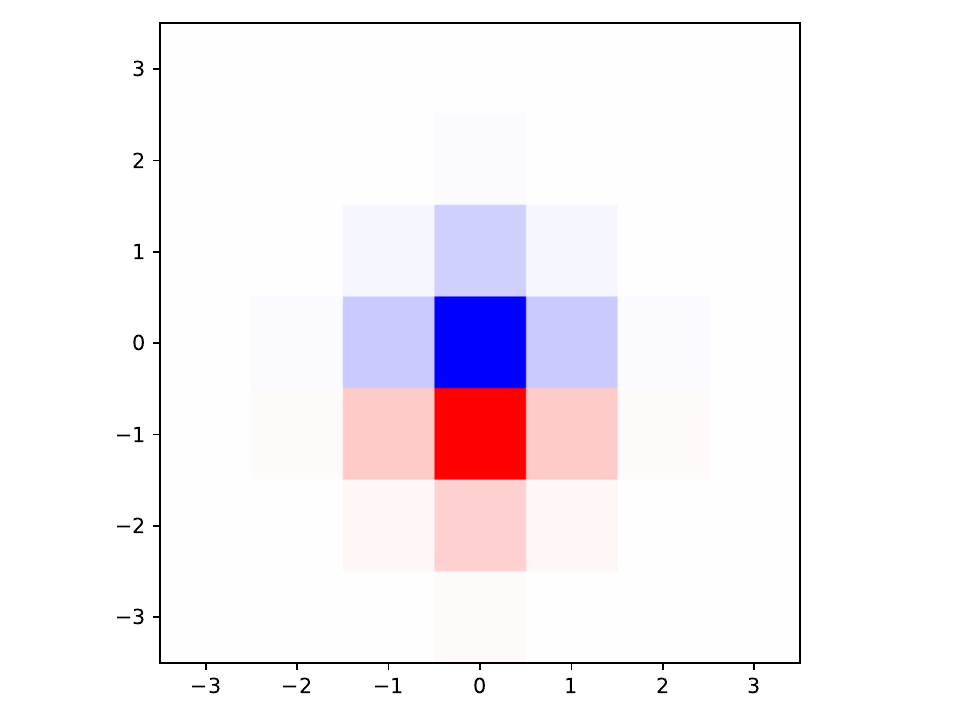}
        & \includegraphics[width=0.13\textwidth]{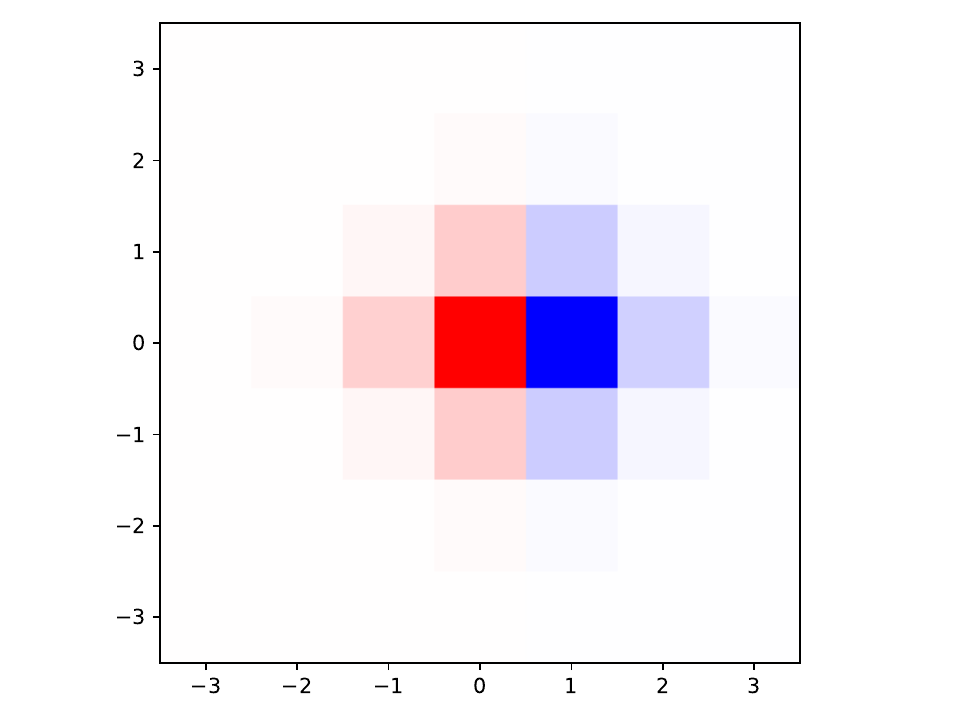}
        & \includegraphics[width=0.13\textwidth]{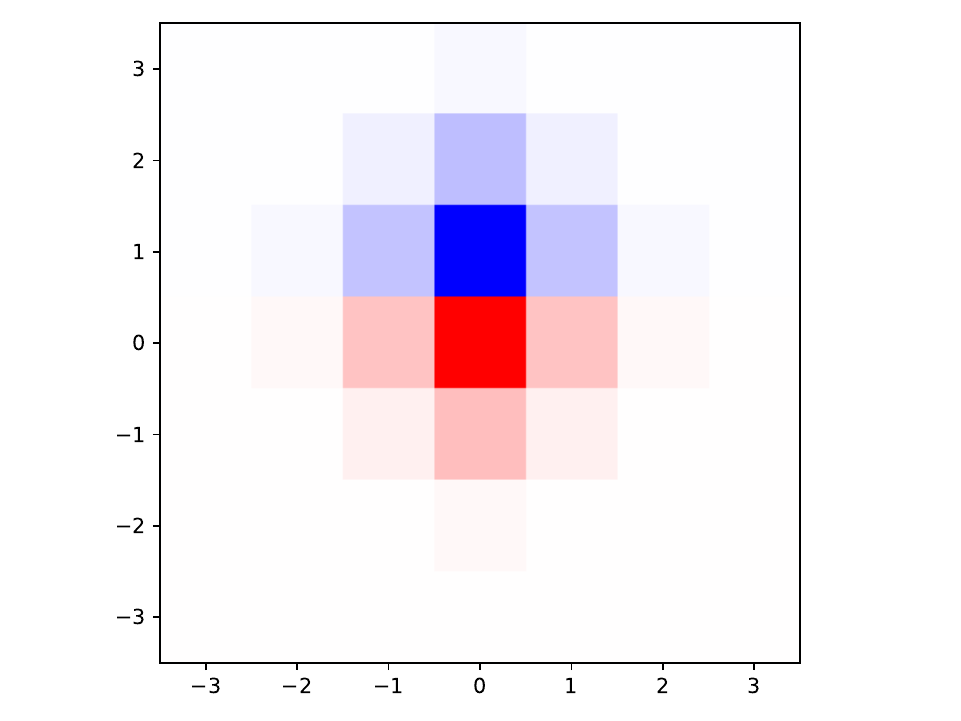}
        & \includegraphics[width=0.13\textwidth]{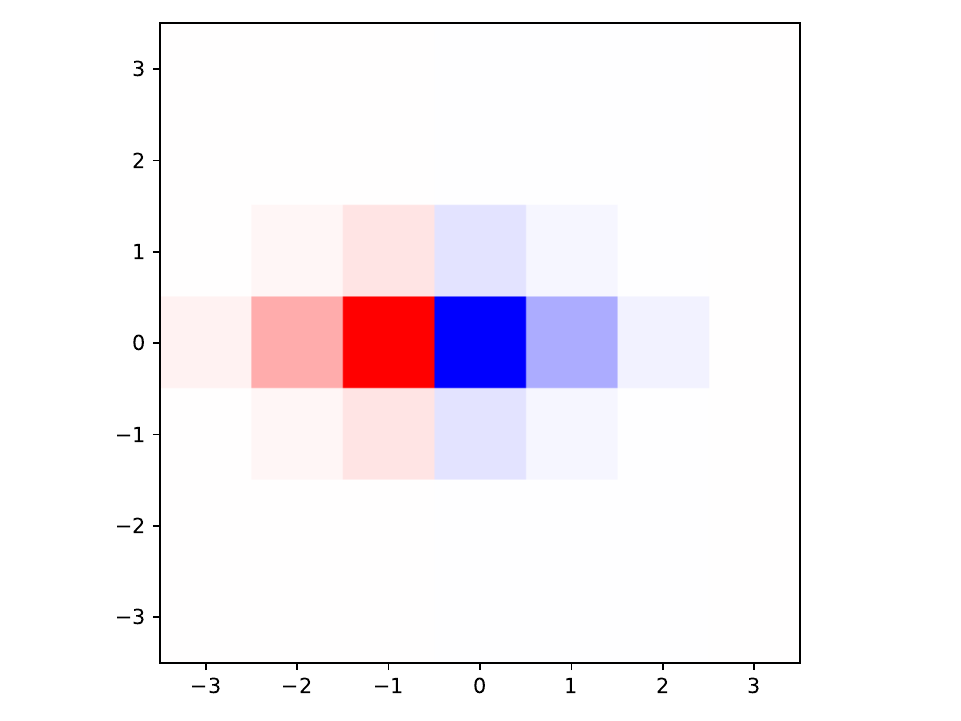} 
       & \includegraphics[width=0.13\textwidth]{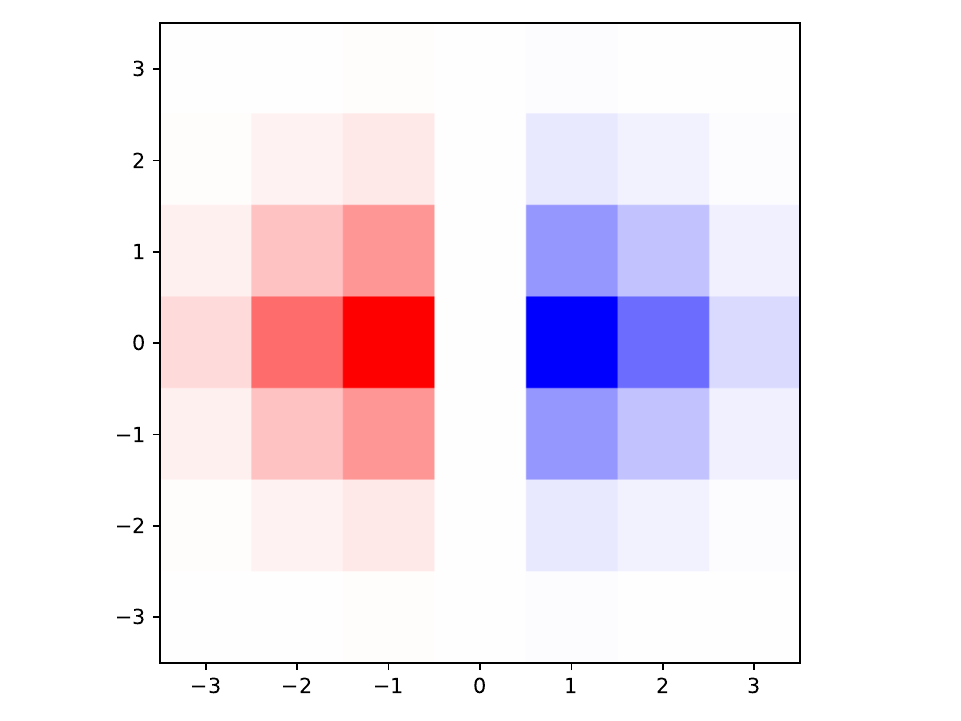}
        & \includegraphics[width=0.13\textwidth]{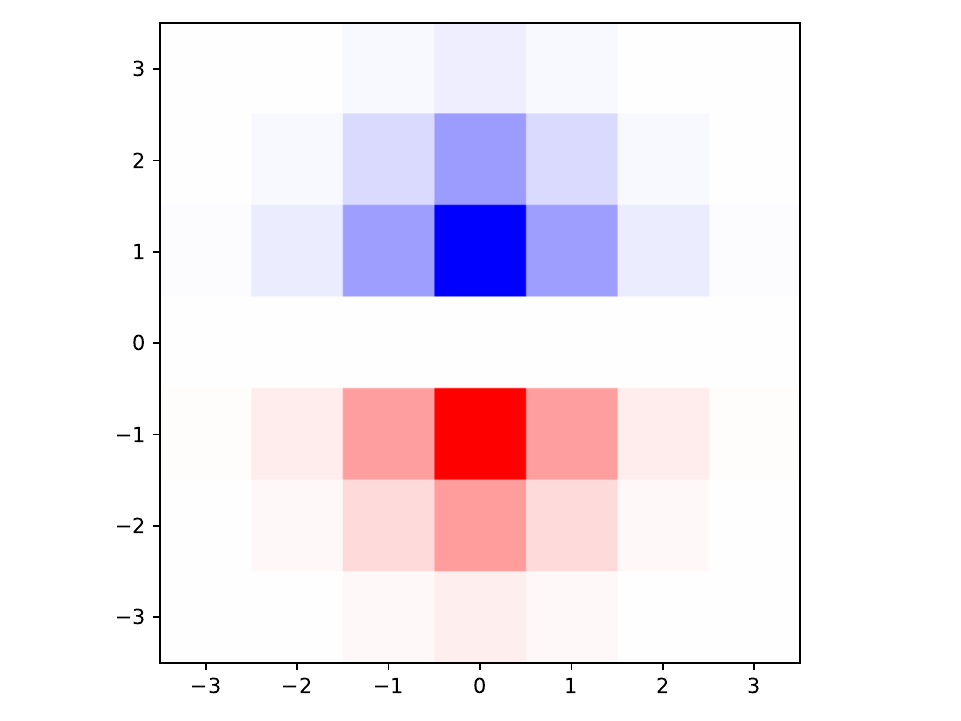}
        & \includegraphics[width=0.13\textwidth]{filter-jointscalefroml2diff7-bluered-eps-converted-to.pdf}
        & \includegraphics[width=0.13\textwidth]{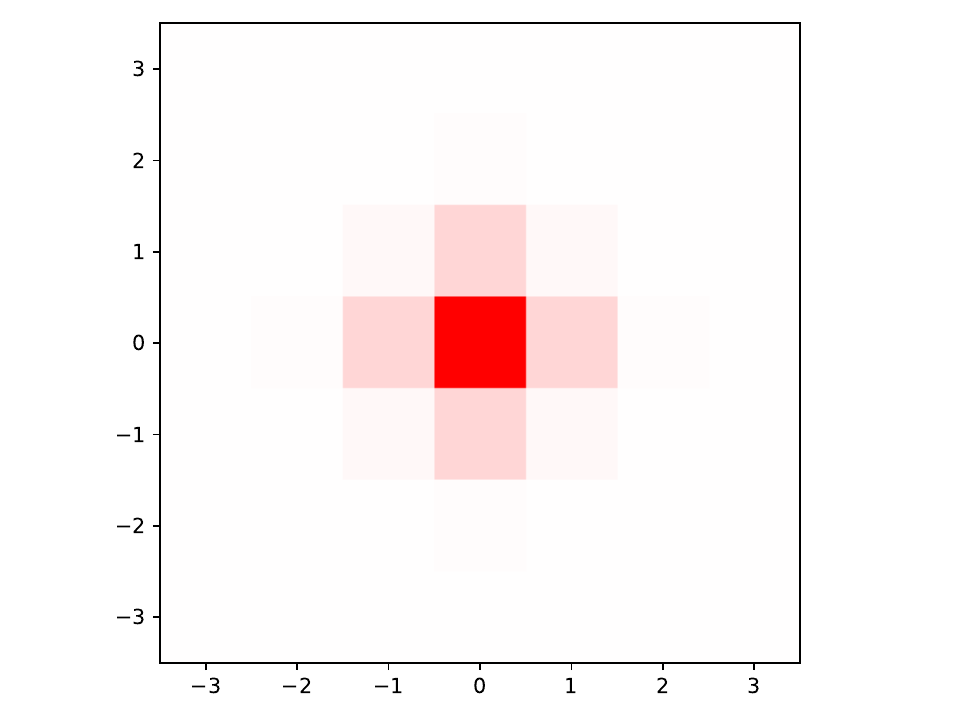} \\
      {\em C1} $\quad$
       & \includegraphics[width=0.13\textwidth]{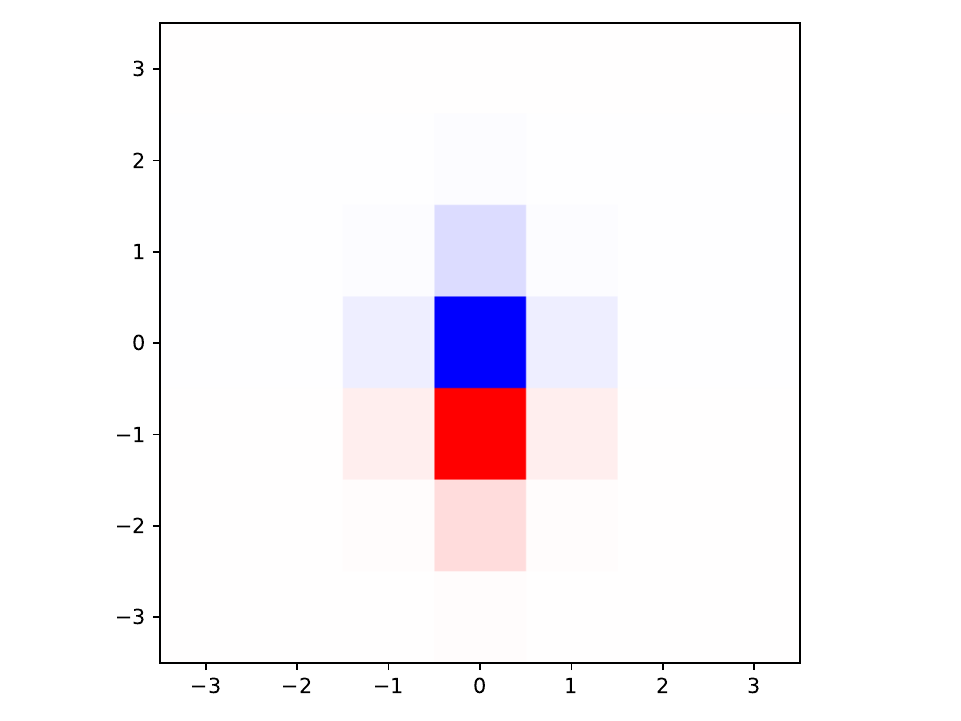}
        & \includegraphics[width=0.13\textwidth]{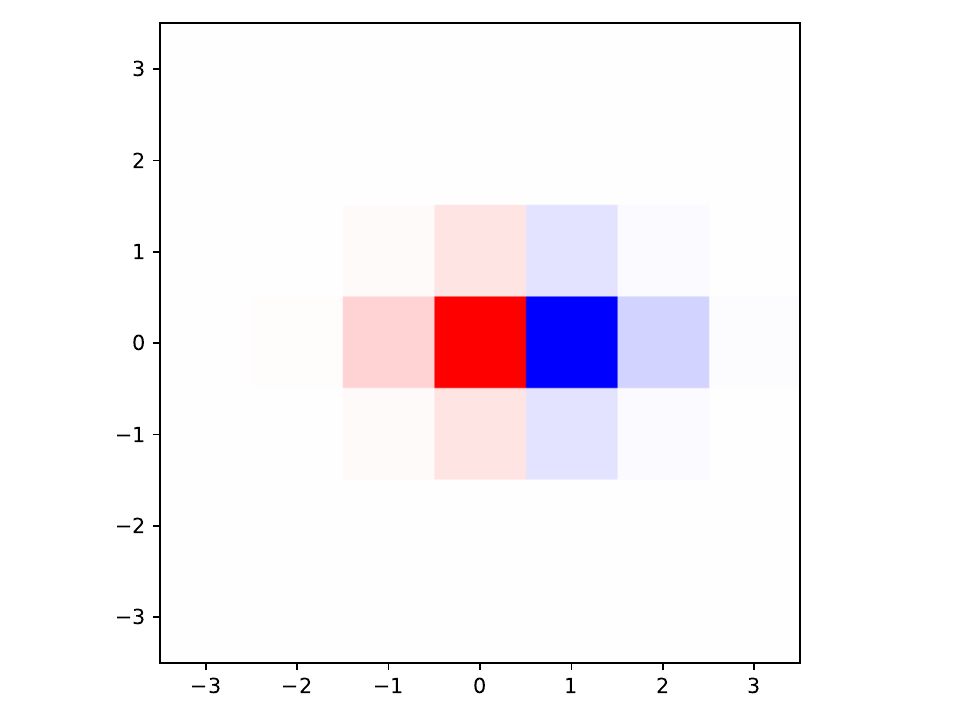}
        & \includegraphics[width=0.13\textwidth]{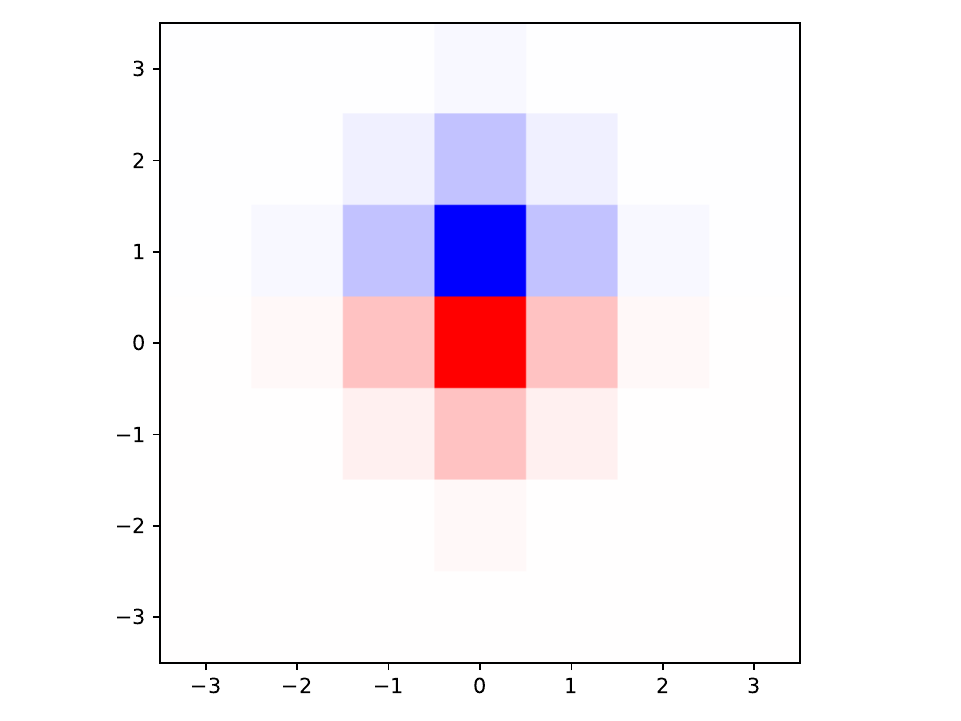}
        & \includegraphics[width=0.13\textwidth]{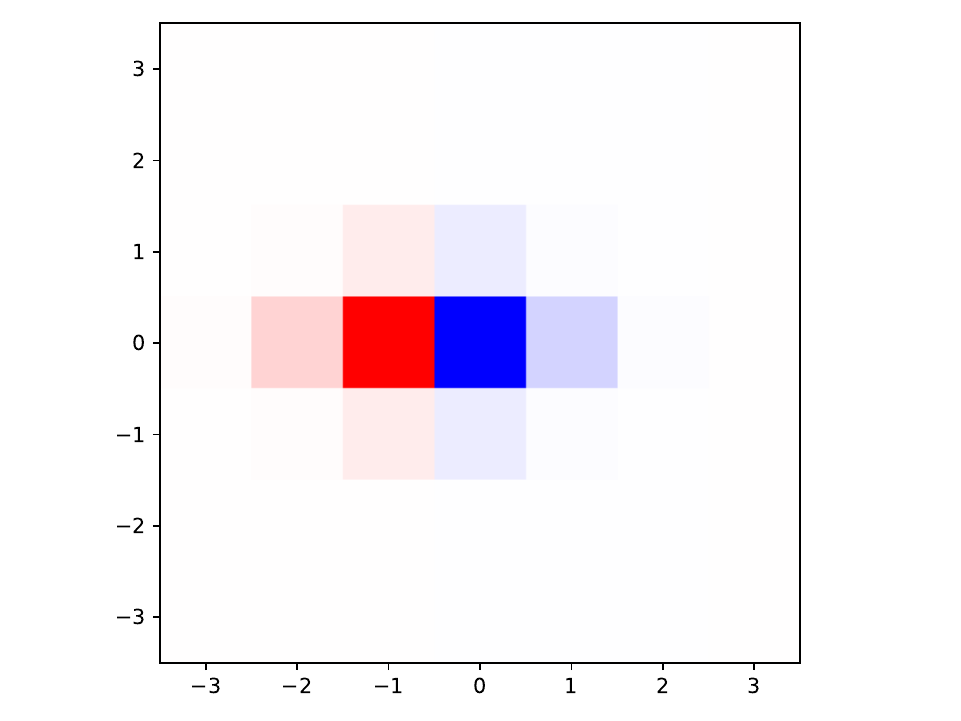} 
       & \includegraphics[width=0.13\textwidth]{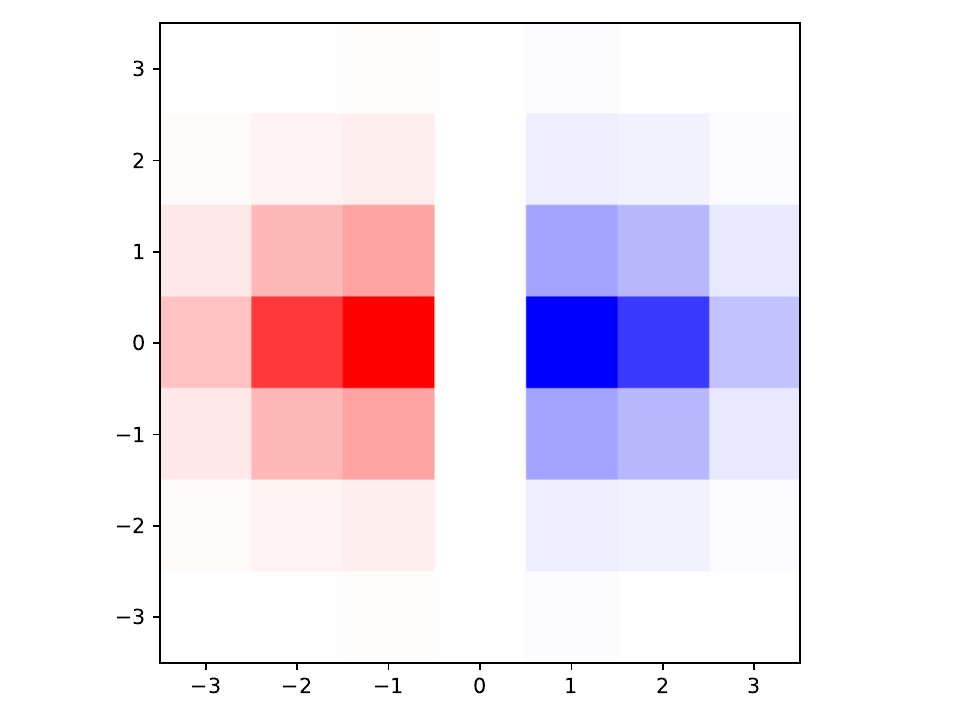}
        & \includegraphics[width=0.13\textwidth]{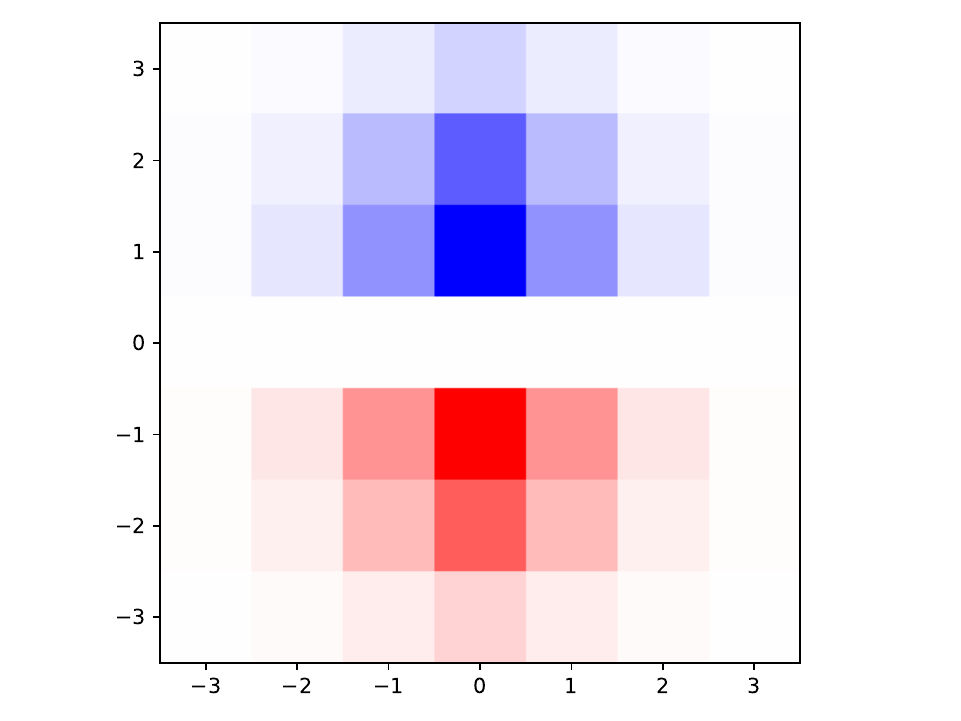}
        & \includegraphics[width=0.13\textwidth]{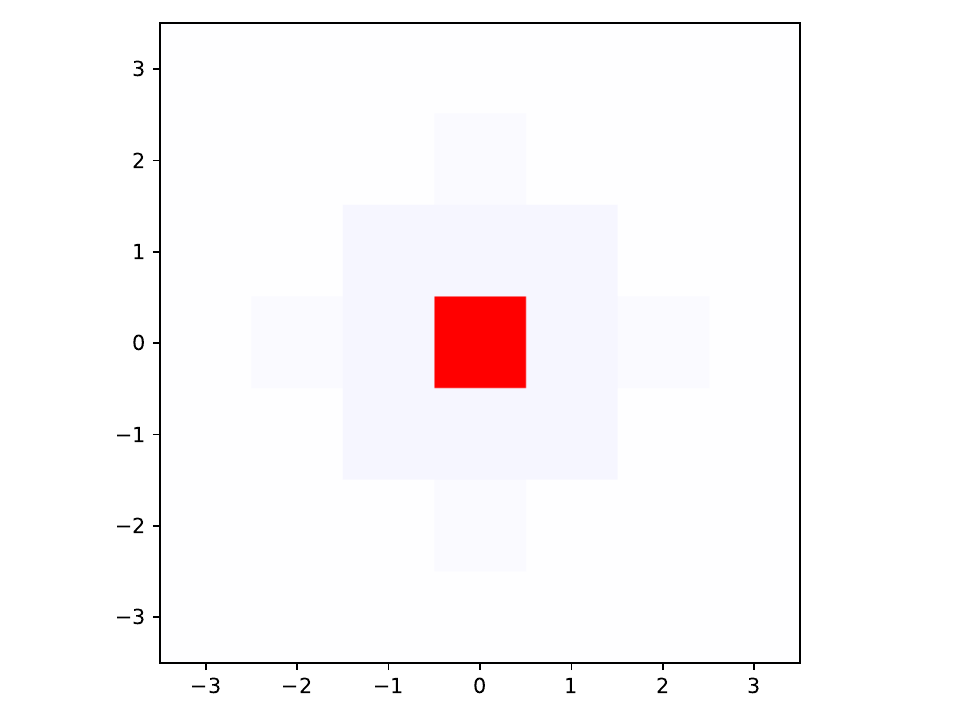}
        & \includegraphics[width=0.13\textwidth]{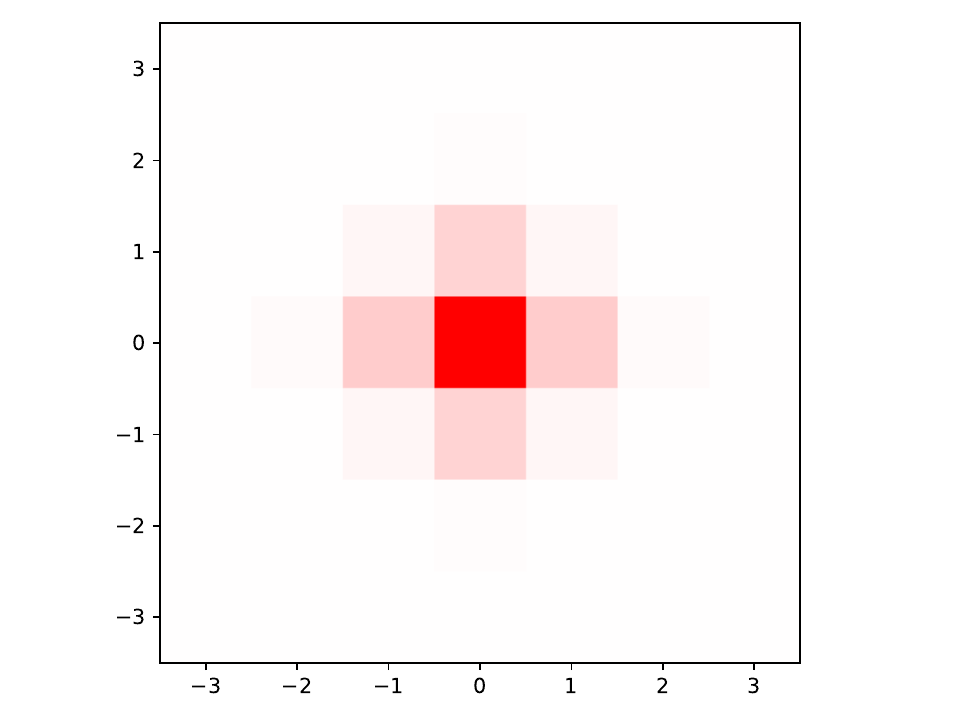} \\
      {\em C2} $\quad$
       & \includegraphics[width=0.13\textwidth]{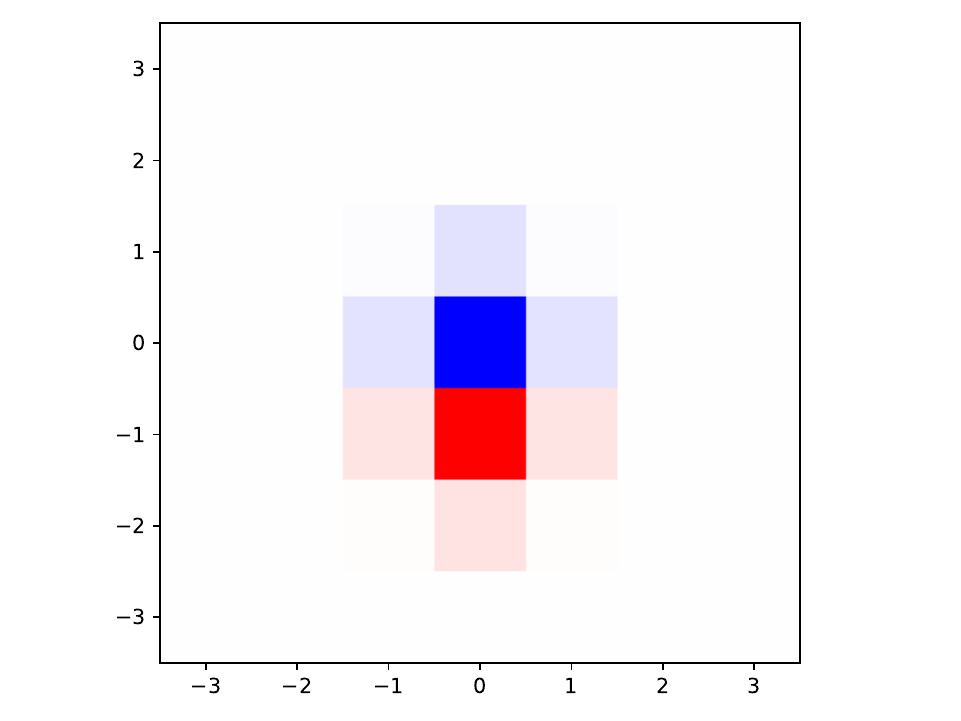}
        & \includegraphics[width=0.13\textwidth]{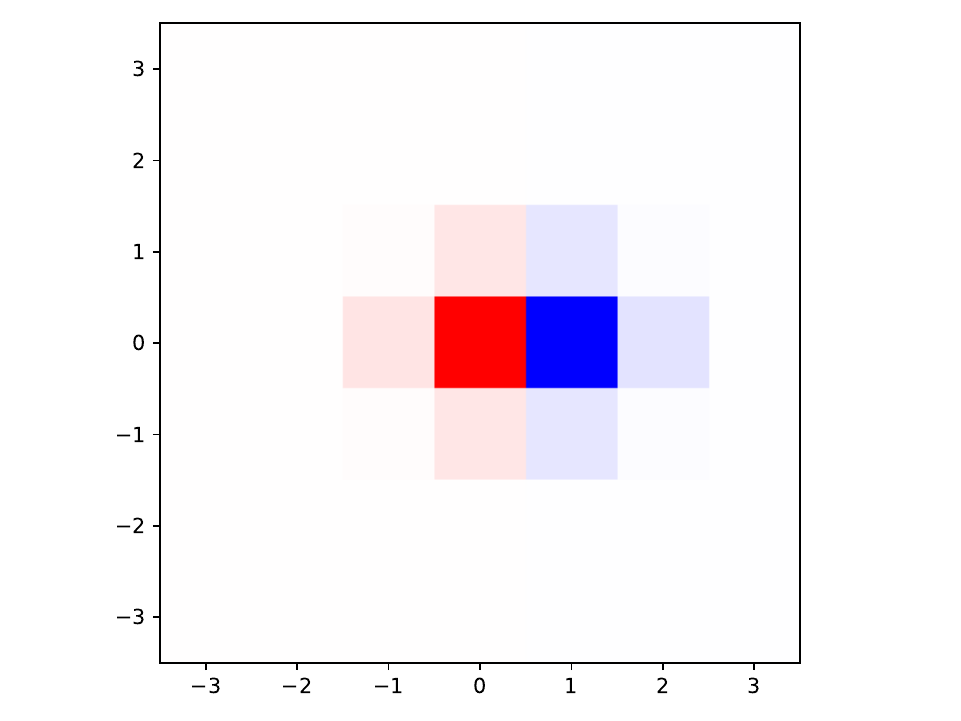}
        & \includegraphics[width=0.13\textwidth]{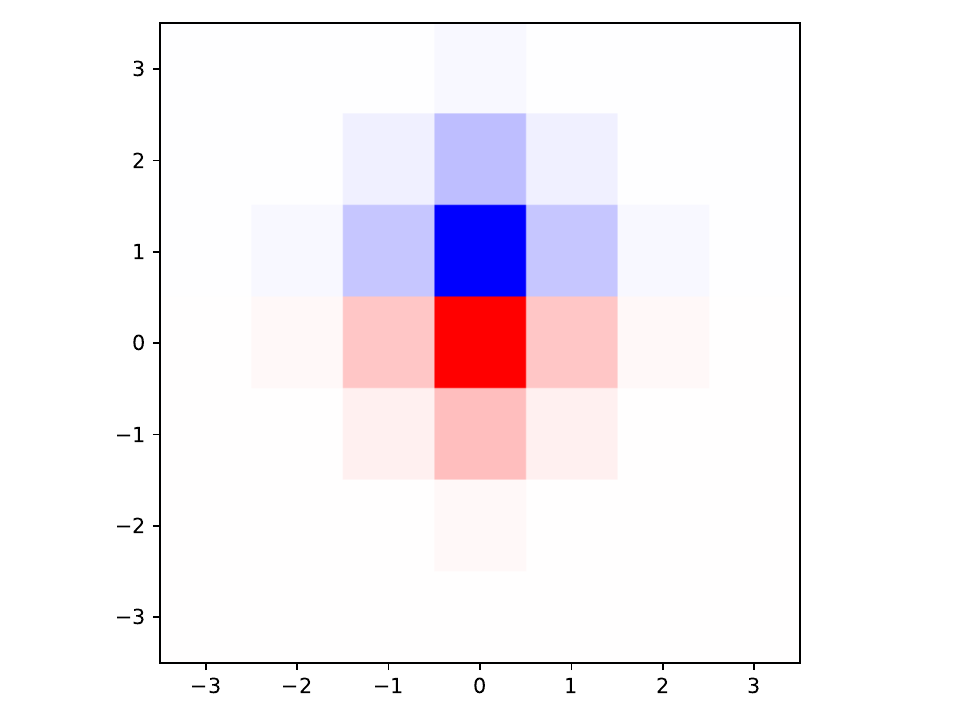}
        & \includegraphics[width=0.13\textwidth]{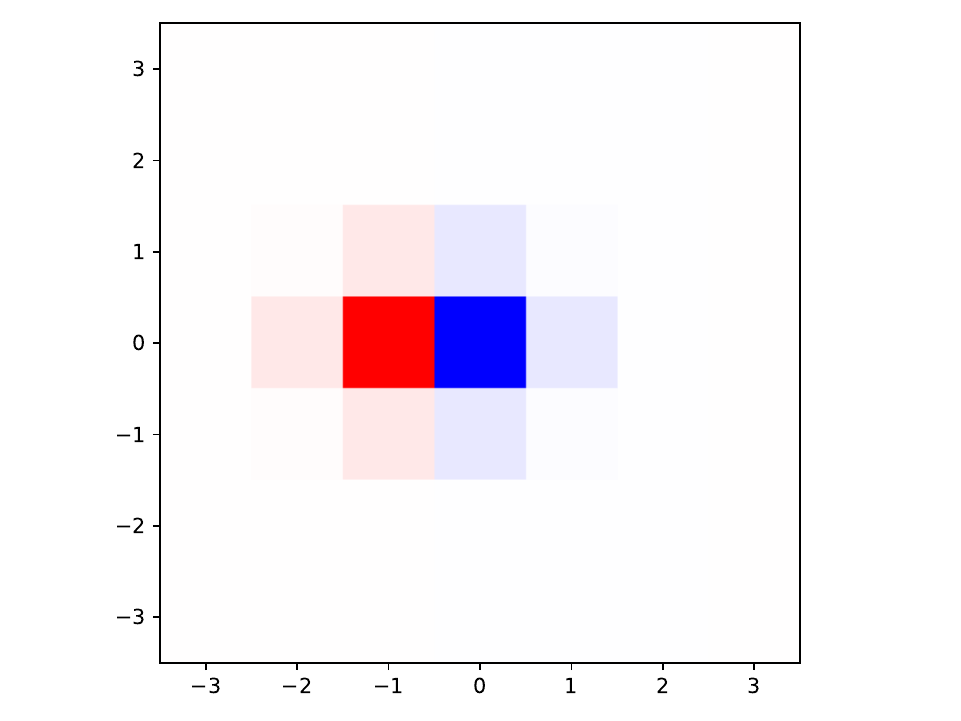} 
       & \includegraphics[width=0.13\textwidth]{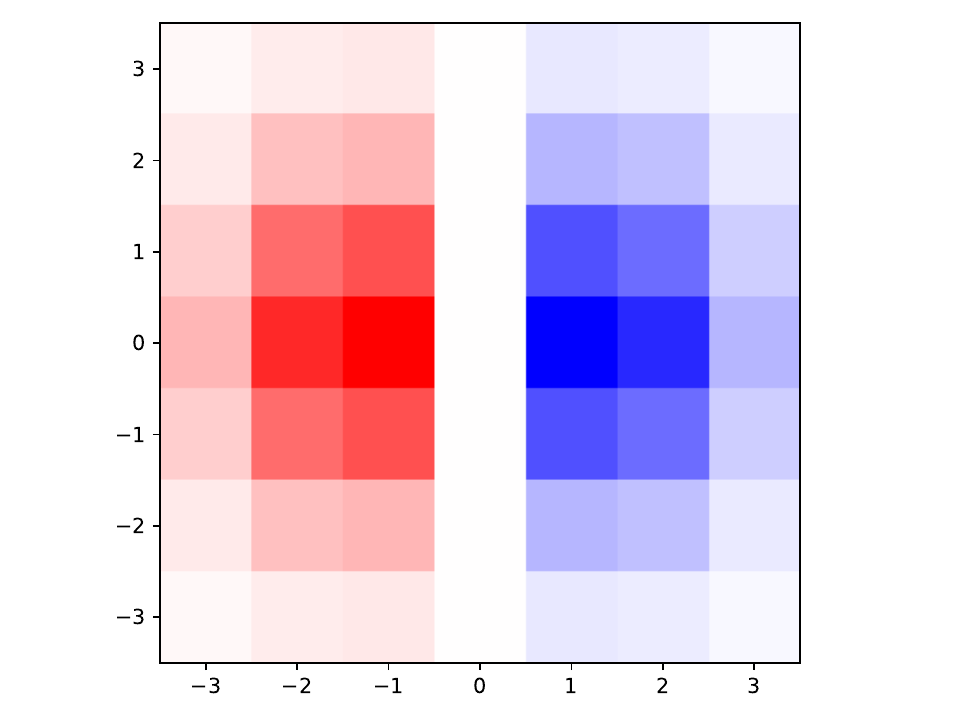}
        & \includegraphics[width=0.13\textwidth]{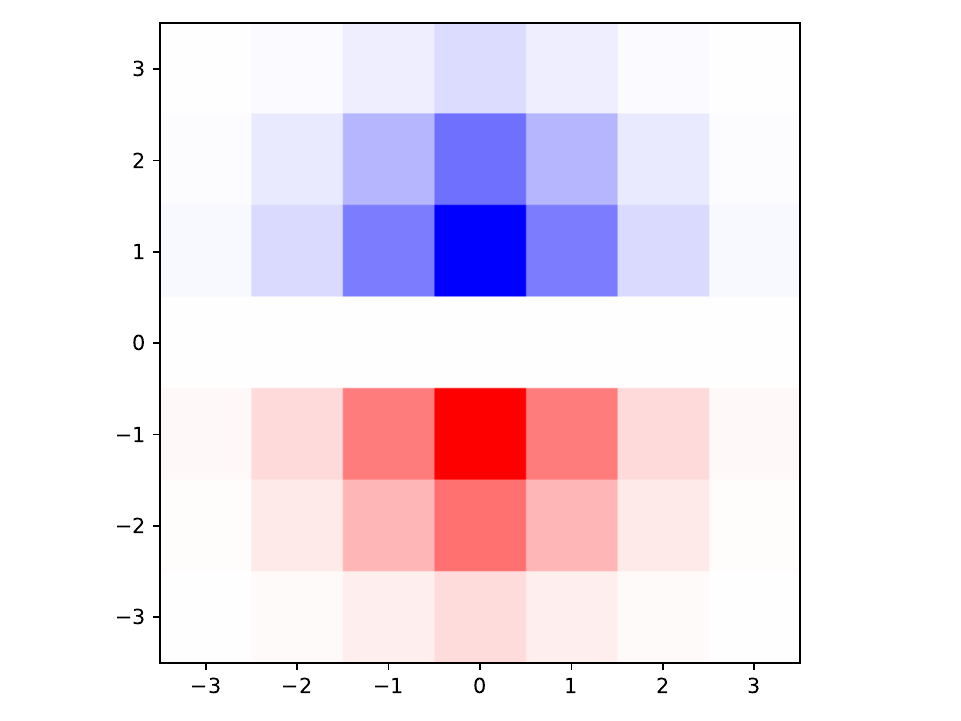}
        & \includegraphics[width=0.13\textwidth]{filter-jointscalefroml1diff7-bluered-eps-converted-to.pdf}
        & \includegraphics[width=0.13\textwidth]{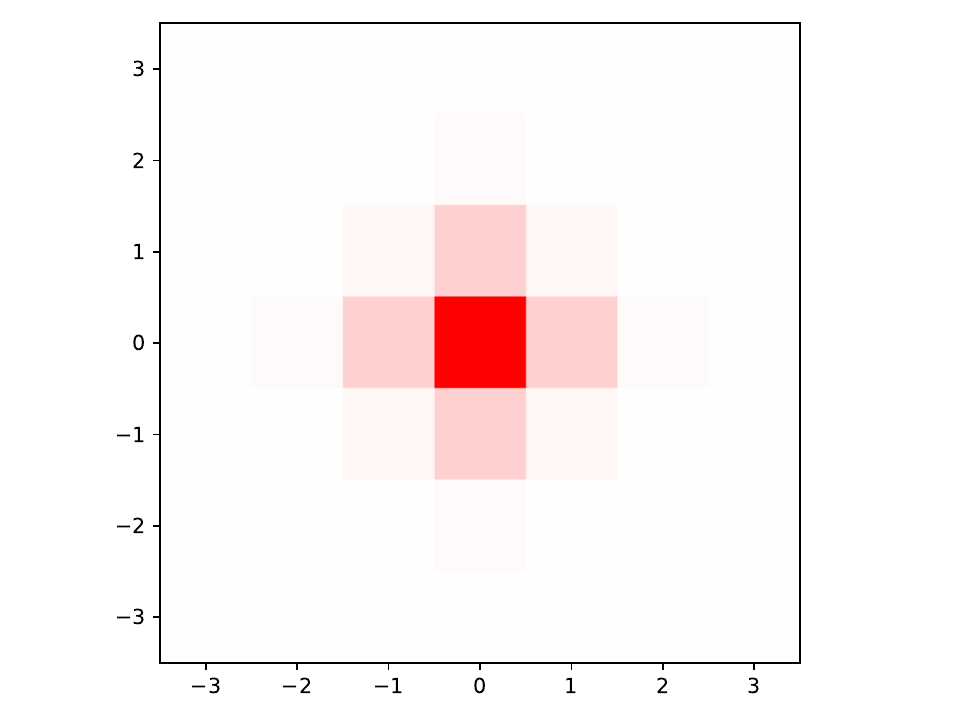} \\
      {\em D1} $\quad$
        & \includegraphics[width=0.13\textwidth]{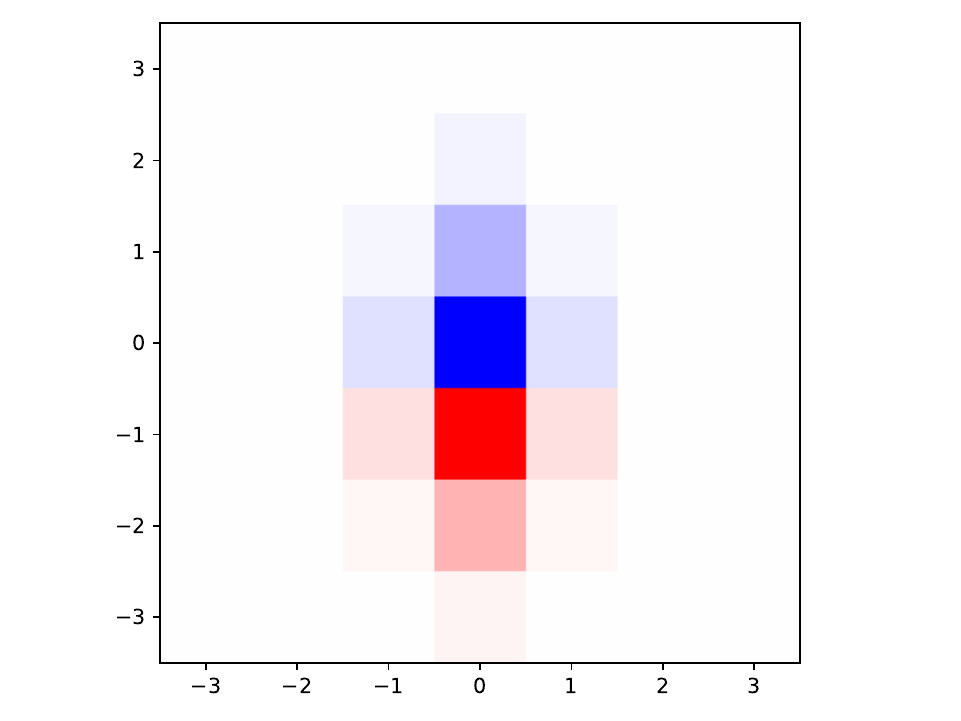}
        & \includegraphics[width=0.13\textwidth]{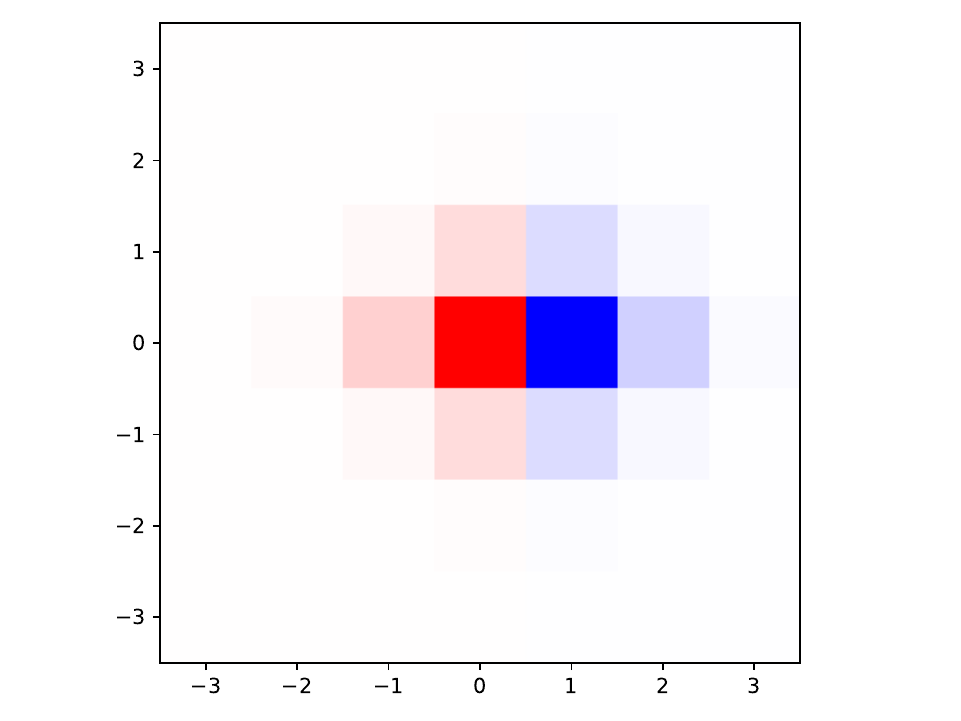}
        & \includegraphics[width=0.13\textwidth]{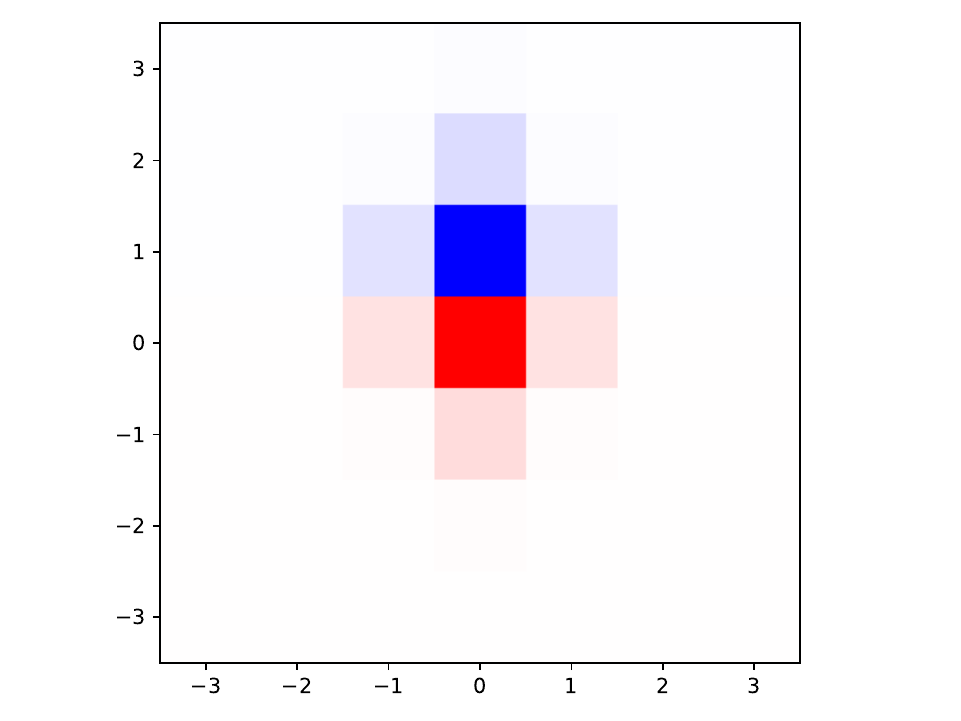}
        & \includegraphics[width=0.13\textwidth]{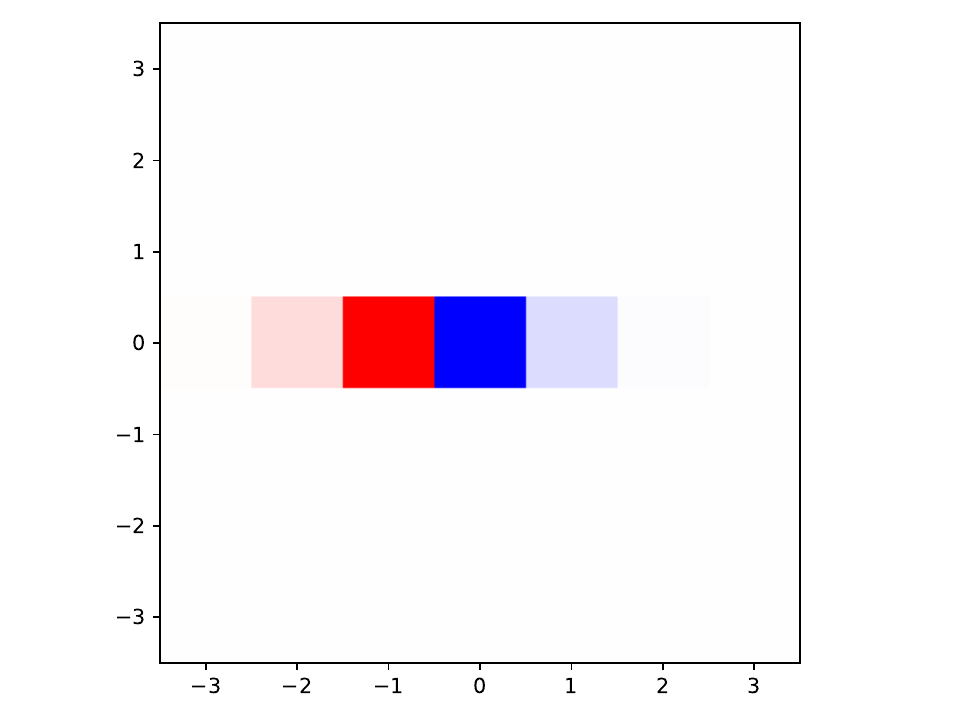} 
       & \includegraphics[width=0.13\textwidth]{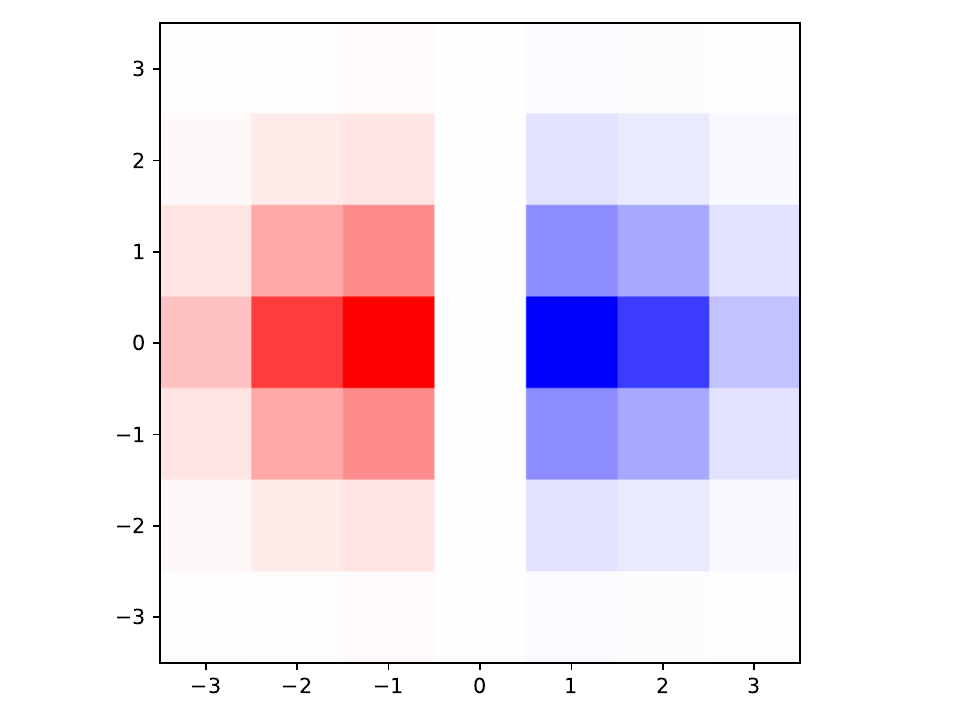}
        & \includegraphics[width=0.13\textwidth]{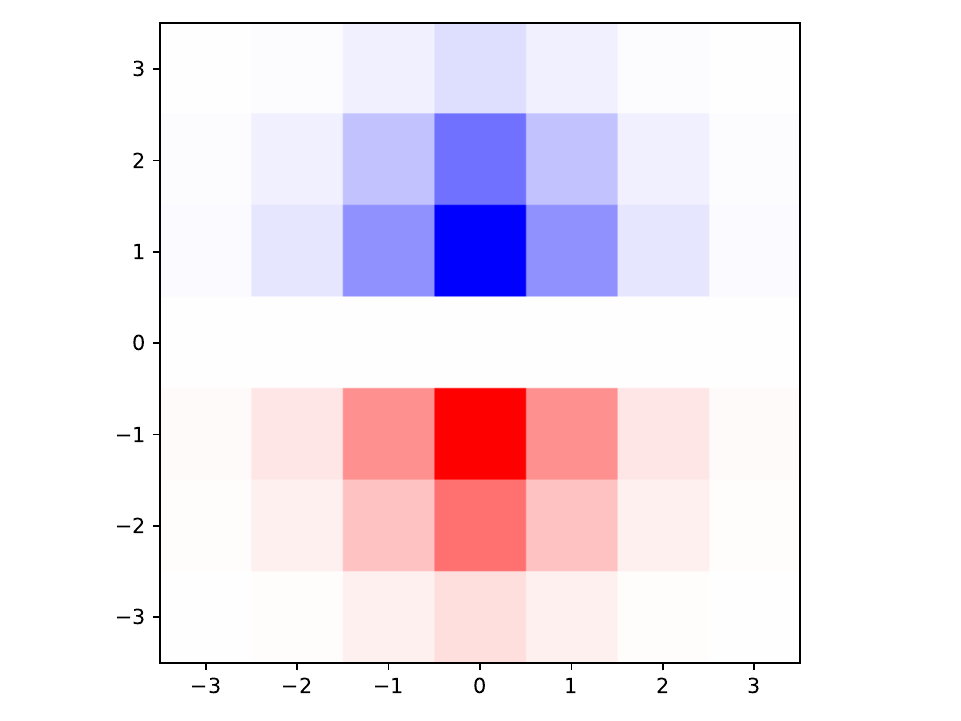}
        & \includegraphics[width=0.13\textwidth]{filter-jointscalefroml2diff7-bluered-eps-converted-to.pdf}
        & \includegraphics[width=0.13\textwidth]{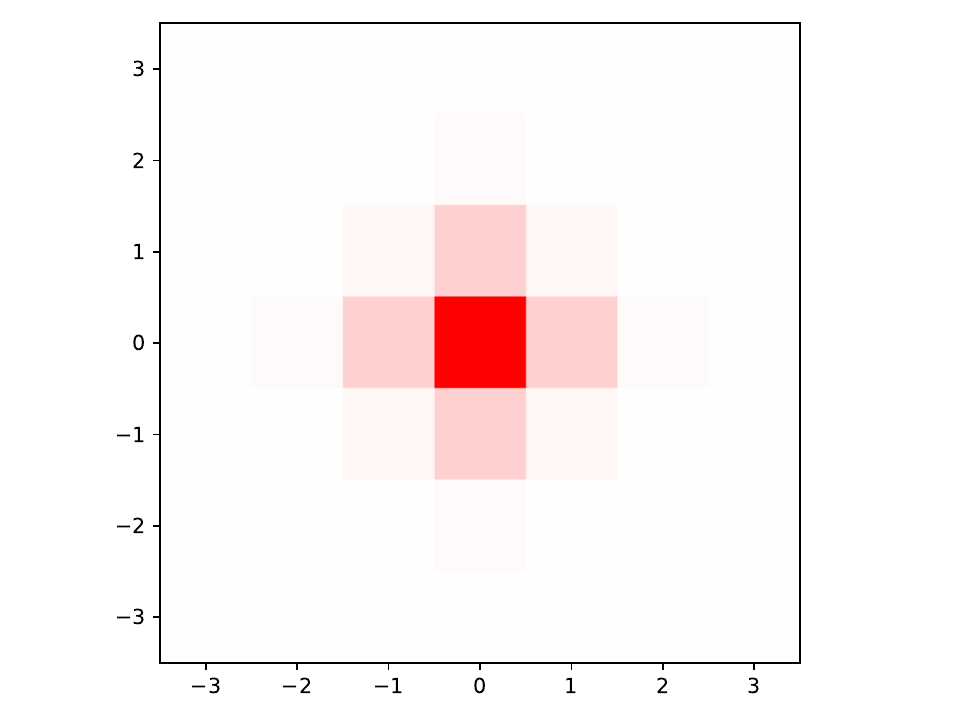} \\
      {\em D2} $\quad$
        & \includegraphics[width=0.13\textwidth]{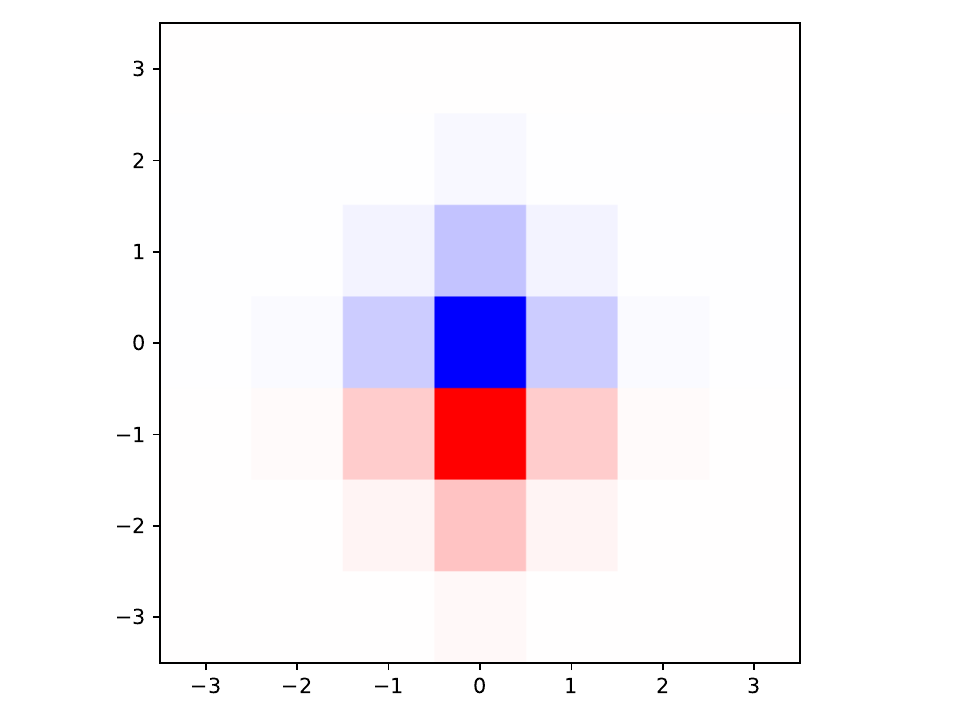}
        & \includegraphics[width=0.13\textwidth]{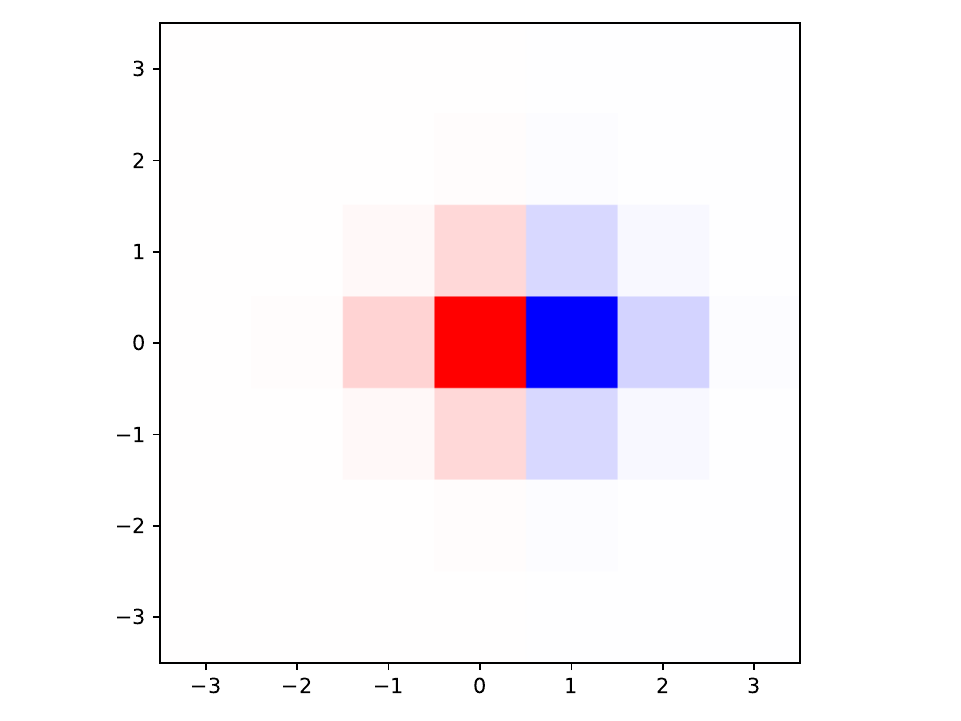}
        & \includegraphics[width=0.13\textwidth]{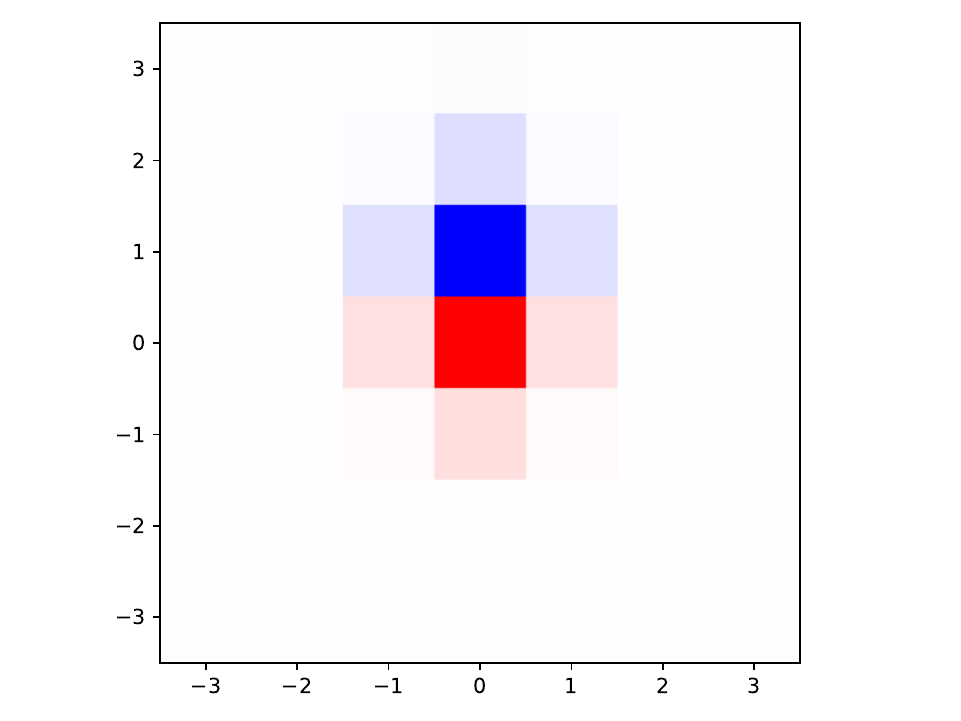}
        & \includegraphics[width=0.13\textwidth]{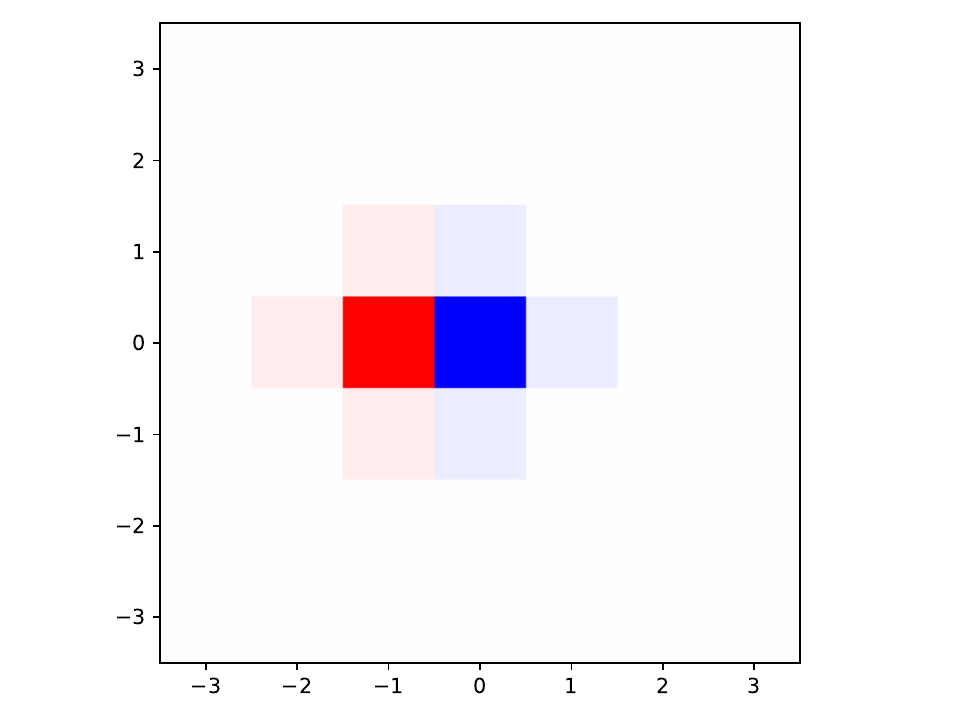} 
        & \includegraphics[width=0.13\textwidth]{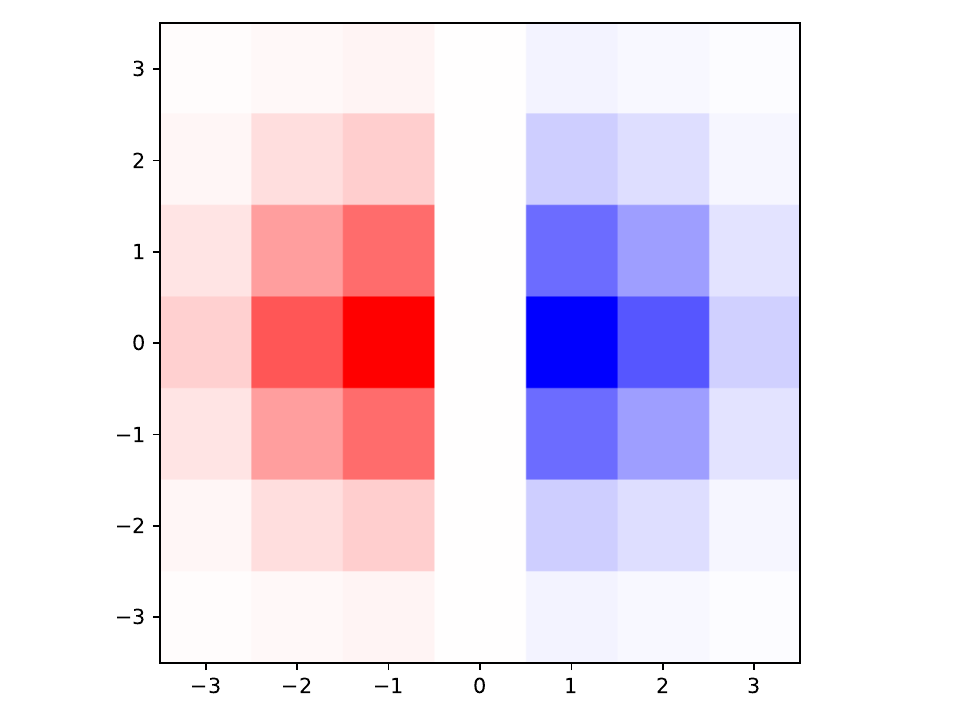}
        & \includegraphics[width=0.13\textwidth]{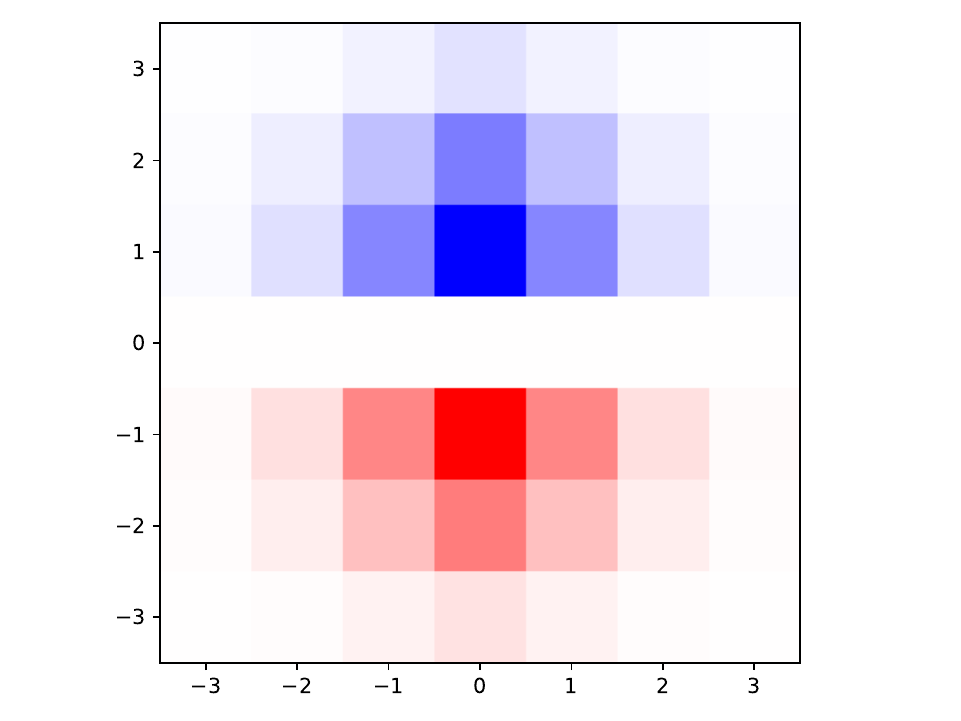}
        & \includegraphics[width=0.13\textwidth]{filter-jointscalefroml2diff7-bluered-eps-converted-to.pdf}
        & \includegraphics[width=0.13\textwidth]{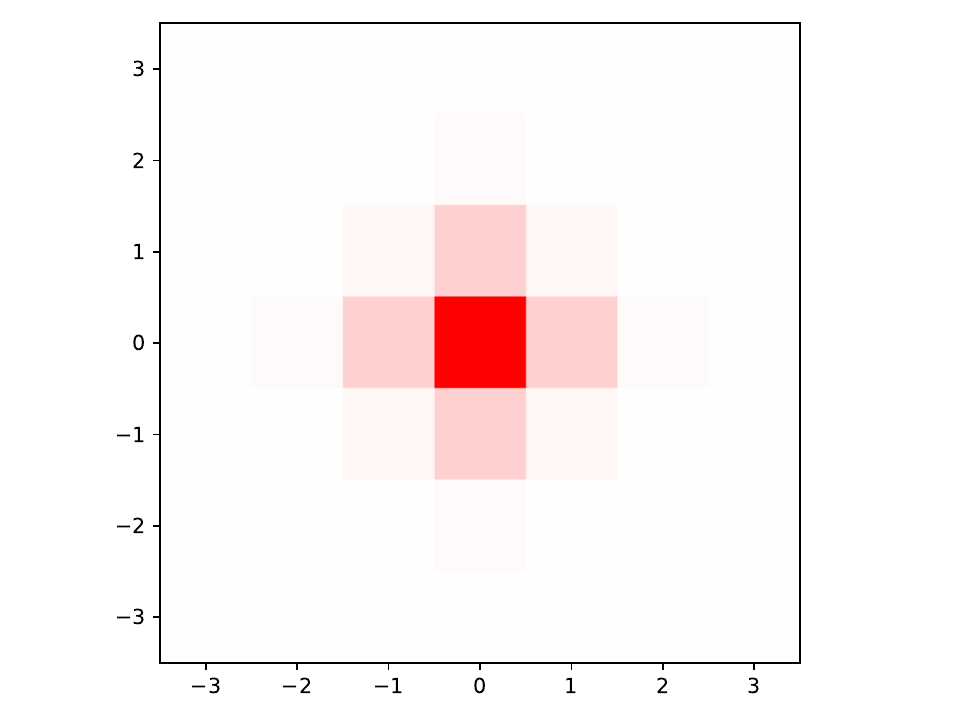} 
    \end{tabular}
  \end{center}
  \caption{Visualizations of the learned filters with the
    corresponding results of fitting idealized models of these filters
    using scale-space operations:
    {\bf (top row)} Alternative visualization of the original set of 8 ``master
    key filters'', extracted by Babaiee {\em et al.\/}
    (\citeyear{BabKiaRusGro25-AAAI-master}), while here complemented
    with a normalization the filters, (i)~by rescaling Filters~1--6 to
    give the same response to matching first-order discrete monomials under
    discrete convolution
    as for convolution of the corresponding continuous monomials with corresponding
    continuous Gaussian derivative operators,
    according to (\ref{eq-def-h1-norm})--(\ref{eq-def-h6-norm}), and
    (ii)~by adding different constants to Filters~7--8, according to
    (\ref{eq-def-h7-dc})--(\ref{eq-def-h8-norm}),
    to minimize the variance-based spatial spread measure
    of the filters, and finally visualizing the data on a blue-red colour scale with the
    value 0 corresponding to white, to better reveal the 
    polarities of the filter values.
    {\bf (rows 2-7)} Idealized scale-space models of the filters, as
    computed with the different types of modelling approaches proposed
    in this paper:
    {\bf Method~A\/} in
    Section~\ref{sec-method-A}, based on direct transfer of scale
    values from the variances for a continuous Gaussian derivative
    model,
    {\bf Method~B\/} in
    Section~\ref{sec-method-B}, based on requiring
    the horizontal {\em vs.\/}\ the vertical discrete weighted
    variance-based spatial spread measures for the idealized
    receptive field models to be equal to the
    weighted discrete spatial spread measures for 
    the corresponding learned filters,
    {\bf Method~C1\/} in
    Section~\ref{sec-method-C1}, based on minimizing the
    discrete $l_1$-norm between the idealized receptive field models
    and the normalized versions of the learned filters and using
    different values of the scale parameters in the horizontal and the
    vertical directions,
    {\bf Method~C2\/} in
    Section~\ref{sec-method-C2}, based on minimizing the
    discrete $l_1$-norm between the idealized receptive field models
    and the normalized versions of the learned filters
    and using 
    the same values of the scale parameters in the horizontal and the
    vertical directions,
    {\bf Method~D1\/} in
    Section~\ref{sec-method-D1}, based on minimizing the
    discrete $l_2$-norm between the idealized receptive field models
    and the normalized versions of the learned filters and using
    different values of the scale parameters in the horizontal and the
    vertical directions, and
    {\bf Method~D2\/} in
    Section~\ref{sec-method-D2}, based on minimizing the
    discrete $l_2$-norm between the idealized receptive field models
    and the normalized versions of the learned filters and using the
    same values of the scale parameters in the horizontal and the
    vertical directions.
    (Note that the contrasts of Filters~2 and~3 are reversed in relation to
    the sign conventions for the corresponding ``master key filters''.)
    (Horizontal axes: horizontal filter indices $m \in [-3, 3]$.
    Vertical axes: vertical filter indices $n \in [-3, 3]$.)}
  \label{fig-8-filters-composed}
\end{figure*}

More recently, Babaiee {\em et al.\/} (\citeyear{BabKiaRusGro24-ICLR},
\citeyear{BabKiaRusGro24-NeurIPSWS-fewer},
\citeyear{BabKiaRusGro25-AAAI-master})
have investigated the properties of
depthwise convolutional filters learned across all the layers of depthwise-separable
convolutional neural networks, as originally proposed by
Chollet (\citeyear{Cho17-CVPR}) and
Howard {\em et al.\/} (\citeyear{HowZhuCheKalWanWeyAndAda17-arXiv}),
with particular focus on the ConvNeXt architectures by Liu {\em et al.\/}
(\citeyear{LiuMaoWuFeiDarXie22-CVPR}) and
Woo {\em et al.\/} (\citeyear{WooDebHuCheLiuKweXi23-CVPR}).
Through unsupervised clustering techniques applied to
millions of trained filters, they discovered that the depthwise convolutional kernels
can be categorized into a few distinct classes, resembling Gaussian functions and their
derivatives. Remarkably, these filter patterns demonstrate generality
regardless of the layer depth, the network architecture, or the
training data domain. Through systematic greedy
search methods, they further demonstrated that this diverse set of learned filters can,
without major loss of performance, be reduced to a smaller set of just 8~so-called
``master key filters'',
see Figure~\ref{fig-8-extract-filters}
and the first row in Figure~\ref{fig-8-filters-composed} 
for two different ways of
visualizing the same data.%
\footnote{Concerning the visualizations of these filters, let us
  remark that that the actual scaling or the base levels of these filters
  is not regarded as essential for the functionality of the resulting
  deep networks in this study, since (i)~the scaling of the filters can both be propagated
  across the layers in the network hierarchy with maintained essential
  functionality regarding the computations
  and can also be compensated for by rescalings of the
  weights that combine the output from the different filters, and
  (ii)~the base levels of the filters can be compensated for with the bias
  terms in the linear combination of filter outputs before the
  non-linearity in each computational unit.
  Instead, for our approach, the spatial variations of the filters are
  regarded as providing the essence of the underlying computations,
  since these filter shapes determine how the local image information is to be
  combined between different spatial points in the image domain.}

Notably, these receptive fields do specifically appear to be qualitatively very
similar to receptive fields obtained from discrete scale-space theory,
in terms of either (see rows 2--7 in
Figure~\ref{fig-8-filters-composed}  
for illustrations):
\begin{itemize}
\item
  anti-symmetric first-order derivative approximations at moderately fine scales (Filters~5 and~6),
\item
  a Gaussian blob at a moderately fine scale (Filter~8),
\item
  a local sharpening operation by subtracting a
  Laplacian-of-the-Gaussian
  at a moderately fine scale (Filter~7), or
\item
  non-centered asymmetric first-order derivative approximations at very fine scales (Filters~1--4).
\end{itemize}
The subject of this paper is to present a quantitative analysis of these learned filters, and specifically to make comparisons to idealized simplifications with different degrees of simplification in idealized models for spatial receptive fields, based on an earlier presented theory for discrete derivative approximations with scale-space properties in Lindeberg (\citeyear{Lin93-Dis}, \citeyear{Lin93-JMIV}, \citeyear{Lin24-JMIV}, \citeyear{Lin25-FrontSignProc}).

In summary, the results to be presented will show that the learned
filters in depthwise-separable deep networks based on the ConvNeXt
architecture can be well modelled by discrete scale-space filters.
By quantitative results regarding deep learning, we also show that
replacement of the learned filter by idealized scale-space filters in
depthwise-separable deep networks also lead to almost as good results as
when using the fully learned filters. In these respects, the results
presented in this paper generalize the previous solid theoretical
foundation of using Gaussian derivative filters as the first layer of
linear filters in a vision system into a combined theoretical and
experimentally based foundation to also using (here discrete
approximations of) Gaussian derivative operators in the higher layers
of deep networks.

Given that the ConvNeXt architecture has been
demonstrated to constitute a modern state-of-the-art architecture for deep networks
with very good ability to compete with Transformer architectures,
a main significance of this work is that it establishes a strong link between
the filters learned in a reduced ConvNeXt architecture and the
idealized filters predicted from theoretically based scale-space theory.

\subsection{Contributions and novelty}

In brief, this paper comprises the following theoretical, conceptual
and experimental contributions:
\begin{itemize}
\item
  Extensions of the discrete scale-space theory to using non-centered
  discrete filter models.
\item
  A methodology for characterizing the properties of both idealized
  receptive field models and learned filters based on scale-space theory.
\item
  Methods for estimating the scale parameters in idealized receptive
  field models from learned filters based on matching of either
  continuous or discrete mean-based and variance-based spatial spread measures.
\item
  The formulation of weighted spatial spread measures, to reduce the
  risk that contributions from spurious values in the background may
  either lead to overestimates (biases) in the scale parameters.
\item
  The set of 8 idealized receptive field models obtained by model
  fitting for the 8 ``master key filters''.
\item
  The proposal of replacing the learned filters in the Conv\-NeXt V2
  architecture with idealized receptive fields based on scale-space
  theory, and evaluating the accuracy obtained when using different types of
  model fitting for determining the scale parameters in the idealized
  models of the receptive fields.
\item
  The result that model fitting based on discrete weighted
  spatial spread measures leads to significantly better predictive
  properties regarding the performance of using idealized receptive
  fields in a deep network architecture, compared to either using
  weighted continuous spatial spread measures or more traditional
  $l_1$- or $l_2$-based model fitting.
\item
  The result that using idealized receptive fields
  obtained from scale-space theory leads to almost as good accuracy as
  when using the originally learned filters in the ConvNeXt V2 Tiny
  architecture.
\item
  The result that only a marginal increase in the accuracy
  is obtained by replacing the values of the scale parameters
  obtained applying the best Method~B to the 8 ``master key filters''
  with the values of the scale parameters obtained by data-driven learning of the
  scale parameters using backpropagation.
\item
  A theoretically based interpretation of the spaces spanned by different sets of
  discrete derivative approximation operators applied to discrete Gaussian
  smoothing filters.
\item
  A set of predictions regarding the choices of the sets of filters to be used in
  further work regarding deep networks based on Gaussian derivatives
  or other closely related scale-space filters.
\end{itemize}
In these ways, the results provide additional support for using
idealized scale-space filters as the receptive field primitives in deep
networks, thereby extending (i)~the previous axiomatic {\em necessity\/} results
of using scale-space filters in the first layer in the visual
hierarchy, as well as (ii)~the previous {\em sufficiency\/} results of using
Gaussian derivative kernels in higher layers of deep networks,
towards (iii)~showing that the receptive fields learned in all the 
layers of a modern state-of-the-art architecture for deep learning can
be {\em very well approximated\/} by idealized scale-space filters.

\subsection{Structure of this paper}

The presentation is organized as follows:
Section~\ref{sec-background} provides a conceptual background 
regarding the ConvNeXt architecture for deep learning and the
procedure for extracting the set of 8 ``master key filters'' from the
filters learned from this architecture.
Section~\ref{sec-char-props-master-filters} then starts by describing a
theoretical background for measuring characteristic properties of the
learned filters, based on spatial spread measures, defined from
spatial means and spatial variances of the
absolute values of the filter coefficients, as well as the responses to
lower-order polynomials of the learned filters.
Section~\ref{sec-model-8-master-key-filt} continues by performing modelling of
the 8 ``master key filters'' using four different types of main
methods in terms of:
\begin{itemize}
\item
  Method~A: Spatial spread measures as matched between discrete spatial spread
  measures of the learned filters and continuous spatial spread
  measures of continuous Gaussian derivatives.
\item
  Method~B: Spatial spread measures as matched between discrete spatial spread
  measures of the learned filters and discrete spatial spread
  measures of discrete scale-space kernels defined by applying small
  support difference operators to the discrete analogue of the
  Gaussian kernel.
\item
  Method~C: Minimization of the difference between the learned filters
  and idealized discrete scale-space filters in discrete $l_1$-norm.
\item
  Method~D: Minimization of the difference between the learned filters
  and idealized discrete scale-space filters in discrete $l_2$-norm.
\end{itemize}
Section~\ref{sec-interpret} then interprets the results obtained in
this way in
terms of the spaces spanned by the receptive field responses in terms
of $N$-jet representations based on Gaussian derivatives.
This section also describes implications and predictions from these
results regarding the construction of more general Gaussian
derivative networks, with the layers defined from linear combinations
of Gaussian derivative responses.

Section~\ref{sec-experiments} thereafter applies the fitted idealized
discrete scale-space filters as the filtering primitives in
depthwise-separable networks based on the ConvNeXt V2 Tiny
architecture applied to the ImageNet dataset, and shows that:
\begin{itemize}
\item
  the idealized filters based on Method~B have the best
  predictive properties out of the four main classes of filter modelling
  methods, and
\item
  using the idealized models of the
  8 ``master key filters'' as the computational primitives in
  depthwise-separable deep networks leads to almost as good results as
  if using the originally learned filters.
\end{itemize}
Finally, Section~\ref{sec-summary} concludes with a summary and
discussion of some of the main results.

\section{Background on depthwise-separable CNNs and the ConvNeXt architecture}
\label{sec-background}

\subsection{Convolutional neural networks and depthwise-separable convolutions}

Convolutional Neural Networks (CNNs) constitute a class of deep
learning architectures in which spatial filtering operations are
performed by learned convolutional kernels applied across multiple
layers (LeCun {\em et al.\/} \citeyear{LecBotBenHaf98-ProcIEEE}). In a standard convolutional layer, each output channel is obtained by convolving all input channels with a kernel tensor of size $k \times k \times C_{\text{in}}$, where $k$ denotes the spatial kernel size and $C_{\text{in}}$ the number of input channels. Thus, spatial filtering and cross-channel mixing are performed jointly.

Depthwise-separable convolutions decouple these two operations into a
spatial and a channel-mixing stage
(Chollet \citeyear{Cho17-CVPR},
Howard {\em et al.\/} \citeyear{HowZhuCheKalWanWeyAndAda17-arXiv}).
Specifically:
\begin{itemize}
    \item \textbf{Depthwise convolution:} A spatial convolution with a $k \times k$ kernel is applied independently to each input channel. If the input has $C$ channels, this stage uses $C$ spatial kernels and does not mix information across channels.
    \item \textbf{Pointwise convolution:} A so-called $1 \times 1$ convolution is subsequently applied to linearly combine information across channels.
\end{itemize}
If both the input and output have $C$ channels, a standard convolution
requires $k^2 \, C^2$ parameters, whereas a depthwise-separable
convolution requires
$k^2 \ C + C^2$ parameters. For moderate $k$ and large $C$, this leads to substantial computational savings.

From a structural viewpoint, the depthwise stage performs the entire spatial filtering operation. The subsequent pointwise convolution operates purely over the channel dimension.

\subsection{The ConvNeXt architecture}

ConvNeXt is a modern convolutional architecture designed to achieve
competitive performance with Transformer-based vision models, while
retaining a fully convolutional structure
(Liu {\em et al.\/} \citeyear{LiuMaoWuFeiDarXie22-CVPR}).
It can be interpreted as a modernization of classical residual
networks (He {\em et al.\/} \citeyear{HeZhaRenSun16-CVPR}),
incorporating architectural refinements inspired by Vision
Transformers
(Dosovitskiy {\em et al.\/} \citeyear{DosBeyKolWeiZhaUntDehMinHeiGelUszHou21-ICLR}).

A ConvNeXt block typically consists of:
\begin{enumerate}
    \item a depthwise convolution (often with kernel size $7 \times 7$),
    \item a layer normalization (LN%
\footnote{Instead of normalizing across the batch, like standard CNNs,
  the LayerNorm (LN) operation in the ConvNeXt V2 architecture
  normalizes across the channels
  for each individual pixel (or spatial location), helping to
  stabilize the training and mimicking the behavior of Vision Transformers.}),
    \item a pointwise ($1 \times 1$) convolution expanding the channel dimension,
    \item a nonlinearity ({\em e.g.\/}, GELU%
\footnote{The Gaussian Error Linear Unit (GELU) is a smoother non-monotonic replacement of the ReLU
  function,
  $\operatorname{GELU}(x) = \frac{x}{2}
  \left( 1 + \operatorname{erf}(\tfrac{x}{\sqrt{2}})\right)$,
  found to enhance the training in deep networks.}),
   \item
     a global response normalization (GRN%
\footnote{Introduced in ConvNeXt V2, the global response normalization (GRN)
  layer enhances the feature diversity, by normalizing the activations of
  the channels based on their global relative importance, thereby
  preventing any single "dead" or over-dominant channel from stalling the network.})
     layer, which enhances the 
     feature diversity, by encouraging competition between the channels
     (Woo {\em et al.\/} \citeyear{WooDebHuCheLiuKweXi23-CVPR}),
   \item a second pointwise convolution, projecting back to the
      original channel dimension, and
    \item a residual connection.
\end{enumerate}
Crucially, the depthwise convolution is the only operation within the
block that performs spatial filtering. All the subsequent operations
act pointwise across spatial locations. Therefore, the spatial
receptive field structure learned by Conv\-NeXt is entirely determined
by its depthwise kernels.

\begin{figure}[hbtp]
  \begin{center}
    \includegraphics[width=0.27\textwidth]{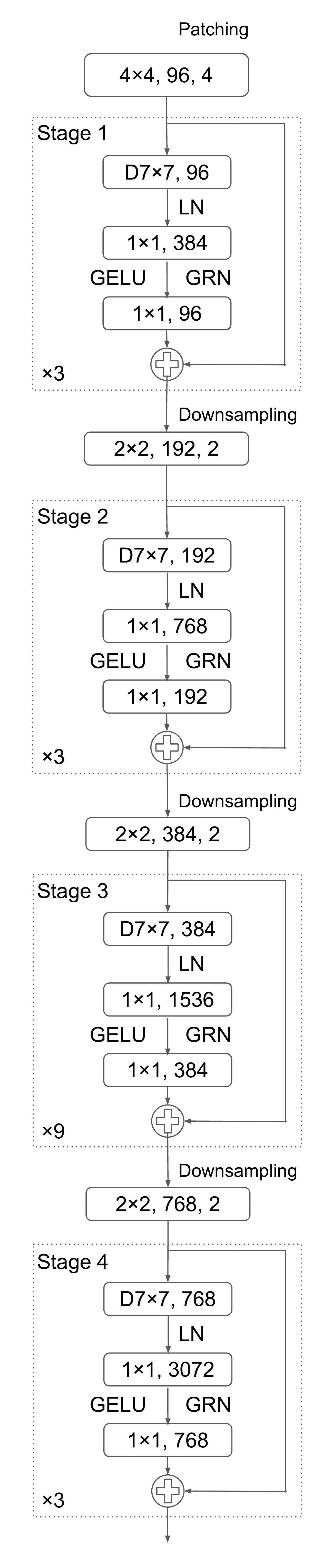}
  \end{center}
  \caption{Architectural overview of the ConvNeXt V2 Tiny network.
    Here, the abbreviation  ``LN'' denotes ``Layer Normalization'',
    while ``GRN'' denotes ``Global Response Normalization''.}
  \label{fig-convnext_v2}
\end{figure}

The global architecture of the ConvNeXt V2 network, that we will
build our work upon, is organized into four hierarchical stages
with a block distribution of $(3, 3, 9, 3)$ for the Tiny variant,
as illustrated in Figure~\ref{fig-convnext_v2}.  Starting from an initial
patching layer with $C=96$ channels, with convolutional kernels and a stride size of
$4$, the channel dimensions double at each stage ($96, 192, 384,
768$), using strided convolutions for downsampling.

\subsection{The Master Key Filters Hypothesis}

Research shows that the trained filters in
depthwise-separable CNNs consistently converge
into a few identifiable clusters across all the
layers (Babaiee {\em et al.\/} \citeyear{BabKiaRusGro24-ICLR}).
Through unsupervised clustering of
millions of filters, these patterns were found to resemble Gaussian
functions and derivatives of Gaussian kernels.
This discovery led to the Master Key Filters Hypothesis,
which proposes that a general set of universal filters exists for
visual data and that depthwise filters in depthwise-separable CNNs naturally tend
toward these "master keys" regardless of the specific dataset,
architecture, or layer depth
(Babaiee {\em et al.\/} \citeyear{BabKiaRusGro25-AAAI-master}).

To find the minimal set of the master key filters, an autoencoder was
used to compress the trained $7 \times 7$ depthwise kernels into a
single-dimensional hidden code
(Babaiee {\em et al.\/} \citeyear{BabKiaRusGro25-NeurIPS}). By sampling
points from this latent 1D space, a spectrum of candidate filters was
constructed. To find the minimal essential set, a greedy search was
performed on the filters learned in the ConvNeXt V2 Tiny
architecture, replacing its original filters with
the closest linear transformations of these sampled candidates. The
least important ones were removed iteratively, by measuring the change
in accuracy, when each candidate was removed from the list. This
process identified a distinct "elbow point" on the accuracy plot at
precisely 8 filters, where the accuracy remained stable up to that
point despite a massive reduction in unique weights. These 8
essential filters, shown in Figure~\ref{fig-8-extract-filters},
can effectively replace thousands of trained
parameters, while maintaining comparable performance on ImageNet.

\subsection{Tangible contributions to ConvNeXt networks}

In this work, we start from the given set of 8 ``master key filters'', extracted
using the procedure described above, and model
these filters in terms of idealized models of receptive fields, as
obtained from scale-space theory. By measuring and characterizing the
spatial extents of these filters in terms of spatial spread measures,
we show that the 8 ``master key filters'' can be well modelled by a set of
8 discrete scale-space filters.

By then replacing the originally
learned filters in the ConvNeXt V2 Tiny architecture with these
idealized models of spatial receptive fields, we show that it is
possible to obtain almost as good results on the ImageNet dataset as
if using the originally learned filters. This result thereby shows
that the filters in the ConvNeXt V2 Tiny architecture can
be very well approximated by discrete scale-space filters.

\section{Measuring characteristic properties of the 8 ``master key
  filters'' in terms of spatial spread measures and responses to
  lower-order polynomials}
\label{sec-char-props-master-filters}

In this section, we will apply a similar methodology, as used for modelling
and characterizing the discrete derivative approximation methods in
Lindeberg (\citeyear{Lin24-JMIV}, \citeyear{Lin25-FrontSignProc}),
to model the 8 ``master key filters'' extracted by
Babaiee {\em et al.\/} (\citeyear{BabKiaRusGro25-AAAI-master}),
while also employing structurally closely related methods for determining
the parameters of idealized receptive field models, as previously used
for matching the parameters between different types of theoretical
models for temporal receptive fields in 
Lindeberg (\citeyear{Lin23-BICY}, Sections~3.3--3.4).

\subsection{Spatial spread measures for characterizing the spatial extent and the spatial offset of receptive fields}

For measuring the overall effect of a non-negative receptive field $h(p) = h(x, y)$
with the image coordinates $p = (x, y)^T$ with
regard to its spatial extent, it is natural to formulate
a spatial spread measure in terms of the spatial mean $M(h(\cdot))$
and the spatial variance $M(h(\cdot, \cdot))$ according to
\begin{align}
  \begin{split}
    \label{eq-def-cont-mean}
    M(h(\cdot)) 
    & = \frac{\int_{p \in \bbbr^2} p \, h(p) \, dp}{\int_{p  \in \bbbr^2} h(p) \, dp},
  \end{split}\\
  \begin{split}
    \label{eq-def-cont-var}
    V(h(\cdot)) 
    & = \frac{\int_{p \in \bbbr^2} p \, p^T \, h(p) \, dp}{\int_{p \in \bbbr^2} h(p) \, dp}
           - \left(
                 \frac{\int_{p \in \bbbr^2} p \, h(p) \, dp}{\int_{p  \in \bbbr^2} h(p) \, dp}
              \right)^2,
  \end{split}
\end{align}
over a continuous image domain and according to
\begin{align}
  \begin{split}
    \label{eq-def-disc-mean}
    M(h(\cdot)) 
    & = \frac{\sum_{p \in \bbbr^2} p \, h(p)}{\sum_{p  \in \bbbr^2} h(p)},
  \end{split}\\
  \begin{split}
    \label{eq-def-disc-var}
    V(h(\cdot)) 
    & = \frac{\sum_{p \in \bbbr^2} p \, p^T \, h(p)}{\sum_{p \in \bbbr^2} h(p)}
           - \left(
                 \frac{\sum_{p \in \bbbr^2} p \, h(p)}{\sum_{p  \in \bbbr^2} h(p)}
              \right)^2,
  \end{split}
\end{align}
over a discrete image domain.

The use of these spatial spread measures
is structurally related to the use of mean values and covariance
matrices for characterizing gross overall properties of statistical distributions.
For our purpose, of capturing essential properties of both the learned
filters in a deep network architecture and the use of idealized
filter models to replace those filters, the motivation behind using
these measures is to in a compact manner characterize how both the
learned filters and the idealized filters weigh the image information from
different spatial points.

In previous work, the use of such spatial spread
measure has been demonstrated to have a predictive ability, to
compare the effective amount of spatial smoothing in receptive fields,
when performing learning of the scale levels in Gaussian
derivative networks using different types
of discretization operations for Gaussian derivative operators,
see Perzanowski and Lindeberg (\citeyear{PerLin25-JMIV}).

\subsubsection{Spatial spread measures applied to zero-order scale-space kernels}

For the continuous 2-D Gaussian kernel
\begin{equation}
  \label{eq-2D-gauss}
  g(x, y;\; \sigma)
  = g_{\oned}(x;\; \sigma) \, g_{\oned}(y;\; \sigma)
  = \frac{1}{2 \pi \sigma^2} \, e^{-(x^2 + y^2)/2\sigma^2},
\end{equation}
with the 1-D Gaussian kernel given by
\begin{equation}
  g_{\oned}(x;\; \sigma) = \frac{1}{\sqrt{2 \pi} \sigma} \, e^{-x^2/2\sigma^2}
\end{equation}
with standard deviation $\sigma$ and the spatial variance $s = \sigma^2$,
the resulting variance-based spread measure is then
\begin{equation}
  \label{eq-var-spread-cont-gauss}
  V(g(\cdot, \cdot;\; \sigma)) = \sigma^2 \, I,
\end{equation}
where $I$ denotes the unit matrix.
Similarly, for the 2-D discrete analogue of the Gaussian kernel
(Lindeberg \citeyear{Lin90-PAMI}, \citeyear{Lin24-JMIV})
\begin{equation}
  \label{eq-2D-disc-gauss}
  T(m, n;\; \sigma)
  = T_{\oned}(m;\; \sigma) \, T_{\oned}(n;\; \sigma) 
\end{equation}
with the 1-D discrete analogue of the Gaussian kernel given by
\begin{equation}
    \label{eq-1D-disc-gauss}
  T_{\oned}(n;\; \sigma) = e^{-s} I_n(\sigma^2),
\end{equation}
where $I_n(\sigma^2)$ denotes the modified Bessel functions of integer order
(see Abramowitz and Stegun \citeyear{AS64}),
the corresponding discrete spatial spread measure is correspondingly
(Lindeberg \citeyear{Lin93-Dis}, \citeyear{Lin24-JMIV})
\begin{equation}
  \label{eq-var-spread-disc-gauss}
  V(T(\cdot, \cdot;\; \sigma)) = \sigma^2 \, I.
\end{equation}
Since both the continuous Gaussian kernel and the discrete analogue of the Gaussian kernel are centered, we also have that their spatial mean values are
\begin{align}
  \begin{split}
    M(g(\cdot, \cdot;\; \sigma))
    = \left( \begin{array}{c} 0 \\ 0 \end{array} \right),
  \end{split}\\
  \begin{split}
    M(T(\cdot, \cdot;\; \sigma))
    = \left( \begin{array}{c} 0 \\ 0 \end{array} \right).
  \end{split}
\end{align}

\subsubsection{Spatial spread measures applied to first-order scale-space kernels}

Following the methodology in Lindeberg (\citeyear{Lin24-JMIV}, \citeyear{Lin25-FrontSignProc}),
we will in this paper also perform corresponding characterizations of receptive fields
$h(x, y)$ that are not guaranteed to be non-negative, by then measuring the mean value
and the variance of the absolute value of the receptive field $|h(x, y)|$,
in terms of the entities $M(|h(\cdot, \cdot)|)$ and $V(|h(\cdot, \cdot)|)$.

By computing these spatial spread measures for the horizontal first-order Gaussian derivative kernel
\begin{equation}
  g_x(x, y;\; \sigma)
  = \partial_x (g(x, y;\; \sigma))
  = - \frac{x}{\sigma^2} \, g(x, y;\; \sigma),
\end{equation}
we obtain
\begin{align}
  \begin{split}
    M(|g_x(\cdot, \cdot;\; \sigma)|) = \left( \begin{array}{cc} 0 \\ 0 \end{array} \right),
  \end{split}\\
  \begin{split}
    V(|g_x(\cdot, \cdot;\; \sigma)|)
    = \left(
          \begin{array}{cc}
            \label{eq-var-based-spread-meas-cont-x-gauss-der}
            2 \sigma^2 & 0 \\
            0 & \sigma^2
          \end{array}
        \right).
  \end{split}
\end{align}
Similarly, for the vertical first-order Gaussian derivative kernel
\begin{equation}
  g_y(x, y;\; \sigma) = \partial_y (g(x, y;\; \sigma)) = - \frac{y}{\sigma^2} \, g(x, y;\; \sigma),
\end{equation}
it holds that
\begin{align}
  \begin{split}
    M(|g_y(\cdot, \cdot;\; \sigma)|) = \left( \begin{array}{cc} 0 \\ 0 \end{array} \right),
  \end{split}\\
  \begin{split}
    \label{eq-var-based-spread-meas-cont-y-gauss-der}
    V(|g_y(\cdot, \cdot;\; \sigma)|)
    = \left(
          \begin{array}{cc}
            \sigma^2 & 0 \\
            0 & 2 \sigma^2
          \end{array}
        \right).
  \end{split}
\end{align}
Later, we will make use of these results in combination with
computed discrete variances of discrete kernels,
that resemble first-order derivative approximations,
for estimating the scale parameter $\sigma$ in corresponding
idealized models of receptive fields in terms of scale-space operations.

\begin{figure*}[hbtp]
  \begin{center}
    \begin{tabular}{cccc}
      $\delta_{x+} 
        = \left(
              \begin{array}{ccc}
                 0   & 0   & 0 \\
                 0  & -1 & +1 \\
                 0   & 0   & 0 
              \end{array}
            \right)$
     &
      $\delta_{x-} 
     = \left(
           \begin{array}{ccc}
              0   & 0   & 0 \\
             -1  & +1 & 0 \\
              0   & 0   & 0 
           \end{array}
         \right)$
     &
       $\delta_{y+} 
        = \left(
              \begin{array}{ccc}
                 0   & +1 & 0 \\
                 0   & -1 & 0 \\
                 0   &  0  & 0 
              \end{array}
            \right)$                         
      &
      $\delta_{y-} 
        = \left(
              \begin{array}{ccc}
                 0   &  0 & 0 \\
                 0   & +1 & 0 \\
                 0   &  -1 & 0 
              \end{array}
            \right)$                         
    \end{tabular}

    \bigskip
    
    \begin{tabular}{ccccc}
      $\delta_{x} 
        = \left(
              \begin{array}{ccc}
                 0     & 0  & 0 \\
                 -1/2  & 0 & +1/2 \\
                  0    &  0  & 0 
              \end{array}
           \right)$
        &
       $\delta_{y} 
        = \left(
              \begin{array}{ccc}
                 0  & +1/2  & 0 \\
                 0  &     0    & 0 \\
                 0  & -1/2  & 0 
              \end{array}
           \right)$
       &
       $\delta_{xx} 
        = \left(
              \begin{array}{ccc}
                 0     & 0  & 0 \\
                 +1  & -2 & +1 \\
                  0    &  0  & 0 
              \end{array}
           \right)$
       &
       $\delta_{xy} 
        = \left(
              \begin{array}{ccc}
                 -1/4  & 0  & +1/4 \\
                 0  & 0 & 0 \\
                 +1/4  &  0 & -1/4
              \end{array}
           \right)$
       &
       $\delta_{yy} 
        = \left(
              \begin{array}{ccc}
                 0  & 1  & 0\\
                 0  & -2 & 0 \\
                 0  &  1 & 0
              \end{array}
           \right)$
    \end{tabular}
  \end{center}
  \caption{Computational molecules visualizing the computational functions of the different (top row) non-centered first-order difference operators and (bottom row) centered first-order and second-order difference operators considered in this paper.}
  \label{fig-comp-mol-diff-ops}
\end{figure*}

\subsubsection{Spatial spread measures applied to zero-order pure difference operators}

In this paper, we will in the modelling and the analysis stages
primarily consider the following non-centered
first-order difference operators
(see Figure~\ref{fig-comp-mol-diff-ops} for an illustration):
\begin{align}
  \begin{split}
    (\delta_{x+} h)(m, n) & = h(m+1, n) - h(m, n),
  \end{split}\\
  \begin{split}
    (\delta_{x-} h)(m, n) & = h(m, n) - h(m-1, n),
  \end{split}\\
  \begin{split}
    (\delta_{y+} h)(m, n) & = h(m, n+1) - h(m, n),
  \end{split}\\
  \begin{split}
    (\delta_{y-} h)(m, n) & = h(m, n) - h(m, n-1),
  \end{split}
\end{align}
and the following centered first-order difference operators:
\begin{align}
  \begin{split}
    (\delta_{x} h)(m, n) & = (h(m+1, n) - h(m-1, n))/2,
  \end{split}\\
  \begin{split}
    (\delta_{y} h)(m, n) & = (h(m, n+1) - h(m, n-1))/2.
  \end{split}
\end{align}
Additionally, in the following interpretations of the results
to be considered in Section~\ref{sec-interpret},
we will also consider the following centered second-order difference operators:
\begin{align}
  \begin{split}
    (\delta_{xx} h)(m, n) & = h(m+1, n) - 2 h(m, n) + h(m-1, n),
  \end{split}\\
  \begin{split}
    (\delta_{xy} h)(m, n) & = (h(m+1, n+1) - h(m+1, n-1)
  \end{split}\nonumber\\
  \begin{split}
     & - h(m-1, n+1) + h(m-1, n-1))/4,
  \end{split}\\
  \begin{split}
    (\delta_{yy} h)(m, n) & = h(m, n+1) - 2 h(m, n) + h(m, n-1).
  \end{split}
\end{align}
For the non-centered first-order difference operators,
we specifically have the following concerning their spatial mean values:
\begin{align}
  \begin{split}
    \label{eq-mean-deltax+}
    M(|\delta_{x+}|) = \left( \begin{array}{cc} +\tfrac{1}{2} \\ 0 \end{array} \right),
  \end{split}\\
  \begin{split}
    \label{eq-mean-deltax-}
    M(|\delta_{x-}|) = \left( \begin{array}{cc} -\tfrac{1}{2} \\ 0 \end{array} \right),
  \end{split}\\
  \begin{split}
    \label{eq-mean-deltay+}
    M(|\delta_{y+}|) = \left( \begin{array}{cc} 0 \\ +\tfrac{1}{2}  \end{array} \right),
  \end{split}\\
  \begin{split}
    \label{eq-mean-deltay-}
    M(|\delta_{y-}|) = \left( \begin{array}{cc} 0 \\ -\tfrac{1}{2} \end{array} \right),
  \end{split}
\end{align}
while for the centered first-order or second-order difference operators, $\delta_{*} \in \{ \delta_{x}, \delta_{y}, \delta_{xx}, \delta_{xy}, \delta_{yy} \}$,
because of their pure mirror symmetries or anti-symmetries,
their spatial mean values are all zero:
\begin{equation}
    M(|\delta_{*}|) = \left( \begin{array}{cc} 0 \\ 0  \end{array} \right).
\end{equation}
For the non-centered first-order difference operators,
we specifically have the following concerning their spatial variances:
\begin{align}
  \begin{split}
    V(|\delta_{x+}|) = V(|\delta_{x-}|) = \left( \begin{array}{cc} \tfrac{1}{4} & 0 \\ 0 & 0 \end{array} \right),
  \end{split}\\
  \begin{split}
    V(|\delta_{y+}|) = V(|\delta_{y-}|) = \left( \begin{array}{cc} 0 & 0 \\ 0 & \tfrac{1}{4}  \end{array} \right),
  \end{split}
\end{align}
the following concerning the centered first-order difference operators:
\begin{align}
  \begin{split}
    V(|\delta_{x}|) = \left( \begin{array}{cc} 1 & 0 \\ 0 & 0 \end{array} \right),
  \end{split}\\
  \begin{split}
    V(|\delta_{y}|) = \left( \begin{array}{cc} 0 & 0 \\ 0 & 1 \end{array} \right),
  \end{split}
\end{align}
and the following concerning the centered second-order difference operators:
\begin{align}
  \begin{split}
    V(|\delta_{xx}|) = \left( \begin{array}{cc} \tfrac{1}{2} & 0 \\ 0 & 0 \end{array} \right),
  \end{split}\\
  \begin{split}
    V(|\delta_{xy}|) = \left( \begin{array}{cc} 1 & 0 \\ 0 & 1 \end{array} \right),
  \end{split}\\
  \begin{split}
    V(|\delta_{yy}|) = \left( \begin{array}{cc} 0 & 0 \\ 0 & \tfrac{1}{2} \end{array} \right).
  \end{split}
\end{align}

\begin{table}[hbtp]
  \begin{center}
  \begin{tabular}{cccc}
    \hline
    Filter & $\| h_i(\cdot, \cdot) \|_1$ & $M(|h_i(\cdot, \cdot)|)$ & $V(|h_i(\cdot, \cdot)|)$ \\
    \hline
    1 & 2.47 
       & $\left( \begin{array}{r} -0.023 \\ -0.602 \end{array} \right)$
       & $\left( \begin{array}{rr} +1.799 & -0.022 \\ -0.022 & +1.632 \end{array} \right)$ \\
    2 & 2.56
       & $\left( \begin{array}{r} +0.559 \\ +0.092 \end{array} \right)$
       & $\left( \begin{array}{rr} +1.328 & +0.012 \\ +0.012 & +1.425 \end{array} \right)$ \\
    3 & 2.73
       & $\left( \begin{array}{r} +0.056 \\ +0.693 \end{array} \right)$
       & $\left( \begin{array}{rr} +1.584 & +0.035 \\ +0.035 & +1.364 \end{array} \right)$ \\
    4 & 2.32
       & $\left( \begin{array}{r} -0.579 \\ +0.098 \end{array} \right)$
       & $\left( \begin{array}{rr} +1.635 & +0.051 \\ +0.051 & +1.253 \end{array} \right)$ \\ 
    5 & 4.44
       & $\left( \begin{array}{r} +0.057 \\ -0.041 \end{array} \right)$
       & $\left( \begin{array}{rr} +3.847 & -0.011 \\ -0.011 & +1.848 \end{array} \right)$ \\
    6 & 3.81
       & $\left( \begin{array}{r} -0.006 \\ +0.086 \end{array} \right)$
       & $\left( \begin{array}{rr} +1.634 & -0.080 \\ -0.080 & +3.347 \end{array} \right)$ \\
    7 & 1.96
       & $\left( \begin{array}{r} +0.043 \\ -0.086 \end{array} \right)$
       & $\left( \begin{array}{rr} +1.276 &  +0.003 \\ +0.003 & +1.547 \end{array} \right)$ \\
    8 & 3.07
       & $\left( \begin{array}{r} -0.029 \\ -0.048 \end{array} \right)$
       & $\left( \begin{array}{rr} +2.381 & -0.020 \\ -0.020 & +2.356 \end{array} \right)$ \\
    \hline                 
  \end{tabular}
  \end{center}
  \caption{The result of computing the discrete $l_1$-norm $\| h(\cdot, \cdot) \|_1$
    as well as the discrete spatial mean value $M(|h(\cdot, \cdot)|)$ according to
    Equation~(\ref{eq-def-disc-mean}) and the discrete spatial
    variance
    $V(|h(\cdot, \cdot)|)$
    according to Equation~(\ref{eq-def-disc-var}) for the 8 ``master key filters''
    shown in Figure~\ref{fig-8-extract-filters}.}
  \label{tab-norm-mean-var-learned-filters}
\end{table}

\subsection{Measuring the spatial extent and the spatial offset of the 8 ``master key filters''}
\label{sec-spat-extent-offset-prel}

Table~\ref{tab-norm-mean-var-learned-filters} shows the result of computing the discrete $l_1$-norm
\begin{equation}
  \| h(\cdot, \cdot) \|_1 = \sum_{(m, n) \in \bbbz^2} |h(m, n)|
\end{equation}
as well as the discrete spatial spread measures $M(|h(\cdot, \cdot)|)$ and $V(|h(\cdot, \cdot)|)$
according to Equations~(\ref{eq-def-disc-mean}) and (\ref{eq-def-disc-var}) for the 8 ``master key filters''
extracted by Babaiee {\em et al.\/} (\citeyear{BabKiaRusGro25-AAAI-master}) and shown
in Figure~\ref{fig-8-extract-filters}.
As can be seen from these results:
\begin{itemize}
\item
  For all the Filters~1--8, the off-diagonal mixed elements in the
  variance-based spatial spread measures in terms of the
  covariance matrices $V(|h(\cdot, \cdot)|)$ are much smaller in magnitude than the diagonal
  elements. In this respect, the learned receptive field appear to be
  strongly aligned with the Cartesian coordinate directions.
\item
  For Filters~1--4, one component of the spatial mean value
  $M(|h(\cdot, \cdot)|)$ is much smaller in magnitude than the other
  component, which is rather close to either $+\tfrac{1}{2}$ or
  $-\tfrac{1}{2}$, thus reflecting a substantial deviation from
  symmetry. In this respect, there is a close structural similarity to
  the corresponding spatial mean values for the non-centered
  first-order difference operators in
  Equations~(\ref{eq-mean-deltax+})--(\ref{eq-mean-deltay-}).

  There is, however, also a notable difference in the sense that the mean values do not very closely approximate $\pm 0.5$, but rather $\pm 0.6$ or $\pm 0.7$.
\item
  For Filters~5--6 and Filters~7--8, the spatial mean values $M(|h(\cdot, \cdot)|)$
  are rather close to zero, in the sense that their magnitudes are all
  smaller than 0.1,
  thus indicating a strong degree of either mirror symmetry or antisymmetry.
\item
  For Filters~5--6, the ratio between the diagonal elements in the variance-based
  spatial spread measure $V(|h(\cdot, \cdot)|)$ is very close to either 2 or $\tfrac{1}{2}$,
  indicating
  a rather close similarity to corresponding continuous variance-based spatial spread
  measures for first-order Gaussian derivative kernels in
  Equations~(\ref{eq-var-based-spread-meas-cont-x-gauss-der}) and
  (\ref{eq-var-based-spread-meas-cont-y-gauss-der}).
  The similarity is, however, not fully perfect, thus reflecting a minor notable difference.
\item
  For Filters~1--4, the ratio between the diagonal elements in the variance-based
  spatial spread measure $V(|h(\cdot, \cdot)|)$ is, however,
  not in any way close to either 2 or $\tfrac{1}{2}$, indicating that a direct transfer of
  the corresponding expressions (\ref{eq-var-based-spread-meas-cont-x-gauss-der}) and
  (\ref{eq-var-based-spread-meas-cont-y-gauss-der}) for the continuous first-order Gaussian derivatives
  does not appear as straightforward.
\item
  For Filter~8, the diagonal components of the variance-based
  spatial spread measure $V(|h(\cdot, \cdot)|)$ are very similar, thus with a very good qualitative agreement
  with Equations~(\ref{eq-var-spread-cont-gauss}) and (\ref{eq-var-spread-disc-gauss}).
\item
  For Filter~7, the diagonal elements of the variance-based
  spatial spread measure $V(|h(\cdot, \cdot)|)$ are somewhat similar, although
  differing in a clearly notable respect.
\item
  The discrete $l_1$-norms of Filters~1--4, which later will be shown
  to be different instances of the same class of filters, are of a similar order of
  magnitude. The discrete $l_1$-norms of Filters~5--6 are also
  of a similar magnitude.
  This property indicates that the data-driven way of
  determining the 8 ``master key filters'' can be regarded as
  reasonably robust between the clustered filters from the same class.
\item
  The discrete $l_1$-norms of all the filters $\{ 1, 2, 3, 4, 5, 6, 7, 8 \}$
  are of a roughly similar order of magnitude, while differing between
  the different classes $\{ 1, 2, 3, 4 \}$, $\{ 5, 6 \}$, $\{ 7 \}$
  and $\{ 8 \}$.%
\footnote{While one could potentially consider normalizing the
  filters with respect to their $l_1$-norms, possibly considering
  the way that the $L_1$-norms of continuous Gaussian derivative
  kernels differ between different order of spatial differentiation,
  we will, however, not here normalize the filters with respect
  to their $l_1$-norms, and instead normalize the filters with respect
  to their responses to monomials, as will be later described in
  Section~\ref{sec-resp-filters-to-monoms}.}
\end{itemize}
Concerning the interpretation of, in particular, the variance-based spatial spread measures
for Filters~7-8, it should, however, be noted that the definitions of
these spatial spread measures assume a background magnitude value
equal to zero, while Filters~7--8 in the set of ``master key filters'' have
background magnitude levels that are significantly different from
zero. If we are to use the spatial spread measures for capturing the
qualitative shapes of both learned and idealized filters in this way,
a potential problem, however, is that the occurence of spurious
non-zero values for the learned filters at image positions, where
the their corresponding idealized filters are close to zero or very
small, can lead to biases in the spatial spread measures.
An explanation for this is that the contribution to the
spatial spread measure will be weighted by the distance from the
origin for the first-order spatial spread measure and by the
distance squared for the second-order spatial spread measure.

Thus, if we would match the parameters of the idealized filters to
the shapes of the learned filters by regular spatial spread measures
only, the resulting idealized filters could become spatially too large,
thereby leading to substantial bias caused by the spatial spread measures,
by substantially overestimating the spatial extent of the filters.
Thus, some modification is needed, to be able to apply the
variance-based spatial spread measures $V(|h(\cdot, \cdot)|)$ for
estimating the filter parameters in the idealized models to be used
for modelling Filters~7--8.
We will return to this issue in Section~\ref{sec-dc-comp}
and Section~\ref{sec-weighted-spread-meas}.

\subsection{Measuring the accuracy in relation to centered or non-centered difference operators combined with spatial smoothing}

A natural consistency requirement, when constructing a centered discrete
derivative approximation operator $\Delta_{x^{\alpha} y^{\beta}}$,
that is to approximate the derivative operator $\partial_{x^{\alpha} y^{\beta}}$
of a given order $(m, n) \in \bbbz_+^2$ over a 2-D spatial domain, is to require that
the result of applying the discrete derivative approximation filter
of order $(m, n)$ applied to a monomial of the same order leads to a
response that is spatially constant, with specifically
\begin{equation}
  \label{eq-consist-req-der-approx-same-order}
  \Delta_{x^{\alpha} y^{\beta}}(x^{\alpha} \, y^{\beta}) = \alpha! \, \beta!,
\end{equation}
where $\alpha! = \alpha \cdot (\alpha - 1) \dots 2 \cdot 1$ denotes the factorial of the horizontal
derivative order $\alpha$ and $\beta! = \beta \cdot (\beta - 1) \dots 2 \cdot 1$ denotes the factorial of
the vertical derivative order $\beta$.
Furthermore, for any integer monomial of order $(a, b) \in \bbbz_+^2$,
if either $a < \alpha$ or $b < \beta$ (or both), we should require that
\begin{equation}
  \label{eq-consist-req-der-approx-lower-order}
  \Delta_{x^{\alpha} y^{\beta}}(x^a \, y^b) = 0.
\end{equation}
These consistency requirements are similar to the results of applying corresponding
continuous pure derivative operations $\partial_{x^{\alpha} y^{\beta}}$ to corresponding monomials.

\subsubsection{Shift-adjusted consistency requirement for non-centered difference operators designed to approximate spatial derivative operators}

With regard to modelling the 8 ``master key filters'' in Figure~\ref{fig-8-extract-filters},
we did according to the analysis in connection with
Table~\ref{tab-norm-mean-var-learned-filters} note that the
8 ``master key filters''
  (i)~are not necessarily centered, and
  (ii)~do not decrease as rapidly with increasing distance from the center
  as the discrete derivative approximation filters considered in
  Lindeberg (\citeyear{Lin24-JMIV}, \citeyear{Lin25-FrontSignProc}).
To handle the off-centering of the filters,
we do for the purpose
of the forthcoming analysis in this paper 
generalize the criterion in Equation~(\ref{eq-consist-req-der-approx-same-order})
into the shift-adjusted consistency requirement
\begin{equation}
  \label{eq-consist-req-der-approx-same-order-shift-adj}
  \Delta_{x^{\alpha} y^{\beta}}((x - m_x)^{\alpha} \, (y - m_y)^{\beta}) = \alpha! \, \beta!.
\end{equation}
We also generalize the criterion in Equation~(\ref{eq-consist-req-der-approx-lower-order})
into the shift-adjusted consistency requirement 
\begin{equation}
  \label{eq-consist-req-der-approx-lower-order-shift-adj}
  \Delta_{x^{\alpha} y^{\beta}}((x - m_x)^a \, (y - m_y)^b) = 0,
\end{equation}
to hold for any $(a, b) \in \bbbz_+^2$,
if either $a < \alpha$ or $b < \beta$ (or both),
where $m_x$ and $m_y$ denote the components in the spatial mean vector of the
absolute value of the discrete filter
$h_{x^{\alpha} y^{\beta}}(m, n)$, that implements the computational function of the
discrete derivative approximation operator $\Delta_{x^{\alpha} y^{\beta}}$
according to
\begin{equation}
  M(|h_{x^{\alpha} y^{\beta}}(\cdot, \cdot)|)
  = \left( \begin{array}{cc} m_x \\ m_y \end{array} \right).
\end{equation}
In the following analysis to be performed, we will specifically evaluate
these criteria at the centers of the responses of the discrete
derivative approximation filters to different types of
polynomials of the form $(x - m_x)^a \, (y - m_y)^b$.

Specifically, these consistency requirements hold for all of the pure
non-central or central difference operators, $\delta_{x+}$, $\delta_{x-}$,
$\delta_{y+}$, $\delta_{y-}$, $\delta_x$, $\delta_y$,
$\delta_{xx}$, $\delta_{xy}$ and $\delta_{yy}$, that we consider in this work.

\subsubsection{Responses to monomials for continuous Gaussian derivative operators}

Due to the fact that the spatial scale-space representation of any image $f(x, y)$,
obtained by Gaussian smoothing
expressed in terms of a parameterization of the scale parameter in terms of the
variance $s = \sigma^2$ of the Gaussian kernel 
\begin{equation}
  L(\cdot, \cdot;\; s) = g(\cdot, \cdot;\; s) * f(\cdot, \cdot),
\end{equation}
satisfies the diffusion equation
\begin{equation}
  \partial_s L = \frac{1}{2} \, \nabla^2 L = \frac{1}{2} \, (L_{xx} + L_{yy}),
\end{equation}
it follows that the evolution over scale for monomial input $f_{ab}(x, y) = x^a \, y^b$
can be described in terms of diffusion polynomials, where we for the lowest
orders $a$ and $b$ have that
\begin{align}
  \begin{split}
    \label{eq-diff-pol-0}
    f_{00}(x, y;\; s) = 1,
   \end{split}\\
 \begin{split}
    f_{10}(x, y;\; s) = x,
  \end{split}\\
  \begin{split}
    f_{01}(x, y;\; s) = y,
   \end{split}\\
 \begin{split}
    f_{20}(x, y;\; s) = x^2 + s,
  \end{split}\\
  \begin{split}
    f_{11}(x, y;\; s) = x \, y
   \end{split}\\
  \begin{split}
    f_{02}(x, y;\; s) = y^2 + s.
  \end{split}
\end{align}
From this summary, we can thus read off that the Gaussian smoothing function
will leave the constant input function $f_{00}(x, y) = 1$ as well as the linear
input functions $f_{10}(x, y) = x$ and $f_{01}(x, y) = y$ unchanged.

This property will then imply that the responses of the zero-order Gaussian kernel
and its first order derivatives in the $x$- and $y$-directions are given by
\begin{align}
  \begin{split}
    g(\cdot, \cdot) * 1 & = 1,
  \end{split}\\
  \begin{split}
    g(\cdot, \cdot) * x & = x,
  \end{split}\\
  \begin{split}
    g(\cdot, \cdot) * y & = y,
  \end{split}\\
  \begin{split}
    g_x(\cdot, \cdot) * 1 & = 0,
  \end{split}\\
  \begin{split}
    g_x(\cdot, \cdot) * x & = 1,
  \end{split}\\
  \begin{split}
    g_x(\cdot, \cdot) * y & = 0,
  \end{split}\\
    \begin{split}
    g_y(\cdot, \cdot) * 1 & = 0,
  \end{split}\\
  \begin{split}
    g_y(\cdot, \cdot) * x & = 0,
  \end{split}\\
  \begin{split}
    g_y(\cdot, \cdot) * y & = 1.
  \end{split}
\end{align}
By superposition, this property will then also
apply to any shift-adjusted input polynomial of the form
$f_{ab}(x, y) = (x - m_x)^a\,  (y - m_y)^b$.

By further differentiating these relationsship, it follows that the
first-order non-centered Gaussian derivative response in the $x$-direction
to the monomial input $f_{10}(x, y) = x - m_x$ as well as
the first-order non-centered Gaussian derivative response in the $y$-direction
to the monomial input $f_{01}(x, y) = y - m_y$ 
will be equal to 1, if the centering for the
non-centered Gaussian derivative responses
is around the point $(m_x, m_y)^T$.

\begin{table}[hbtp]
  \begin{center}
    \begin{tabular}{cccc}
      \hline
      Filter & $h_i(\cdot, \cdot) * 1$
      & $h_i(\cdot, \cdot) * (x - m_x)$
      & $h_i(\cdot, \cdot) * (y - m_y)$ \\
      \hline
        1 & $0 \cdot 10^{-5}$   & -0.079 & +1.556 \\
        2 & $0 \cdot 10^{-5}$   & -1.380 & +0.396 \\
        3 & $0 \cdot 10^{-5}$   & +0.129 & -1.091 \\
        4 & $0 \cdot 10^{-5}$ & +0.839 & +0.029 \\
        5 & $0 \cdot 10^{-5}$   & +7.750 & -0.063 \\
        6 & $0 \cdot 10^{-5}$  & -0.160 & +6.137 \\
        7 & $0 \cdot 10^{-5}$  & +0.085 & -0.169 \\
        8 & $0 \cdot 10^{-5}$ & -0.072 & -0.167 \\
      \hline
      \end{tabular}
  \end{center}
  \caption{The result of computing the responses to the monomials $\theta_{00} = 1$,
    $\theta _{10} = x - m_x$ and $\theta_{01} = y - m_y$ for the 8 ``master key filters'' shown in
    Figure~\ref{fig-8-extract-filters}, where the spatial offset
    $(m_x, m_y)^T = M(|h(\cdot, \cdot)|)$
    represents the spatial mean of the absolute value of each filter.}
  \label{tab-monom-resp-key-filters}
\end{table}

\begin{table}[hbtp]
  \begin{center}
    \begin{tabular}{cccc}
      \hline
      Filter
      & $h_{i,\norm}(\cdot, \cdot) * (x - m_x)$
      & $h_{i,\norm}(\cdot, \cdot) * (y - m_y)$ \\
      \hline
        1 & -0.051 & +1.000 \\
        2 & +1.000 & -0.287 \\
        3 & -0.118 & +1.000 \\
        4 & +1.000 & +0.035 \\
        5 & +1.000 & -0.008 \\
        6 & -0.026 & +1.000 \\
      \hline
      \end{tabular}
  \end{center}
  \caption{The result of computing the responses to the monomials $\theta_{00} = 1$,
    $\theta _{10} = x - m_x$ and $\theta_{01} = y - m_y$ for
    {\em normalized versions\/} of the first 6 ``master key filters'' shown in
    Figure~\ref{fig-8-extract-filters}, where the spatial offset
    $(m_x, m_y)^T = M(|h(\cdot, \cdot)|)$
    represents the spatial mean of the absolute value of each filter.
    For this table, the amplitudes of the original first 6 ``master key filters'' have been rescaled,
    to make the response equal to 1 for the monomial of $\theta_{10} = x - m_x$
    or $\theta_{01} = y - m_y$ that best matches an idealized approximation of the filter
    as either a derivative in the $x$-direction or the $y$-direction.}
  \label{tab-monom-resp-key-filters-norm}
\end{table}

\subsubsection{Responses to monomials for the 8 ``master key filters'' operators}
\label{sec-resp-filters-to-monoms}

Table~\ref{tab-monom-resp-key-filters} shows the result of computing the
responses to the monomials $\theta_{00} = 1$, $\theta _{10} = x - m_x$
and $\theta_{01} = y - m_y$ for the 8 ``master key filters''
in Figure~\ref{fig-8-extract-filters}, with the
spatial offset $(m_x, m_y)^T$ determined from the spatial mean of the
absolute value of each filter according to $(m_x, m_y)^T = M(|h(\cdot, \cdot)|)$.
As can be seen from these results:
\begin{itemize}
\item
  The responses of the filters to the constant function $\theta_{00} = 1$ are
  very close to zero for all the filters, indicating that the filters in this
  respect can be regarded as DC-balanced.%
\footnote{In the greedy search procedure for extracting the ``master key
  filters'', a normalization of the filter coefficients was used to
  having their sum equal to zero. In the data format used for storing
  the ``master key filters'' to be used as input for the modelling
  step, the filter coefficients were additionally rounded off to
  6~decimals. Thereby, the responses of the filters to the monomial 1
  will be quantized in units of $1 \cdot 10^{-6}$.

  In Table~\ref{tab-monom-resp-key-filters}, we have rounded off
  all these values to $0 \cdot 10^{-5}$, implying consistency
  with the data-driven way that this analysis has been performed,
  without any use of such prior information, and
  the quantization errors and the floating-point errors caused by the
  finite precision in the
  storage of the ``master key filters'' and the computations on them.}
\item
  For Filters~2, 4 and 5, the largest response is obtained for the
  monomial $\theta _{10} = x - m_x$, indicating that these filters
  respond as strongest in a way rather closely related to first-order
  derivatives in the horizontal $x$-direction.
\item
  For Filters~1, 2 and 6,  the largest response is obtained for the
  monomial $\theta _{01} = y - m_y$, indicating that these filters
  respond as strongest in a way rather closely related to first-order
  derivatives in the vertical $y$-direction.
\item
  The Filters~1--6, which approximate first-order derivatives in
  either the horizontal $x$-direction or the vertical $y$-direction,
  are, however, not normalized to return derivative responses
  to the functions $\theta _{10} = x - m_x$ and $\theta_{01} = y - m_y$,
  to be equal to one in the respective cases of matching
  orders between the monomials and the derivative approximations.
\item
  For Filters~7 and 8, which have the qualitatively similar visual
  appearance as a discrete Gaussian kernel, or a discrete delta
  function with a subtracted Laplacian-of-the-Gaussian, the DC-balancing
  of the ``master key filters'' is, however, not fully consistent with
  discrete Gaussian models of those filters, since both the
  continuous integral of the continuous Gaussian kernel
  and the discrete sum of the filter coefficients in discrete approximations
  of Gaussian kernels should be equal to 1
\end{itemize}
Table~\ref{tab-monom-resp-key-filters-norm} shows the result of
an extension of this analysis, where we have for the first 6 ``master key filters''
normalized the amplitudes of the receptive fields, to
make the response equal to 1 for the monomial of $\theta_{10} = x - m_x$
or $\theta_{01} = y - m_y$ that best matches an idealized approximation of the filter
as either a derivative in the $x$-direction or the $y$-direction:
\begin{align}
  \begin{split}
    \label{eq-def-h1-norm}
    h_{1,\norm}(m, n) = \frac{h_{1}(m, n)}{h_{1}(\cdot, \cdot)  * (y - m_y) |_{(x, y)^T = (0, 0)^T}},
  \end{split}\\
  \begin{split}
    \label{eq-def-h2-norm}    
    h_{2,\norm}(m, n) = \frac{h_{2}(m, n)}{h_{2}(\cdot, \cdot)  * (x - m_x) |_{(x, y)^T = (0, 0)^T}},
  \end{split}\\
  \begin{split}
    \label{eq-def-h3-norm}    
    h_{3,\norm}(m, n) = \frac{h_{3}(m, n)}{h_{3}(\cdot, \cdot)  * (y - m_y) |_{(x, y)^T = (0, 0)^T}},
  \end{split}\\
  \begin{split}
    \label{eq-def-h4-norm}
    h_{4,\norm}(m, n) = \frac{h_{4}(m, n)}{h_{4}(\cdot, \cdot)  * (x - m_x) |_{(x, y)^T = (0, 0)^T}},
  \end{split}\\
  \begin{split}
    \label{eq-def-h5-norm}
    h_{5,\norm}(m, n) = \frac{h_{5}(m, n)}{h_{5}(\cdot, \cdot)  * (x - m_x) |_{(x, y)^T = (0, 0)^T}},
  \end{split}\\
  \begin{split}
    \label{eq-def-h6-norm}
    h_{6,\norm}(m, n) = \frac{h_{6}(m, n)}{h_{6}(\cdot, \cdot)  * (y - m_y) |_{(x, y)^T = (0, 0)^T}}.
  \end{split}
\end{align}
As can be seen from these results:
\begin{itemize}
\item
  The complementary response to a first-order monomial in the direction
  orthogonal to the preferred direction of the receptive field is comparably
  rather low (of the order of $5~\%$ or below) for Filters~1, 4, 5 and~6.

  In this respect, the responses to first-order monomials are rather well
  consistent as approximations to first-order derivative operators for
  Filters~1, 4, 5 and~6. 
\item
  For Filter~2, the response to a first-order monomial in the direction
  orthogonal to the preferred direction of the receptive field is
  notable (of the order of $12~\%$) for Filter~3, and very notable
  (of the order of $30~\%$) for Filter~2.

  In this respect, there are notable deviations from approximations to
  first-order derivative operators for Filter~3 and specifically for Filter~2.
\end{itemize}
To conclude, Filters~1--6 show qualitatively rather good or very good structural
similarities to first-order derivative approximation filters along the
Cartesian coordinate direction.

The matches to the corresponding fully idealized models in terms of
Gaussian derivatives or discrete approximations thereof is,
however, far from perfect. A possible source to deviations from
the idealized theory could 
also be because of from variations in the background periphery
of the learned filters, which, by a lack of symmetry in those spurious
variations, could contribute significantly to the
responses to the first-order monomials.

\subsection{Reducing the bias in the estimates of the spatial variance of the Gaussian-smoothing-like receptive fields by DC-compensation}
\label{sec-dc-comp}

As previously remarked, for Filters~7 and~8, the estimates of the spatial
extents of the filters obtained from direct application of the
variance-based
spatial spread measures will be significantly biased, because of the lack of
a determination of the DC-level of these filters to correspond to a
background level of about zero. Instead, as we saw from the experimental
results in Table~\ref{tab-monom-resp-key-filters}, the response of
these filters to the constant function $\theta_{00} = 1$ is very close to zero,
as opposed to being close to one, as it should be for a continuous Gaussian
filter or a discrete approximation thereof.

To address this issue, we will in the following DC-correct Filters~7 and~8
according to
\begin{align}
  \begin{split}
    \label{eq-def-h7-dc}
    h_{7,\dc}(m, n;\; C_7) = h_{7}(m, n) - C_7,
  \end{split}\\
  \begin{split}
    \label{eq-def-h8-dc}
    h_{8,\dc}(m, n;\; C_8) = h_{8}(m, n) - C_8,
  \end{split}
\end{align}
by determining the values of the DC-correction-constants $\hat{C}_7$ and $\hat{C}_8$
so as to minimize the determinant of the variance-based spatial spread measure
$V(|h(\cdot, \cdot)|)$ according to
\begin{equation}
  \label{eq-def-crit-determ-Chat}
  \hat{C_i} = \operatorname{argmin}_{C_i} \det V(|h_i(\cdot, \cdot) - C_i|),
\end{equation}
for $i \in \{ 7, 8 \}$, and then further normalize the resulting
DC-corrected filters as
\begin{align}
  \begin{split}
    \label{eq-def-h7-norm}
    h_{7,\norm}(m, n) = \frac{h_{7,\dc}(m, n;\;
      \hat{C}_7)}{h_{7,\dc}(\cdot, \cdot;\; \hat{C}_7) * 1 |_{(x, y)^T = (0, 0)^T}},
  \end{split}\\
  \begin{split}
    \label{eq-def-h8-norm}   
    h_{8,\norm}(m, n) = \frac{h_{8,\dc}(m, n;\; \hat{C}_8)}{
      h_{8,\dc}(\cdot, \cdot;\; \hat{C}_8) * 1 |_{(x, y)^T = (0, 0)^T}},
  \end{split}
\end{align}
so that we obtain the following desirable properties
\begin{align}
  \begin{split}
    h_{7,\norm}(m, n) \star 1 = 1,
  \end{split}\\
  \begin{split}
    h_{8,\norm}(m, n) \star 1 = 1,
  \end{split}
\end{align}
in analogy to corresponding relationships for continuous Gaussian kernels
and appropriate discrete approximations thereof.

\begin{figure}[hbtp]
  \begin{center}
    \begin{tabular}{c}
      $\sqrt{\det V(|h_7(\cdot, \cdot) - C_7|)}$ \\
      \includegraphics[width=0.40\textwidth]{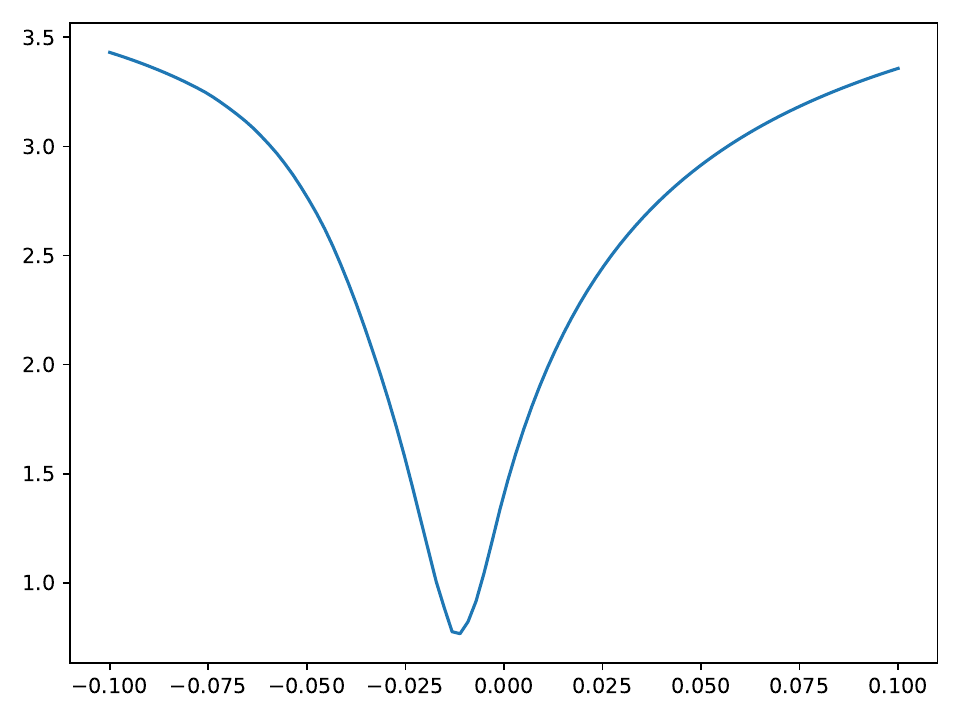} \\
      $\,$ \\
      $\sqrt{\det V(|h_8(\cdot, \cdot) - C_8|)}$ \\
      \includegraphics[width=0.40\textwidth]{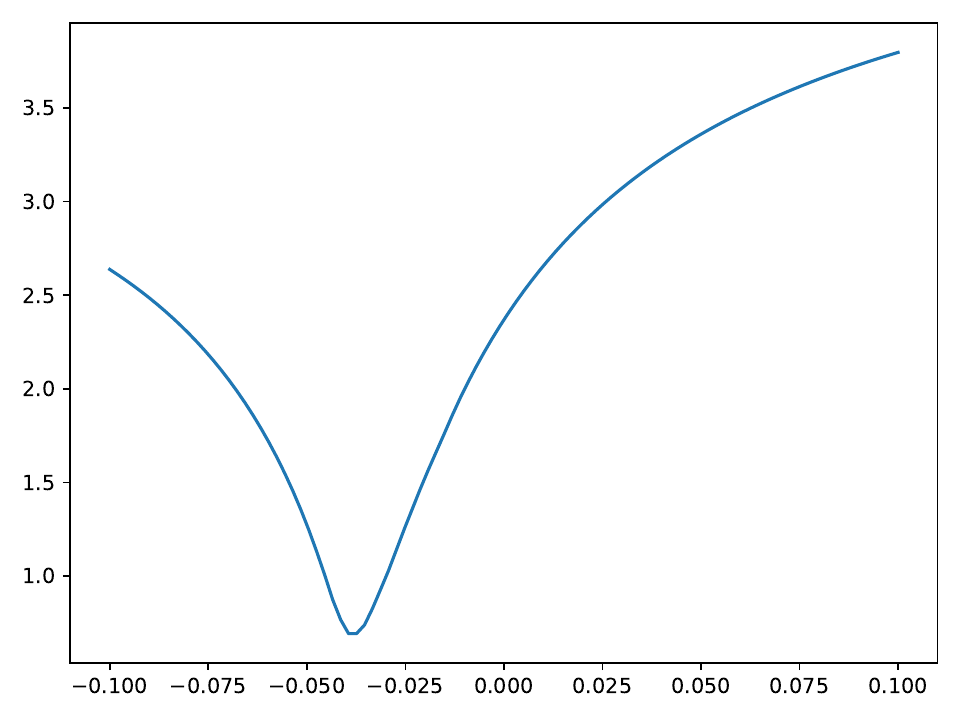}
    \end{tabular}
  \end{center}
  \caption{Graphs of the error measure $\sqrt{\det V(|h_i(\cdot, \cdot) - C_i|)}$
    for $ i \in \{ 7, 8 \}$ when determining the DC-compensation constants
    $C_7 \approx -0.0118$ and $C_8 \approx -0.0386$ according
    to (\ref{eq-def-crit-determ-Chat}), in order
    to later renormalize the Gaussian-like ``master key filters'' to unit $l_1$-norm,
    as regular Gaussian kernels obey.
    (Horizontal axes: parameter value $C_i \in [-0.1, 0.1]$.
    Vertical axes: error measure: $\sqrt{\det V(|h_i(\cdot, \cdot) - C_i|)}$.)}
  \label{fig-determ-C7-C8}
\end{figure}

\begin{table}[hbtp]
  \begin{center}
  \begin{tabular}{cccc}
    \hline
    Filter & $\| h_{i,\dc}(\cdot, \cdot) \|_1$ & $M(|h_{i,\dc}(\cdot, \cdot)|)$ & $V(|h_{i,\dc}(\cdot, \cdot)|)$ \\
    \hline
    7 & 2.686
       & $\left( \begin{array}{r} +0.050 \\ -0.021 \end{array} \right)$
        & $\left( \begin{array}{rr} +0.600 & -0.018 \\ -0.018 & +0.962 \end{array} \right)$ \\
    8 & 1.055
       & $\left( \begin{array}{r} +0.019 \\ +0.028 \end{array} \right)$
        & $\left( \begin{array}{rr} +0.696 & +0.028 \\ -0.018 & +0.680 \end{array} \right)$ \\
    \hline                 
  \end{tabular}
  \end{center}
  \caption{The result of computing the discrete $l_1$-norm $\| h(\cdot, \cdot) \|_1$
    as well as the discrete spatial mean value $M(|h(\cdot, \cdot)|)$ according to
    Equation~(\ref{eq-def-disc-mean}) and the discrete spatial
    variance
    $V(|h(\cdot, \cdot)|)$
    according to Equation~(\ref{eq-def-disc-var}) for the {\em DC-compensated\/}
    ``master key filters'' 7 and 8 according to (\ref{eq-def-h7-dc}) and (\ref{eq-def-h8-dc}).}
    \label{tab-norm-mean-var-dc-comp-filters}
\end{table}
    
\begin{table*}[hbtp]
  \begin{center}
  \begin{tabular}{cccc}
    \hline
    Filter
    & $h_{i,\norm}(\cdot, \cdot) \star 1$
    & $h_{i,\norm}(\cdot, \cdot) \star (x - m_x)$
    & $h_{i,\norm}(\cdot, \cdot) \star (y - m_y)$ \\
    \hline
    7 & 1.000 & -0.098 & -0.272 \\
    8 & 1.000 & -0.057 & -0.117 \\
    \hline
  \end{tabular}
  \end{center}
  \caption{The result of computing the responses to the monomials $\theta_{00} = 1$,
    $\theta _{10} = x - m_x$ and $\theta_{01} = y - m_y$ for
    {\em normalized versions\/} of the last 2 ``master key filters'' shown in
    Figure~\ref{fig-8-extract-filters}, where the spatial offset
    $(m_x, m_y)^T = M(|h(\cdot, \cdot)|)$
    represents the spatial mean of the absolute value of each filter.
    For this table, the amplitudes of the original last 2 ``master key filters'' have been rescaled,
    to make the response equal to 1 for the monomial of $\theta_{00} = 1$
    equal to 1.}
  \label{tab-monom-resp-key-filters-norm-78}
\end{table*}
  
Figure~\ref{fig-determ-C7-C8} show graphs of the error measure
\begin{equation}
  \sqrt{\det V(|h_i(\cdot, \cdot) - C_i|)}
\end{equation}
according to
(\ref{eq-def-crit-determ-Chat}) for determining
the DC-compensation constants for $\hat{C_7}$ and $\hat{C_8}$
in this way.
Table~\ref{tab-norm-mean-var-dc-comp-filters} shows the result of computing the spatial spread measures
for the resulting DC-compensated filters
$h_{7,\dc}(m, n)$ and $h_{8,\dc}(m, n)$
according to (\ref{eq-def-h7-dc}) and (\ref{eq-def-h8-dc}).
As can be seen from the results,
the use of DC compensation substantially reduces the variance-based spatial spread
measures for Filters~7 and~8, compared to the variance-based spread measures
of the corresponding not DC-compensated filters shown in Table~\ref{tab-norm-mean-var-learned-filters}.

Table~\ref{tab-monom-resp-key-filters-norm-78} additionally shows the responses to monomials
up to order 1 for the resulting DC-compensated and
$l_1$-normalized filters $h_{7,\norm}(m, n)$ and $h_{8,\norm}(m, n)$
according to (\ref{eq-def-h7-norm}) and (\ref{eq-def-h8-norm}).
As can be seen from the table, for the larger-size Gaussian-like
Filter~8, the responses to the first-order monomials $x$ and $y$
are rather low. For the smaller-size Gaussian-like
Filter~7, the deviations of those responses from the ideal
value 0 are, however, notable.

The top row in Figure~\ref{fig-8-filters-composed}
shows a visualization of
the resulting normalized versions of the 8 ``master key filters'',
which we will then approximate by idealized receptive field models in
different ways in Section~\ref{sec-model-8-master-key-filt}.
Before proceeding to the modelling stage, we will, however,
first in the next section introduce a way of defining spatial
spread measures, that are less influenced by spurious variabilities in
the peripheries of the learned filters.

\subsection{Reducing the bias in the estimates of the spatial variances, by windowed estimates of the spatial spread measures}
\label{sec-weighted-spread-meas}

As previously explained towards the end of
Section~\ref{sec-spat-extent-offset-prel},
that due to the spurious variations in the background of the learned
filters, there may be substantial biases in the
mean-based and variance-based spread measures, because
of these spurious variations being weighted by either a linear
or a quadratic factor in the first- and second-order moments
that define the spatial spread measures. This situation is, thus,
in contrast to previous use of such spatial spread measures
in Lindeberg (\citeyear{Lin24-JMIV}, \citeyear{Lin25-FrontSignProc})
for characterizing the properties of idealized mathematical
models of receptive fields with regard to variabilities in their
spatial extent under variations of the spatial scale parameter.

The underlying reason to this problem is that the idealized filter
models, that we are to replace the learned filters with, do all tend
to zero further away from the origin, whereas the learned filters
contain distinctly non-zero values for image positions where the
corresponding idealized filters with similar qualitative shape would
assume values very close to zero. Thus, without compensating for this
effect, a direct matching of idealized filter shapes to the learned
filters based on the regular spatial spread measures in 
Equations~(\ref{eq-def-cont-mean})--(\ref{eq-def-disc-var}) may overestimate the
spatial size of the idealized filters that are to replace the learned
filters.

The underlying reason for this is that the spurious variations
in the background are all transformed to their absolute values, when
computing the spatial spread measures. Specifically, regarding the
variance-based spatial spread measure, these absolute values are
weighted with the squared distance from the origin. Thereby, such
spurious variations may contribute in a different way to the spatial
spread measures than their effects in the learned filters, where
spurious variations of both positive and negative sign could be
expected to in some statistical sense counteract each other.

\subsubsection{Weighted spatial spread measures}

To reduce this potentially substantial source of bias, we will in this section
replace the previous uniformly weighted first- and second-order
moments by weighted first- and second-order moments.
For this purpose, we will use weight functions determined from discrete approximations
of Gaussian functions with similar values of the scale parameters,
as we would obtain by estimating the scale parameters of
idealized filter models from the spatial variances of the corresponding
discrete kernels.
Thus, we define weighted spatial mean values and spatial variance
of continuous or discrete filter kernels $h(\cdot, \cdot)$ according to:
\begin{equation}
  \label{eq-def-cont-mean-weigth}
  M_{g(\cdot)}(h(\cdot)) 
  = \frac{\int_{p \in \bbbr^2} p \, h(p) \, g(p) \, dp}{\int_{p  \in \bbbr^2} h(p) \, g(p) \, dp},
\end{equation}
\begin{multline}
  \label{eq-def-cont-var-weigth}
  V_{g(\cdot)}(h(\cdot)) 
  = \frac{\int_{p \in \bbbr^2} p \, p^T \, h(p) \, g(p) \, dp}{\int_{p \in \bbbr^2} h(p) \, g(p) \, dp} \\
      - \left(
           \frac{\int_{p \in \bbbr^2} p \, h(p) \, g(p) \, dp}{\int_{p  \in \bbbr^2} h(p) \, g(p) \, dp}
         \right)^2,
\end{multline}
over a continuous image domain and according to
\begin{equation}
  \label{eq-def-disc-mean-weigth}
  M_{T(\cdot)}(h(\cdot)) 
  = \frac{\sum_{p \in \bbbr^2} p \, h(p) \, T(p) \, dp}{\sum_{p  \in \bbbr^2} h(p) \, T(p) \, dp},
\end{equation}
\begin{multline}
  \label{eq-def-disc-var-weigth}
  V_{T(\cdot)}((h(\cdot)) 
  = \frac{\sum_{p \in \bbbr^2} p \, p^T \, h(p) \, T(p)}{\sum_{p \in \bbbr^2} h(p) \, T(p)} \\
      - \left(
            \frac{\sum_{p \in \bbbr^2} p \, h(p) \, T(p)}{\sum_{p  \in \bbbr^2} h(p) \, T(p)}
         \right)^2,
\end{multline}
over a discrete image domain, with the intention that peripheral values
corresponding to spurious structures in the learned filters should
have lower influence on the spatial spread measures.

Specifically, if the continuous receptive field $h(\cdot)$ would be equal to the
continuous weighting filter $g(\cdot)$, then the weighted spatial mean value
and the weighted spatial variance would correspond to the mean value
and the standard deviation in $L_2$-norm.
Similarly, if the discrete receptive field $h(\cdot)$ would be equal to the
discrete weighting filter $T(\cdot)$, then the weighted spatial mean value
and the weighted spatial variance would correspond to the mean value
and the standard deviation in $l_2$-norm.
For our purpose, we do on the other hand formulate a hybrid approach,
where we use the desired shape of the filter model as a prior
to suppress the influence of spurious responses in the learned
filter spatial positions in the peripheries of the support regions
of the learned filters.

\subsubsection{Determining idealized shapes for the weighting kernels from the qualitative shapes of the ``master key filters''}
\label{sec-weight-kernel-shapes-from-master-key-shapes}

The intention is then to choose either continuous or discrete weighting
filters that match the qualitative shapes of the filters to be modelled
in the following ways:
\begin{itemize}
\item
  For Filter~8, we will either use anisotropic
  continuous Cartesian-aligned anisotropic Gaussian kernels
  \begin{equation}
    g(x, y;\; \sigma_x, \sigma_y)
    = g_{\oned}(x;\; \sigma_x) \, g_{\oned}(y;\; \sigma_y)
  \end{equation}
  or discrete Cartesian-aligned anisotropic Gaussian kernels
  \begin{equation}
    \label{eq-2d-disc-gauss-x-y}
    T(m, n;\; \sigma_x, \sigma_y)
    = T_{\oned}(m;\; \sigma_x) \, T_{\oned}(n;\; \sigma_y)
  \end{equation}
  having possibly different
  values of the scale parameters $\sigma_x$ and $\sigma_y$
  in the $x$- and $y$-directions, respectively.
\item
  For Filters~2, 4 and 5, we will either use the absolute value of
  either a first-order horizontal $x$-derivative
  of a continuous Gaussian kernel
  \begin{equation}
    g_x(x, y;\; \sigma_x, \sigma_y)
    = g_{\oned,x}(x;\; \sigma_x) \, g_{\oned}(y;\; \sigma_y)
  \end{equation}
  or the absolute value of a discrete approximation of a horizontal $x$-derivative of a Gaussian kernel
  \begin{equation}
    (\Delta_x T)(m, n;\; \sigma_x, \sigma_y)
    = (\Delta_x T_{\oned})(m;\; \sigma_x) \, T_{\oned}(n;\; \sigma_y)
  \end{equation}
  for appropriately matched difference
  operators
  $\Delta_x \in \{ \delta_{x-}, \delta_{x+}, \delta_x \}$
  in the horizontal $x$-direction, with the scale parameters $\sigma_x$ and $\sigma_y$
  determined from variance-based spatial spread measures
  of the corresponding filters.
\item
  For Filters~1, 3 and 6, we will either use the absolute value of a first-order vertical $y$-derivative
  of continuous Gaussian kernel
  \begin{equation}
    g_y(x, y;\; \sigma_x, \sigma_y)
    = g_{\oned}(x;\; \sigma_x) \, g_{\oned,y}(y;\; \sigma_y)
  \end{equation}  
  or the absolute value of a discrete approximation of a vertical $y$-derivative of a Gaussian kernel
   \begin{equation}
    (\Delta_y T)(m, n;\; \sigma_x, \sigma_y)
    = T_{\oned}(m;\; \sigma_x) \, (\Delta_y T_{\oned})(n;\; \sigma_y)
  \end{equation}
  for appropriately matched difference operators
  $\Delta_y \in \{ \delta_{y+}, \delta_{y-}, \delta_y \}$
  in the vertical $y$-direction, with the scale parameters $\sigma_x$ and $\sigma_y$
  determined from variance-based spatial spread measures
  of the corresponding filters.
\item
  For Filter~7, we will use either a continuous isotropic%
\footnote{The reason for using an isotropic Laplacian operator here is
  mainly to obtain a minimum number of parameters in the model. More
  generally, one could also conceive replacing the isotropic
  Gaussian kernel in this model with a separable anisotropic Gaussian
  kernel, that makes
  use of different scale parameters in the horizontal and the vertical
  directions. Possibly, one could also consider extending such a model
  by making use of different scaling constants $\gamma_{x}$ and
  $\gamma_{y}$, that weigh the contributions from the terms
  $\partial_{xx}(g(x;\; \sigma_{x} g(y;\; \sigma_{y}))$ and
  $\partial_{yy}(g(x;\; \sigma_{x} g(y;\; \sigma_{y}))$,
  respectively,
  from the horizontal {\em vs.\/}\ the vertical directions.
  The discrete sharpening filter
  (\ref{eq-disc-sharp-filter}) could possibly also be extended to
  corresponding anisotropic image operations. We do,
  however, leave the possible investigations of such design
  options regarding the idealized filter models
  to future work.}
  sharpening filter of the form
  \begin{equation}
    T_{\text{sharp}}(x, y;\; \sigma, \gamma)
    = 1 - \gamma \, (\nabla^2 g)(x, y;\; \sigma, \sigma),
  \end{equation}
  for an appropriately determined scale parameter $\sigma$ with an
  appropriately determined scaling constant $\gamma$,
  where $\nabla^2 = \partial_{xx} + \partial_{yy}$ denotes the
  isotropic Laplacian operator,
  or a discrete isotropic sharpening filter of the form
  \begin{multline}
    \label{eq-disc-sharp-filter}
    T_{\text{sharp}}(x, y;\; \sigma, \gamma) = \\
    = 1 - \gamma_7 \,
    \nabla_5^2 (T_{\oned}(m;\; \sigma) \, T_{\oned}(n;\; \sigma)),
  \end{multline}
  where $\nabla_5^2 = \delta_{xx} + \delta_{yy}$ denotes the 5-point
  discrete approximation of the Laplacian operator.
\end{itemize}

\subsection{Estimating variance parameters of receptive fields from weighted spatial spread measures}

When using non-uniform weighting kernels for
computing the spatial spread measures,
the resulting variance-based measures will, however,
be affected by the spatial extent of the weighting kernels.
In the continuous case, we specifically have the
following relations:
\begin{align}
  \begin{split}
    \label{eq-weight-mean-cont-gauss}
    M_{g(\cdot, \cdot;\; \sigma_x, \sigma_y)}(g(\cdot, \cdot;\; \sigma_x, \sigma_y))
    = \left(
            \begin{array}{c}
               0 \\
               0
            \end{array}
      \right),
  \end{split}\\
  \begin{split}
    \label{eq-weight-var-cont-gauss}
    V_{g(\cdot, \cdot;\; \sigma_x, \sigma_y)}(g(\cdot, \cdot;\; \sigma_x, \sigma_y))
    = \left(
          \begin{array}{cc}
             \sigma_x^2/2 & 0 \\
              0 & \sigma_y^2/2 
          \end{array}
        \right),
    \end{split}\\
  \begin{split}
    \label{eq-weight-mean-cont-gauss-derx}
    M_{|g_x(\cdot, \cdot;\; \sigma_x, \sigma_y)|}(|g_x(\cdot, \cdot;\; \sigma_x, \sigma_y)|)
    = \left(
            \begin{array}{c}
               0 \\
               0
            \end{array}
      \right),
  \end{split}\\
  \begin{split}
    \label{eq-weight-var-cont-gauss-derx}
    V_{|g_x(\cdot, \cdot;\; \sigma_x, \sigma_y)|}(|g_x(\cdot, \cdot;\; \sigma_x, \sigma_y)|)
    = \left(
          \begin{array}{cc}
             3 \, \sigma_x^2/2 & 0 \\
              0 & \sigma_y^2/2 
          \end{array}
        \right),
  \end{split}\\
  \begin{split}
    \label{eq-weight-mean-cont-gauss-dery}
    M_{|g_y(\cdot, \cdot;\; \sigma_x, \sigma_y)|}(|g_y(\cdot, \cdot;\; \sigma_x, \sigma_y)|)
    = \left(
            \begin{array}{c}
               0 \\
               0
            \end{array}
      \right),
  \end{split}\\
  \begin{split}
    \label{eq-weight-var-cont-gauss-dery}
    V_{|g_y(\cdot, \cdot;\; \sigma_x, \sigma_y)|}(|g_y(\cdot, \cdot;\; \sigma_x, \sigma_y)|)
    = \left(
          \begin{array}{cc}
             \sigma_x^2/2 & 0 \\
              0 & 3 \, \sigma_y^2/2 
          \end{array}
        \right).
    \end{split}   
\end{align}
Our intention is then to use these expressions for the weighted
variance-based spatial spread measures for estimating the scale
parameters $\sigma_x$ and $\sigma_y$ in idealized models of the
learned ``master key filters'', based on weighted variance measurements
for the learned filters.

\begin{table}[hbtp]
  \begin{center}
  \begin{tabular}{ccc}
    \hline
    Filter & $M_{T(\cdot, \cdot)}(|h_i(\cdot, \cdot)|)$ & $V_{T(\cdot, \cdot)}(|h_i(\cdot, \cdot)|)$ \\
    \hline
    1 & $\left( \begin{array}{r} -0.005 \\ -0.753 \end{array} \right)$
       & $\left( \begin{array}{rr} +0.169 & -0.010 \\ -0.010 & +0.466 \end{array} \right)$ \\
    2 & $\left( \begin{array}{r} +0.665 \\ -0.017 \end{array} \right)$
       & $\left( \begin{array}{rr} +0.469 & +0.010 \\ +0.010 & +0.168 \end{array} \right)$ \\
    3 & $\left( \begin{array}{r} +0.011 \\ +0.666 \end{array} \right)$
       & $\left( \begin{array}{rr} +0.190 & +0.001 \\ +0.001 & +0.541 \end{array} \right)$ \\
    4 & $\left( \begin{array}{r} -0.536 \\ +0.005 \end{array} \right)$
       & $\left( \begin{array}{rr} +0.619 & +0.000 \\ +0.000 & +0.090 \end{array} \right)$ \\ 
    5 & $\left( \begin{array}{r} -0.011 \\ -0.020 \end{array} \right)$
       & $\left( \begin{array}{rr} +1.732 & -0.011 \\ -0.011 & +0.320 \end{array} \right)$ \\
    6 & $\left( \begin{array}{r} -0.006 \\ +0.114 \end{array} \right)$
       & $\left( \begin{array}{rr} +0.297 & -0.002 \\ -0.002 & +1.510 \end{array} \right)$ \\
    8 & $\left( \begin{array}{r} -0.002 \\ +0.008\end{array} \right)$
       & $\left( \begin{array}{rr} +0.152 & +0.001\\ +0.001 & +0.149 \end{array} \right)$ \\
    \hline                 
  \end{tabular}
  \end{center}
  \caption{The result of computing the discrete {\em weighted\/}
    spatial mean value
    $M_{T(\cdot, \cdot)}(|h(\cdot, \cdot)|)$ according to
    Equation~(\ref{eq-def-disc-mean-weigth}) and the
    discrete {\em weighted\/} spatial variance $V_{T(\cdot, \cdot)}(|h(\cdot, \cdot)|)$
    according to Equation~(\ref{eq-def-disc-var-weigth})
    for the normalized versions of the 8 ``master key filters''
    according to (\ref{eq-def-h1-norm})--(\ref{eq-def-h6-norm})
    and 
    (\ref{eq-def-h8-norm}) when using matching shapes of
    the spatial weighting filters $T(\cdot, \cdot)$ to the qualitative shapes of the
    ``master key filters'' according to
    Section~\ref{sec-weight-kernel-shapes-from-master-key-shapes}.}
  \label{tab-norm-weighted-mean-var-learned-filters}
\end{table}

Table~\ref{tab-norm-weighted-mean-var-learned-filters}
shows weighted spatial spread measures computed
in this way, with the weighting functions for the respective
normalized versions of the ``master key filters'' according
to (\ref{eq-def-h1-norm})--(\ref{eq-def-h6-norm})
and 
(\ref{eq-def-h8-norm})
determined from the absolute values of the corresponding
idealized filter models according to
(\ref{eq-def-h1-ideal})--(\ref{eq-def-h8-ideal}),
with the scale parameters of the weighting functions
set to  $\sigma_x = \sigma_y = \sigma_0 = 1$ in the spatial
weighting kernels for all the ``master key filters''.

As can be seen from a comparison with the corresponding
non-weighted spatial spread measures in
Tables~\ref{tab-norm-mean-var-learned-filters}
and~\ref{tab-norm-mean-var-dc-comp-filters},
the use of spatial weighting in the computation of
the variance-based spread measures leads to substantially
lower estimates of the corresponding scale parameters,
and thus to a substantial decrease of the previous bias,
caused by spurious variations in the peripheries
of the learned filters.

We can also see that the elements of the weighted mean value vectors are,
as for the previously computed non-weighted mean values,
either around $\pm 1/2$ or 0, although again with slight
offsets from the ideal values by up to 0.1 or 0.2 grid units.
In this respect, both the weighted and the unweighted mean value
give rise to qualitatively similar results, indicating that the
spurious variations in the backgrounds of the ``master key filters''
seem to have roughly symmetric statistical properties under
mirroring through the coordinate axes.

\section{Modelling the 8 ``master key filters'' in terms of idealized scale-space operations}
\label{sec-model-8-master-key-filt}

In the following, we will use the modelling and analysis results from
the previous section for modelling the 8 ``master key filters''
in terms of idealized scale-space operations.
Specifically, we will explore the implications of using different
types of modelling criteria and analyze how they give rise to
different sets of scale parameters in the idealized filter models.
Overall, we will either consider separate modelling of the
different filters of the following forms:
\begin{align}
  \begin{split}
    \label{eq-def-h1-ideal}
    h_{1,\ideal}(m, n;\; \sigma_{x,1}, \sigma_{y,1})
    = (\delta_{y+} T)(m, n;\; \sigma_{x,1}, \sigma_{y,1}),
  \end{split}\\
  \begin{split}
    \label{eq-def-h2-ideal}    
    h_{2,\ideal}(m, n;\; \sigma_{x,2}, \sigma_{y,2})
    = (\delta_{x-} T)(m, n;\; \sigma_{x,2}, \sigma_{y,2}),
  \end{split}\\
  \begin{split}
    \label{eq-def-h3-ideal}    
    h_{3,\ideal}(m, n;\; \sigma_{x,3}, \sigma_{y,3})
    = (\delta_{y-} T)(m, n;\; \sigma_{x,3}, \sigma_{y,3}),
  \end{split}\\
  \begin{split}
    \label{eq-def-h4-ideal}    
    h_{4,\ideal}(m, n;\; \sigma_{x,4}, \sigma_{y,4})
    = (\delta_{x+} T)(m, n;\; \sigma_{x,4}, \sigma_{y,4}),
  \end{split}\\
  \begin{split}
    \label{eq-def-h5-ideal}    
    h_{5,\ideal}(m, n;\; \sigma_{x,5}, \sigma_{y,5})
    = (\delta_{x} T)(m, n;\; \sigma_{x,5}, \sigma_{y,5}),
  \end{split}\\
  \begin{split}
    \label{eq-def-h6-ideal}    
    h_{6,\ideal}(m, n;\; \sigma_{x,6}, \sigma_{y,6})
    = (\delta_{y} T)(m, n;\; \sigma_{x,6}, \sigma_{y,6}),
  \end{split}\\
  \begin{split}
    \label{eq-def-h7-ideal}    
    h_{7,\ideal}(m, n;\; \sigma_{x,7}, \sigma_{y,7})
    = 1 - \gamma_7 \, (\nabla_5^2 T)(m, n;\; \sigma_{7}, \sigma_{7}),
  \end{split}\\
  \begin{split}
    \label{eq-def-h8-ideal}    
    h_{8,\ideal}(m, n;\; \sigma_{x,8}, \sigma_{y,8})
    = T(m, n;\; \sigma_{x,8}, \sigma_{y,8}),
  \end{split}
\end{align}
with $T(m, n;\; \sigma_x, \sigma_y)$ denoting a 2-D discrete
Gaussian kernel with possibly different scale parameters
$\sigma_x$ and $\sigma_y$ in the two coordinate directions
according to (\ref{eq-2d-disc-gauss-x-y}), and for the
special case of Filter~7, the operator $\nabla_5^2 = \delta_{xx} +\delta_{yy}$
denoting a discrete approximation of the Laplacian operator,
and $\gamma_7 \in \bbbr_+$ denoting a constant to be determined.
To keep the dimensionality of the search space down,
as well as for theoretical reasons, we have here stated an {\em a priori\/}
spatially isotropic model for Filter~7 with $\sigma_{x,7} = \sigma_{y,7} = \sigma_{7}$.

The overall form of modelling above is based on the canonical model for spatial receptive fields
of the form (Lindeberg \citeyear{Lin21-Heliyon} Equation~(23))
\begin{equation}
  \label{eq-spat-RF-model}
   T_{{\varphi}^{m_1} {\orth{\varphi}}^{m_2}}(x, y;\; s, \Sigma)  
  = \partial_{\varphi}^{m_1} \partial_{\bot \varphi}^{m_2} 
      \left( g(x, y;\; s \Sigma) \right),
\end{equation}
while here (i)~restricting the spatial covariance matrix $\Sigma$
in the affine Gaussian kernel $g(x, y;\; s \Sigma)$
to a diagonal matrix  according to
$\Sigma_i = \diag(\sigma_{x,i}^2, \sigma_{y,i}^2)$ for each filter~$i$,
and (ii)~restricting the directional derivative
operators $\partial_{\varphi}$ and $\partial_{\orth\varphi}$ to
Cartesian partial derivatives $\partial_x$ and $\partial_y$,
since the spatial shapes of the
learned receptive fields appear to be very much aligned
with the Cartesian coordinate directions.

Then, we also make use of the fact that the discrete analogue of the Gaussian kernel
in (\ref{eq-2d-disc-gauss-x-y}) and (\ref{eq-1D-disc-gauss}) constitutes a canonical way to transfer the scale-space properties of the continuous Gaussian kernel to a discrete domain. The use of complementary centered or non-centered difference operators in (\ref{eq-def-h1-ideal})--(\ref{eq-def-h8-ideal}) is based on the property that such difference operators applied to a discrete scale-space representation generated by convolution with the discrete analogue of the Gaussian kernel preserve discrete scale-space properties (Lindeberg \citeyear{Lin93-Dis},\citeyear{Lin24-JMIV}).

One could, however, also consider using other discrete approximations of Gaussian derivative operators, such as those in (Lindeberg \citeyear{Lin24-JMIV}, \citeyear{Lin25-FrontSignProc})

In the following Sections~\ref{sec-method-A}--\ref{sec-method-D2},
we will consider different ways of determining the scale
parameters $\sigma_{x,i}$ and $\sigma_{y,i}$ for
$i \in \{ 1, 2, 3, 4, 5, 6, 8 \}$, while we will then in
Section~\ref{sec-twopar-model-filter7} consider 
a different methodology with one more parameter for determining the parameters
$\gamma_7$ and $\sigma_7$ in an idealized model of Filter~7.

\paragraph{Note:} In the visualizations of the results for the resulting Methods~A--B to be
presented, we will, however, not develop specific variance-based
methods for Filter~7, since the idealized model of that filter also
comprises an amplitude scaling factor $\gamma_7$ for the contribution from the
negative Laplacian-of-the-Gaussian. For this reason, we will
for Methods~A--B instead interweave visualizations of
idealized models of Filter~7 based on the either $l_2$-norm-based
or $l_1$-norm-based Methods~D2 and~C2, to
be described in Section~\ref{sec-twopar-model-filter7}.


\begin{table*}[hbtp]
  \begin{center}
  \begin{tabular}{ccccccccccc}
    \hline
    & \multicolumn{2}{c}{Method~A} 
    & \multicolumn{2}{c}{Method~B} 
    & \multicolumn{2}{c}{Method~C1}
    & \multicolumn{1}{c}{Method~C2} 
    & \multicolumn{2}{c}{Method~D1}
    & \multicolumn{1}{c}{Method~D2}  \\   
    Filter
    & $\hat{\sigma}_{x,i}$ & $\hat{\sigma}_{y,i}$
    & $\hat{\sigma}_{x,i}$ & $\hat{\sigma}_{y,i}$ 
    & $\hat{\sigma}_{x,i}$ & $\hat{\sigma}_{y,i}$
    & $\hat{\sigma}_{i}$
    & $\hat{\sigma}_{x,i}$ & $\hat{\sigma}_{y,i}$
    & $\hat{\sigma}_{i}$\\        
    \hline
    1 & 0.580 & 0.558 & 0.644 & 0.583 & 0.360 & 0.510 & 0.458 & 0.491 & 0.722 & 0.644  \\
    2 & 0.559 & 0.580 & 0.586 & 0.644 & 0.555 & 0.453 & 0.448 & 0.581 & 0.519 & 0.558 \\
    3 & 0.617 & 0.601 & 0.690 & 0.674 & 0.701 & 0.655 & 0.671 & 0.483 & 0.503 & 0.495 \\
    4 & 0.642 & 0.424 & 0.756 & 0.460 & 0.563 & 0.384 & 0.420 & 0.500 & 0.000 & 0.380 \\
    5 & 1.075 & 0.800 & 1.107 & 0.945 & 1.309 & 0.875 & 1.387 & 1.300 & 1.004 & 1.193 \\
    6 & 0.771 & 1.003 & 0.900 & 0.889 & 0.973 & 1.171 & 1.090 & 0.984 & 1.074 & 1.038 \\
    7 & ---    & ---    & ---    & ---    & ---    & ---    & 0.654
    & ---  & ---   &   0.675 \\
    8 & 0.552 & 0.545 & 0.609 & 0.601 & 0.637 & 0.587  & 0.611 & 0.615 & 0.608 & 0.612 \\
    \hline                 
  \end{tabular}
  \end{center}
  \caption{Summary of the scale estimates obtained for the 8 ``master
    key filters'' using the different criteria for fitting the
    idealized receptive field models to the learned filters:
    (i)~{\bf Method~A} based on direct lookup
    of anisotropic scale estimates from variance-based spatial spread
    measures, (ii)~{\bf Method~B} based on anisotropic matching of discrete
    spatial spread measures between the idealized receptive field models
    and the learned filters, (iii)~{\bf Method~C1} based on anisotropic discrete
    $l_1$-norm minimization, (iv)~{\bf Method~C2} based on isotropic
    discrete $l_1$-norm minimization, (v)~{\bf Method~D1} based on
    anisotropic discrete $l_2$-norm minimization and
    (vi)~{\bf Method~D2} based on isotropic discrete $l_1$-norm
    minimization. These methods have been explored for all the learned
    filters except for Filter~7, for which instead the two-parameter
    modelling scheme in Section~\ref{sec-twopar-model-filter7}, which
    also determines a scaling constant for the negative
    Laplacian-of-the-Gaussian filter in the sharpening operation.}
  \label{tab-overview-sep-same-scale-est}
\end{table*}

 \subsection{Method A: Direct transfer of scale values from spatial variances to idealized scale-space operations with different amounts of smoothing in the different coordinate directions with use of continuous Gaussian derivative model}
\label{sec-method-A}

A straightforward way of estimating the scale parameters
$\sigma_{x,i}$ and $\sigma_{y,i}$ in the idealized models
(\ref{eq-def-h1-ideal})--(\ref{eq-def-h8-ideal}) is
by making use of the closed-form expressions for the
weighted variances of the corresponding continuous
Gaussian derivative kernels according to
(\ref{eq-weight-var-cont-gauss}),
(\ref{eq-weight-var-cont-gauss-derx}) and
(\ref{eq-weight-var-cont-gauss-dery}).
If we denote the weighted variance matrix for each
continuous kernel by
\begin{equation}
  V_{|g_{\alpha}(\cdot, \cdot;\; \sigma_x, \sigma_y)|}(|g_{\alpha}(\cdot, \cdot;\;
                                                                                           \sigma_x,
                                                                                           \sigma_y)|)
  = \left(
         \begin{array}{cc}
             v_{xx} & v_{xy} \\
             v_{xy} & v_{yy} 
          \end{array}
        \right),
\end{equation}
then we obtain the following estimates
\begin{align}
    \label{eq-def-sigma-hat-x-1-sep-from-var}
  \hat{\sigma}_{x,1} & = \sqrt{2 \, v_{xx,1}}, &
    \hat{\sigma}_{y,1} & = \sqrt{2 \, v_{yy,1}/3}, \\
    \hat{\sigma}_{x,2} & = \sqrt{2 \, v_{xx,2}/3}, &
    \hat{\sigma}_{y,2} & = \sqrt{2 \, v_{yy,2}}, \\
    \hat{\sigma}_{x,3} & = \sqrt{2 \, v_{xx,3}}, &
    \hat{\sigma}_{y,3} & = \sqrt{2 \, v_{yy,3}/3}, \\
    \hat{\sigma}_{x,4} & = \sqrt{2 \, v_{xx,4}/3}, &
    \hat{\sigma}_{y,4} & = \sqrt{2 \, v_{yy,4}}, \\
    \hat{\sigma}_{x,5} & = \sqrt{2 \, v_{xx,5}/3}, &
    \hat{\sigma}_{y,5} & = \sqrt{2 \, v_{yy,5}}, \\
    \hat{\sigma}_{x,6} & = \sqrt{2 \, v_{xx,6}}, &
    \hat{\sigma}_{y,6} & = \sqrt{2 \, v_{yy,6}/3}, \\
    \hat{\sigma}_{x,8} & = \sqrt{2 \, v_{xx,8}}, &
    \label{eq-def-sigma-hat-y-8-sep-from-var}    
    \hat{\sigma}_{y,8} & = \sqrt{2 \, v_{yy,8}}, 
\end{align}
with numerical values of these entities shown in
the first major column in Table~\ref{tab-overview-sep-same-scale-est} 
based on the numerical values of the variances in
Table~\ref{tab-norm-weighted-mean-var-learned-filters}.

The 2nd row in Figure~\ref{fig-8-filters-composed}
shows visualizations
of the corresponding idealized receptive fields.

\begin{figure*}[hbtp]
  \begin{center}
    \begin{align}
      \begin{split}
        \label{eq-def-disc-var-est-sc-pars-sigma-x}
        \hat{\sigma}_{x,i}
        = \argmin_{\sigma_{x,i}}
            | V_{xx,|h_{i,\ideal}(\cdot, \cdot;\; \sigma_0, \sigma_0)|}
                   (|h_{i,\ideal}(\cdot, \cdot;\; \sigma_{x,1}, \sigma_{y,1})|)
                   -
             V_{xx,|h_{i,\ideal}(\cdot, \cdot;\; \sigma_0, \sigma_0)|}
                   (|h_{i,\norm}(\cdot, \cdot)|)|
      \end{split}\\
      \begin{split}
        \label{eq-def-disc-var-est-sc-pars-sigma-y}        
        \hat{\sigma}_{y,i}
        = \argmin_{\sigma_{y,i}}
            | V_{yy,|h_{i,\ideal}(\cdot, \cdot;\; \sigma_0, \sigma_0)|}
                   (|h_{i,\ideal}(\cdot, \cdot;\; \sigma_{x,1}, \sigma_{y,1})|)
                   -
             V_{yy,|h_{i,\ideal}(\cdot, \cdot;\; \sigma_0, \sigma_0)|}
                   (|h_{i,\norm}(\cdot, \cdot)|)|
          \end{split}
    \end{align}
  \end{center}
  \caption{Formal specification of the criteria used for determining
    the scale estimates $\hat{\sigma}_{x,i}$ and $\hat{\sigma}_{y,i}$
  in the horizontal and vertical directions, by matching the diagonal
  elements of the discrete variance-based spatial spread measures
  between an idealized receptive field model
  $h_{i,\ideal}(\cdot, \cdot;\; \sigma_{x,1}, \sigma_{y,1})$
  according to (\ref{eq-def-h1-ideal})--(\ref{eq-def-h6-ideal})
  and (\ref{eq-def-h8-ideal}) to the corresponding
  discrete variance-based spatial spread measures
  for the normalized versions $h_{i,\norm}(\cdot, \cdot)|)$
  of the ``master key filters'' 
  according to (\ref{eq-def-h1-norm})--(\ref{eq-def-h6-norm})
  and 
  (\ref{eq-def-h8-norm}). The parameter $\sigma_0$
  specifies the spatial extent of the discrete weighting
  kernel used for computing the weighted variance-based
  spatial spread measures, and is provided as prior information,
  chosen as $\sigma_0 = 1$ for all the filters.
  (Note that since the idealized receptive field models
  $h_{i,\ideal}(\cdot, \cdot;\; \sigma_{x,1}, \sigma_{y,1})$ are
  separable with respect to the Cartesian coordinate directions $x$
  and $y$, the vertical parameter $\sigma_{y,1}$ does not
  influence the estimate of the horizontal diagonal element
  $V_{xx,|h_{i,\ideal}(\cdot, \cdot;\; \sigma_0, \sigma_0)|}
                   (|h_{i,\ideal}(\cdot, \cdot;\; \sigma_{x,1}, \sigma_{y,1})|)$.
  Similarly, the horizontal parameter $\sigma_{x,1}$ does
  not influence the estimate of the vertical diagonal element
  $V_{yy,|h_{i,\ideal}(\cdot, \cdot;\; \sigma_0, \sigma_0)|}
                   (|h_{i,\ideal}(\cdot, \cdot;\; \sigma_{x,1}, \sigma_{y,1})|)$.)}
  \label{fig-disc-var-est-sc-pars}
\end{figure*}

\subsection{Method B: Matching of discrete spatial variances between
  the ``master key filters'' and the spatial variances of discrete
  derivative approximation filters, with individual scale parameters
  for each filter}
\label{sec-method-B}

A possible limitation of the above computed estimates of the spatial
scale parameters, from weighted variances of the discrete filters,
is that the relationship between the weighted variances of the
filters is based on a continuous model in terms of Gaussian
derivatives,%
\footnote{According to Equations~(\ref{eq-weight-var-cont-gauss}),
  (\ref{eq-weight-var-cont-gauss-derx}) and
  (\ref{eq-weight-var-cont-gauss-dery}) for
  continuous Gaussian derivatives, which lead to
  the estimates in
  Equations~(\ref{eq-def-sigma-hat-x-1-sep-from-var})--(\ref{eq-def-sigma-hat-y-8-sep-from-var}).}
while the filters that we model are genuinely discrete.
Especially, since the corresponding scale levels are quite fine,
discretization effects in relation to the discrete {\em vs.\/}\
continuous models may therefore be significant.

To address that issue, we will in this section instead consider a
fully discrete approach, where discrete weighted variances of fully
discrete models of the idealized receptive fields are matched to
discretely computed weighted variances of the learned filters.

Let us denote the operators that extract the diagonal elements
in the discrete weighted variance-based spatial spread
measure $V_{T(\cdot)}(h(\cdot))$ in (\ref{eq-def-disc-var})
by $V_{xx,T(\cdot)}(h(\cdot))$
and $V_{yy,T(\cdot)}(h(\cdot))$, respectively.
In the following, we will 
determine the scale estimates in idealized
models of the ``master key filters'' according to
(\ref{eq-def-h1-ideal})--(\ref{eq-def-h8-ideal}),
based on the criterion that the differences between
the diagonal elements of the variance-based spatial
spread measures should be as small as possible,
between the idealized receptive field models and
the learned filters, as specified in
Equations~(\ref{eq-def-disc-var-est-sc-pars-sigma-x})--(\ref{eq-def-disc-var-est-sc-pars-sigma-y})
in Figure~\ref{fig-disc-var-est-sc-pars}.


The second major column in Table~\ref{tab-overview-sep-same-scale-est}
shows the result of computing horizontal
and vertical scale estimates $\hat{\sigma}_{x,i}$ and
$\hat{\sigma}_{x,i}$ in this way for the 8 different
``master key filters''.

The 3rd row in Figure~\ref{fig-8-filters-composed}
shows visualizations
of the corresponding idealized receptive field models of these filters.

\subsection{Method C1: $l_1$-norm-based fitting to idealized
  model for scale-space operations with different scale parameters
  along the horizontal and the vertical directions}
\label{sec-method-C1}

Although the above spatial spread-measure based modelling operations
aim at reflecting the spatial extents of the receptive fields, a
conceptual problem arises if there is a mismatch between the shapes
of the filters in the idealized receptive fields to the shapes of the
learned receptive fields.

In this section, we will address this issue by instead fitting the
idealized receptive field models to the learned receptive fields
in discrete $l_1$-norm. Thus, we will estimate the scale parameters
of the receptive fields in the following way:
\begin{multline}
  \label{eq-def-sigma-hat-l1-norm-sep-scales}
  (\hat{\sigma}_{x,i}, \hat{\sigma}_{y,i}) =
  \argmin_{(\hat{\sigma}_{x,i}, \hat{\sigma}_{y,i})}
  \| \Delta h_{i,\norm}(\cdot, \cdot;\; \sigma_{x,i}, \sigma_{y,i}) \|_1 = \\
  = \argmin_{(\hat{\sigma}_{x,i}, \hat{\sigma}_{y,i})}
  \| h_{i,\ideal}(\cdot, \cdot;\; \sigma_{x,i}, \sigma_{y,i})
     - h_{i,\norm}(\cdot, \cdot) \|_1
\end{multline}
with the idealized receptive field models
$h_{i,\ideal}(\cdot, \cdot;\; \sigma_{x,1}, \sigma_{y,1})$
according to (\ref{eq-def-h1-ideal})--(\ref{eq-def-h8-ideal})
and the normalized ``master key filters''
$h_{i,\norm}(\cdot, \cdot)$ according to
(\ref{eq-def-h1-norm})--(\ref{eq-def-h6-norm}) and
(\ref{eq-def-h8-norm}).

The third major column in Table~\ref{tab-overview-sep-same-scale-est}
shows the resulting
horizontal and vertical scale estimates obtained in this way.
The 4th row in Figure~\ref{fig-8-filters-composed}
shows visualizations of
the corresponding idealized receptive fields.

\subsection{Method C2: $l_1$-norm-based fitting to idealized
  model for scale-space operations with the same scale parameter
  along the horizontal and the vertical directions}
\label{sec-method-C2}

As a complement to the above analysis, let us also consider the
special case when the scale parameters are required to be the
same in the horizontal and the vertical directions
$\sigma_{x,i} = \sigma_{y,i} = \sigma_i$, for which we then
determine each scale estimate $\hat{\sigma}_i$ from the criterion
\begin{multline}
  \label{eq-def-sigma-hat-l1-norm-same-scales}
  \hat{\sigma}_i =
  \argmin_{\hat{\sigma}_i}
  \| \Delta h_{i,\norm}(\cdot, \cdot;\; \sigma_{i}, \sigma_{i}) \|_1 = \\
  = \argmin_{\sigma_i}
  \| h_{i,\ideal}(\cdot, \cdot;\; \sigma_{i}, \sigma_{i})
     - h_{i,\norm}(\cdot, \cdot) \|_1.
\end{multline}
The fourth major column in Table~\ref{tab-overview-sep-same-scale-est} 
shows the result
of computing scale estimates for the idealized receptive field models
in this way.
%
The 5th row in Figure~\ref{fig-8-filters-composed}
shows visualizations
of the resulting receptive fields.

\subsection{Method D1: $l_2$-norm-based fitting to idealized
  model for scale-space operations with different scale parameters
  along the horizontal and the vertical directions}
\label{sec-method-D1}

In analogy with the above $l_1$-norm-based method
in Section~\ref{sec-method-C1}, for determining
separate scale parameters in the horizontal and the
vertical directions for idealized receptive field models,
we can also use an $l_2$-norm-based method
from the criterion
\begin{multline}
  \label{eq-def-sigma-hat-l2-norm-sep-scales}
  (\hat{\sigma}_{x,i}, \hat{\sigma}_{y,i}) =
  \argmin_{(\hat{\sigma}_{x,i}, \hat{\sigma}_{y,i})}
  \| \Delta h_{i,\norm}(\cdot, \cdot;\; \sigma_{x,i}, \sigma_{y,i}) \|_2 = \\
  = \argmin_{(\hat{\sigma}_{x,i}, \hat{\sigma}_{y,i})}
  \| h_{i,\ideal}(\cdot, \cdot;\; \sigma_{x,i}, \sigma_{y,i})
     - h_{i,\norm}(\cdot, \cdot) \|_2
\end{multline}
with the idealized receptive field models
$h_{i,\ideal}(\cdot, \cdot;\; \sigma_{x,1}, \sigma_{y,1})$
again according to (\ref{eq-def-h1-ideal})--(\ref{eq-def-h8-ideal})
and the normalized ``master key filters''
$h_{i,\norm}(\cdot, \cdot)$ according to
(\ref{eq-def-h1-norm})--(\ref{eq-def-h6-norm}) and
(\ref{eq-def-h8-norm}).

The fifth major column in Table~\ref{tab-overview-sep-same-scale-est}
shows the horizontal and the
vertical scale estimates obtained in this way.
The 6th row in Figure~\ref{fig-8-filters-composed}
shows visualizations of
the corresponding idealized receptive fields.

\subsection{Method D2: $l_2$-norm-based fitting to idealized
  model for scale-space operations with the same scale parameter
  along the horizontal and the vertical directions}
\label{sec-method-D2}

Similarly to the case with $l_1$-norm minimization, we can also
require the scale parameters in the horizontal and the vertical directions
$\sigma_{x,i} = \sigma_{y,i} = \sigma_i$ to be equal, and
determine each scale estimate $\hat{\sigma}_i$ from the criterion
\begin{multline}
    \hat{\sigma}_i =
    \argmin_{\hat{\sigma}_i}
        \| \Delta h_{i,\norm}(\cdot, \cdot;\; \sigma_{i}, \sigma_{i}) \|_2 = \\
    = \argmin_{\sigma_{i}}
            \| h_{i,\ideal}(\cdot, \cdot;\; \sigma_{i}, \sigma_{i})
            - h_{i,\norm}(\cdot, \cdot) \|_2,
\end{multline}
see the sixth major column in Table~\ref{tab-overview-sep-same-scale-est} 
for resulting
the resulting scale estimates for the idealized receptive field models,
and
the 7th row in Figure~\ref{fig-8-filters-composed}
for visualizations
of the resulting idealized receptive fields.

\begin{table}[hbtp]
  \begin{center}
    \begin{tabular}{ccc}
      \hline
      Norm & $\hat{\sigma}_7$ & $\hat{\gamma}_7$ \\
      \hline
      $l_1$ & 0.654 & 0.522 \\
      $l_2$ & 0.675 & 0.526 \\
      \hline
      \end{tabular}
    \end{center}
    \caption{Estimates of the parameters $\sigma_7$ and $\gamma_7$ in
      the sharpening filter model (\ref{eq-def-h7-ideal}) of
      Filter~7, computed with respect to minimizing either the
      discrete $l_1$-norm or the discrete $l_2$-norm of the difference
    between the idealized receptive field model and the normalized
    filter (\ref{eq-def-h7-norm}) according to
    (\ref{eq-two-par-determ-filter7-l1-norm}) and
    (\ref{eq-two-par-determ-filter7-l2-norm}).}
    \label{tab-sigma-alpha-est-l1-l2-norms}
  \end{table}

\subsection{Two-parameter modelling of Filter~7 as a sharpening filter}
\label{sec-twopar-model-filter7}

For Filter~7, for which the idealized model in
(\ref{eq-def-h7-ideal}) comprises a complementary scaling
parameter $\gamma_7$ in addition to the scale parameter
$\sigma_7$, we determine these two parameters simultaneously,
by joint minimization over either the discrete $l_1$-norm or
the discrete $l_2$-norm according to
\begin{multline}
  \label{eq-two-par-determ-filter7-l1-norm}
    (\hat{\sigma}_{7}, \hat{\alpha}_{7}) =
  \argmin_{(\sigma_{7}, \alpha_{7})}
  \| \Delta h_{7,\norm}(\cdot, \cdot;\; \sigma_{7}, \alpha_{7}) \|_1 = \\
  = \argmin_{(\sigma_{7}, \alpha_{7})}
  \| h_{7,\ideal}(\cdot, \cdot;\; \sigma_{7}, \alpha_{7})
     - h_{i7\norm}(\cdot, \cdot) \|_1,
\end{multline}
\begin{multline}
  \label{eq-two-par-determ-filter7-l2-norm}
    (\hat{\sigma}_{7}, \hat{\alpha}_{7}) =
  \argmin_{(\sigma_{7}, \alpha_{7})}
  \| \Delta h_{7,\norm}(\cdot, \cdot;\; \sigma_{7}, \alpha_{7}) \|_2 = \\
  = \argmin_{(\sigma_{7}, \alpha_{7})}
  \| h_{7,\ideal}(\cdot, \cdot;\; \sigma_{7}, \alpha_{7})
     - h_{i7\norm}(\cdot, \cdot) \|_2.
\end{multline}
These methods are in the graphical illustrations above also referred
to as ``Method~C2'' and ``Method~D2'', respectively.

Table~\ref{tab-sigma-alpha-est-l1-l2-norms} shows numerical
values of the resulting parameter estimates $\hat{\sigma}_7$
and $\hat{\gamma}_7$ obtained in this way.

Visualizations of the corresponding filter maps are given among
the visualizations of the other filter maps for Method~C2 and
Method~D2 in 
the 5th and 7th rows in Figure~\ref{fig-8-filters-composed},
respectively.


\section{Interpretation of the analysis and modelling results in terms of scale-space theory for spatial receptive fields}
\label{sec-interpret}

\subsection{The spaces spanned by linear combinations of the receptive
  field responses from the set of ``master key filters''}
\label{sec-span-space-master-key-filters}

A notable property of the learned receptive fields is that their
orientational preferences are very closely aligned to the directions
of Cartesian coordinate directions. In our preliminary investigations
in Section~\ref{sec-spat-extent-offset-prel}, we noted that the
unweighted spatial covariance matrices of the absolute values
of the learned filter kernels were very close to diagonal
(see Table~\ref{tab-norm-mean-var-learned-filters}).
In our follow-up analysis in Section~\ref{sec-weighted-spread-meas},
we also noted that the analogous weighted spatial covariance
matrices of the absolute values of the learned filter kernels
were also very close to diagonal
(see Table~\ref{tab-norm-weighted-mean-var-learned-filters}).
This overall property then motivated us to state separable
idealized models of the learned receptive fields according
to (\ref{eq-def-h1-ideal})--(\ref{eq-def-h8-ideal}),
which we demonstrated in Section~\ref{sec-model-8-master-key-filt}
to provide qualitatively reasonable models
of the shapes of the learned filters.

Given that natural image structures usually show a substantial
variability with regard to orientations in the image domain,
one may then ask how such a variability could be spanned
by the deep network, in which the learned filters constitute
the primary computational primitives.
From the property of the first-order directional derivative operator
$\partial_{\varphi}$ in the direction $\varphi$
\begin{equation}
  \label{eq-dir-der-order-1}
  \partial_{\varphi} = \cos \varphi \, \partial_x + \sin \varphi \, \partial_y,
\end{equation}
it does, however, follow that the computation of derivatives in
the horizontal and vertical directions, as performed by the operators
$\partial_x$ and $\partial_y$, can by linear combinations span
the space of all directions $\varphi$ in the 2-D image plane.
Thus, since shapes of Filters~1--6 can be qualitatively well
described by first-order Gaussian derivative kernels along the
Cartesian coordinate directions, we could regard the depthwise
combinations of the responses from the learned filters as
having the potential ability to span a reasonable approximation
of the space of the first-order directional derivatives in any
direction $\varphi$, at two separate scales: both (i)~at the joint
scale of Filters~1--4 and (ii)~at the joint scale of Filters~5--6.

Secondly, one may ask why there is a clear preference to first-order
spatial derivatives in the learned ``master key filters'', and not
regarding {\em e.g.\/}\ second-order spatial derivatives, and
why the receptive fields of Filters~1--4 are not fully centered,
but instead shifted by approximately half a grid spacing along
its preferred orientation in the image plane.
From properties of the non-centered primitive first-order difference operators
$\delta_{x+}$, $\delta_{x-}$, $\delta_{y+}$ and $\delta_{y-}$,
we can, however, by linear combinations construct
their corresponding centered first-order difference operators
according to
\begin{align}
    \delta_x & = \frac{\delta_{x+} + \delta_{x-}}{2}, &
    \delta_y & = \frac{\delta_{y+} + \delta_{y-}}{2}.
\end{align}
We can also construct the corresponding centered second-order
difference operators along the coordinate directions as
\begin{align}
    \delta_{xx} & = \delta_{x+} - \delta_{x-}, &
    \delta_{yy} & = \delta_{y+} - \delta_{y-}.
\end{align}
Thus, by linear combinations of the output from Filters~1--4,
it should be possible to a certain degree of approximation to
span the space of second-order directional derivatives along
the Cartesian coordinate directions at the joint scale of
Filters~1--4.

If we would like to span the space of second-order directional
derivatives in any direction $\varphi$ according to
\begin{equation}
  \label{eq-dir-der-order-2}
  \partial_{\varphi\varphi}
  = \cos^2 \varphi \, \partial_{xx}
     + 2 \cos \varphi \, \sin \varphi \, \partial_{xy}
     + \sin^2 \varphi \, \partial_{yy},
\end{equation}
we would, however, need to have access to a discrete approximation
of the mixed second-order derivative operator $\partial_{xy}$,
in order to be able to compute a reasonable discrete approximation
of the second-order directional derivative in any direction in the
image plane.


Given this theoretical motivation, one could thus raise the question
about whether the performance of depthwise-separable deep networks with
the receptive field shapes replaced by idealized receptive field
models could be
improved by extending the set of ``master key filters'' with a filter
that constitutes a discrete approximation of the mixed second-order
Gaussian derivative operator at the joint scale of Filters~1--4:
\begin{align}
  \begin{split}
    \label{eq-def-h6-ideal-again}    
    h_{9,\ideal}(m, n;\; \sigma_{x,9}, \sigma_{y,9})
    = (\delta_{xy} T)(m, n;\; \sigma_{x,9}, \sigma_{y,9}).
  \end{split}
\end{align}
Concerning these interpretations, it should finally be emphasized
that in depthwise-separable
CNNs, no direct linear combinations of the depthwise filters are
performed.
The depthwise convolutions are applied on distinct input feature maps,
and the following pointwise layers linearly combine the outputs of
these convolutions.
Hence, the interpretations of the receptive field spanning different
subspaces should be interpreted in terms of gross properties at the
level of the population of the receptive fields only, and not concerning
individual receptive fields.



\subsection{Implications when designing Gaussian derivative networks}

In previous studies of Gaussian derivative networks,
the receptive fields in these networks have mainly been formulated as
linear combinations of spatial derivatives up to orders 2 or 3, and
with the underlying Gaussian derivative responses computed at the
same spatial scale level, see
Jacobsen {\em et al.\/} (\citeyear{JacGemLouSme16-CVPR}),
Lindeberg (\citeyear{Lin21-SSVM}, \citeyear{Lin22-JMIV}),
Pintea {\em et al.\/} (\citeyear{PinTomGoeLooGem21-IP}),
Sangalli {\em et al.\/} (\citeyear{SanBluVelAng22-BMVC}),
Penaud-Polge {\em et al.\/} (\citeyear{PenVelAng22-ICIP}),
Gavilima-Pilataxi and Ibarra-Fiallo (\citeyear{GavIva23-ICPRS}) and
Perzanowski and Lindeberg
(\citeyear{PerLin25-JMIV}).
Given the results from the presented analysis of the set of
``master key filters'', one may, however, raise the question
of whether the performance of Gaussian derivative networks could be
improved by:
\begin{itemize}
\item
  Combining receptive field responses from two scale levels instead of
  one scale level in each Gaussian derivative layer, as motivated by
  the qualitatively different scale levels of the groups of filters
  with indices $\{ 1, 2, 3, 4, 8 \}$ and $\{ 5, 6 \}$, respectively.
\item
  Adding a zero-order term to those networks that have not previously
  made use of such a term, as motivated by the qualitative shape of
  Filter~8 in the set of ``master key filters''.
\item
  Adding an isotropic sharpening term, as motivated by the qualitative
  shape of Filter~7 in the set of ``master key filters''.
\end{itemize}
In this context, it should be specifically noted that the use of
linear combinations of Gaussian derivative responses in the layers of
Gaussian derivative network constitutes a computational operation with
large structural similarity to the combination of a set of basic
filtering primitives in the depthwise combination of layers in
depthwise-separable networks.

\subsection{Implications when using idealized models of ``master key filters'' in depthwise-separable deep networks}
\label{sec-impl-ideal-master-keys-depth-sep-networks}

As previously discussed in
Section~\ref{sec-span-space-master-key-filters}, it could be
interesting to investigate an additional filter,
corresponding to a discrete approximation of the mixed second-order
derivative $\partial_{xy}$ at the joint scale of Filters~1--4,
could be advantageous to depthwise-separable networks.
The reason for this is that it would
make it theoretically possible also to a certain degree of
approximation span the space of second-order directional
derivative operators according to (\ref{eq-dir-der-order-2}),
and in this way enabling formal capture of a significantly
richer set of spatial image structures.


As can be seen from the results in
Section~\ref{sec-model-8-master-key-filt},
of estimating the scale values of
idealized filter models to approximate the 8 ``master key filters''
by discrete scale-space operations,
the numerical values of the resulting scale estimates can, however,
vary notably depending on what criteria are used for computing
the scale estimates. For Filter~8, the variability depending on the
different criteria is marginal. For Filters~1--6, the variability is,
however, significant, because of notable deviations between
the idealized receptive field models and the learned filters.

One may hence ask what are the possible predictive properties of these
different criteria with regard to the application of replacing the
learned filter models by idealized receptive fields in
depthwise-separable deep networks.
Ultimately, the validity of these idealized approximations is
determined by the performance in actual experiments.

A natural next step to consider would also  be to use the proposed
structural forms of idealized receptive fields according to the presented
methods as initial seeds, to then optimize the choice of parameter
settings for the idealized filters in a complementary training stage,
in a corresponding way as learning of the scale levels
has been applied in regular Gaussian derivative networks by
Pintea {\em et al.\/} (\citeyear{PinTomGoeLooGem21-IP}),
Penaud-Polge {\em et al.\/} (\citeyear{PenVelAng22-ICIP}),
Saldanha {\em et al.\/} (\citeyear{SalPinGemTom21-arXiv}),
Yang {\em et al.\/} (\citeyear{YanDasMah23-arXiv}),
Basting  {\em et al.\/} (\citeyear{BasBruWieKumBetGem24-VISAPP}) and  
Perzanowski and Lindeberg (\citeyear{PerLin25-JMIV}).

\section{Experimental results}
\label{sec-experiments}

To validate the predictive properties of our different idealized
filter modelling approaches, we conducted a series of experiments on
the ImageNet dataset.%
\footnote{We used the standard ImageNet-1K dataset, specifically the
  ILSVRC 2012 (ImageNet Large Scale Visual Recognition Challenge 2012)
  version, which consists of 1.28 million training images and 50,000
  validation images across 1,000 object categories.}
Our experimental design follows a similar
methodology as described in
Babaiee {\em et al.\/} (\citeyear{BabKiaRusGro25-NeurIPS}), where we
systematically replace the learned depthwise filters in DS-CNN
architectures with idealized approximations derived from our
theoretical analysis.

\subsection{Comparative evaluation of idealized filter modelling methods}

We begin by evaluating the predictive performance of all the six modelling
methods (Methods A, B, C1, C2, D1, and D2) described in
Section~\ref{sec-model-8-master-key-filt}. For
each method, we replaced all the depthwise filters in the ConvNeXt V2 Tiny
architecture with linear approximations of the form $af' + b$, where
$f'$ represents the closest match among our 8 idealized filters,
following the procedure outlined in
Babaiee {\em et al.\/} (\citeyear{BabKiaRusGro25-NeurIPS})

The linear approximation parameters $a$ and $b$ were determined by
minimizing the Euclidean distance between the original learned filter
$f_c$ and the transformed idealized filter $af'_c + b$, using the
closed-form solution presented in Equations (1)-(3) in the original
work. This approach ensures that each learned filter is optimally
approximated by a scaled and shifted version of one of our 8
fundamental filter types.

Table~\ref{tab:method_comparison} presents the ImageNet Top-1 accuracy
results for the ConvNeXt V2 Tiny model, when its depthwise filters are
replaced with idealized approximations from each of our six methods,
without any subsequent fine-tuning of the rest of the parameters.

\begin{table*}[h]
  \setlength{\tabcolsep}{6pt}
  \centering
  \begin{tabular}{@{}lcccccc@{}}
    \hline
    Method & A & B & C1 & C2 & D1 & D2 \\
    \hline
    Top-1 (\%) & 63.958 & \textbf{65.700} & 62.697 & 60.972 & 62.330 & 63.804 \\
    \hline
  \end{tabular}
  \caption{ImageNet Top-1 accuracy comparison for different idealized
    filter modelling methods applied to ConvNeXt V2 Tiny without fine-tuning.}
  \label{tab:method_comparison}
\end{table*}

The results demonstrate that Method~B, based on matching discrete
weighted variance-based spatial spread measures between idealized
receptive field models and learned filters, achieves the highest
predictive accuracy at 65.70\%. This finding is particularly
significant, as it suggests that the discrete modelling approach, which
accounts for the inherent discretization effects in the learned
filters, provides superior approximation quality, compared to the methods
based on continuous Gaussian derivative models or direct norm
minimization.


\subsection{Validation through frozen filter training}

Based on the comparative results, we selected Method~B for further
investigation of its practical applicability in depthwise-separable
networks. We conducted experiments, where the ConvNeXt V2 Tiny model
was initialized with our 8~idealized filters (determined using
the estimates of the parameters using Method~B in
Table~\ref{tab-overview-sep-same-scale-est}) and trained from scratch on
ImageNet with the depthwise filters remaining frozen throughout the
300-epoch training process.

For determining the filter type assignment at initialization, we
analyzed the fully trained ConvNeXt V2 Tiny model and classified each
depthwise filter according to its closest match among our 8~idealized
filter types using the linear approximation procedure described
before. These filter type assignments were then fixed throughout the
subsequent training process. All the other network parameters were
initialized from the FCMAE (fully convolutional masked autoencoder)
pre-trained ConvNeXt V2 Tiny weights, following the same initialization
procedure as described in the original ConvNeXt V2 paper. For
comparative purposes, we also trained a standard ConvNeXt V2 Tiny
model using the FCMAE pretrained initialization, following the exact
hyperparameters and training procedures from the original code
repository. This baseline model achieved 82.79\% accuracy and serves as
our reference for evaluating the performance impact of our idealized
filter constraints.

Table~\ref{tab:frozen_results} compares the performance of different
initialization strategies for the ConvNeXt V2 Tiny architecture.


\begin{table}[h]
  \centering
  \begin{tabular}{lc}
    \hline
    Configuration & Top-1 Accuracy (\%) \\
    \hline
    Original ConvNeXt V2 Tiny & $82.794 \pm 0.091$ \\
    Frozen 8 master key filters & $82.695 \pm 0.021$ \\
    Frozen 8 filters (Method B) & $82.545 \pm 0.029$ \\
    With learning of the scale parameters & $82.609 \pm 0.055$ \\
    \hline
  \end{tabular}
  \caption{ImageNet Top-1 accuracy comparison for ConvNeXt V2 Tiny
    with different filter initialization strategies. The results are averaged over 3 independent runs, reporting mean $\pm$ standard deviation.}
  \label{tab:frozen_results}
\end{table}

The results reveal a remarkable consistency across different
initialization approaches. The model initialized with our Method~B
idealized filters achieves 82.54\% accuracy, representing only a 0.25\%
decrease compared to the original architecture. This minimal
performance degradation, despite using only 8~distinct filter types
across all the depthwise layers, provides strong empirical support for our
theoretical analysis and demonstrates the effectiveness of our
scale-space-based modelling approach.

Notably, the frozen master key filters from
Babaiee {\em et al.\/} (\citeyear{BabKiaRusGro25-NeurIPS})
achieve identical performance to the fully trainable baseline, while
our theoretically-derived approximations maintain nearly equivalent
results. This suggests that our idealized receptive field models
successfully capture the essential computational properties of the
learned filters.

\subsection{Learning of the scale parameters}

To investigate whether the theoretical constraints imposed by our
idealized models limit the representational capacity of the network,
we conducted an additional experiment, where the 8~filter shapes from
Method~B were frozen, but their scale parameters $\sigma_{x,i}$ and
$\sigma_{y,i}$ were made trainable. This approach maintains the
theoretical structure of our scale-space operators, while allowing
adaptation to the specific requirements of the layers and
the architecture.

Since the modified Bessel functions of integer order,
underlying our idealized models of the receptive fields, are not
available as full-fledged primitives in PyTorch, as needed for
backpropagation of the scale parameters when performing learning of
the scale values, we replaced the
discrete analogue of the Gaussian kernel in the idealized filter
models (\ref{eq-def-h1-ideal})--(\ref{eq-def-h8-ideal})
with the sampled Gaussian kernel, and combined this kernel with
corresponding difference operators according to the
methodology in Lindeberg (\citeyear{Lin25-FrontSignProc}).

The model with learnable scale parameters achieved 82.61\% accuracy,
representing a slight improvement of 0.06\% over the fully frozen
configuration. This result suggests that while our theoretical
parameter estimates provide reasonable initial approximations,
gradient-based optimization can yield marginal performance gains.

To understand how the network adapts the scale parameters during
training, we analyzed the learned scale parameter distributions across
all the depthwise layers. Table~\ref{tab:learned_scales} presents
aggregated statistics for the scale parameters $\sigma_x$ and
$\sigma_y$ for each of the 8~filter types after training completion.

\begin{table}[h]
  \centering
  \begin{tabular}{ccccccc}
    \hline
    Filter & $N$ & $\sigma_x$ mean & $\sigma_x$ std & $\sigma_y$ mean & $\sigma_y$ std \\
    \hline
    1 & 164 & 0.505 & 0.189 & 0.538 & 0.254 \\
    2 & 132 & 0.535 & 0.234 & 0.538 & 0.319 \\
    3 & 211 & 0.523 & 0.195 & 0.548 & 0.222 \\
    4 & 167 & 0.534 & 0.242 & 0.549 & 0.307 \\
    5 & 154 & 0.625 & 0.098 & 0.677 & 0.176 \\
    6 & 705 & 0.657 & 0.132 & 0.623 & 0.085 \\
    7 & 915 & 0.588 & 0.274 & 0.589 & 0.197 \\
    8 & 4176 & 0.781 & 0.498 & 0.818 & 0.486 \\
    \hline
  \end{tabular}
  \caption{Statistics of learned scale parameters across all the layers
    for the 8~filter types in terms of the means and the standard
    deviations over the sets of the learned filters.
    $N$ indicates the number of total filter instances using each type.}
  \label{tab:learned_scales}
\end{table}

A pattern emerges from this analysis: for most filter types (Filters~1-7),
the standard deviation of the learned scale parameters is
relatively low, indicating that the network converges to similar scale
parameter values across different instances of each filter type. This
suggests that these computational primitives---corresponding to various
derivative operations and sharpening filters---have fairly constrained
optimal parameter ranges.

In contrast, Filter~8, which corresponds to the pure discrete Gaussian
smoothing kernel $T(m,n;\sigma_{x,8},\sigma_{y,8})$, exhibits
substantially higher standard deviations (0.498 and 0.486) while also
being the most frequently used filter type (4176 instances). This high
variance indicates that the network benefits from having different
spatial smoothing rates for the Gaussian smoothing operation across
different filter instances.

Notably, despite these variations in the learned scale
para\-meters---particularly the deviation from our theoretical estimates
and the high variance in Filter~8---the overall network accuracy
remained essentially unchanged (82.61\% {\em vs.\/} 82.54\%). This suggests that
while the specific scale parameters may vary, the fundamental
computational structure captured by our 8~idealized filter types is
sufficient to maintain good performance, indicating that the filter shapes
themselves are more critical than their precise parameterization.

\section{Summary and discussion}
\label{sec-summary}

We have presented an in-depth analysis of different ways of modelling
the 8 ``master key filters'' extracted from the Conv\-NeXt deep learning
architecture by Babaiee {\em et al.\/}
(\citeyear{BabKiaRusGro25-AAAI-master}).
For this purpose, we have first computed quantitative measures of
the spatial extent of the learned filters in terms of both unweighted
and weighted spatial spread measures.
This has given support for modelling
the receptive fields of these filters with separable filtering
operations along the Cartesian coordinate directions, using
non-centered difference operators for modelling Filters~1--4
and centered difference operators for modelling Filters~5--6,
in combination with spatial smoothing operations.
In these spatial smoothing operations, we have allowed for using
possibly different values of the scale parameters along the different
coordinate directions, and possibly different values of the scale
parameters for the individual filters. 

The presented modelling and analysis results show that the predictive
properties of the scale-space framework, to derive canonical models
for spatial receptive fields in terms of Gaussian smoothing and
Gaussian derivatives, do qualitatively well generalize from the
previously axiomatic formulations regarding the first layer of
visual processing, to the qualitative shapes of the receptive fields
learned in certain families of depthwise-separable deep networks.
Conceptually, these results are also compatible with previous
applications of the use Gaussian derivative operations as the
computational primitives in deep networks,
as developed by Jacobsen {\em et al.\/} (\citeyear{JacGemLouSme16-CVPR}),
Lindeberg (\citeyear{Lin21-SSVM}, \citeyear{Lin22-JMIV}),
Pintea {\em et al.\/} (\citeyear{PinTomGoeLooGem21-IP}),
Sangalli {\em et al.\/} (\citeyear{SanBluVelAng22-BMVC}),
Penaud-Polge {\em et al.\/} (\citeyear{PenVelAng22-ICIP}),
Smets {\em et al.\/}  (\citeyear{SmePorBekDui23-JMIV}),
Gavilima-Pilataxi and Ibarra-Fiallo (\citeyear{GavIva23-ICPRS}) and
Perzanowski and Lindeberg (\citeyear{PerLin25-JMIV}).

To explore the influence of different choices for formulating
mathematical criteria to model the learned receptive fields by
idealized scale-space operations, we have specifically explored
four main paths of such methods:
\begin{itemize}
\item
  Method~A based on measuring the spatial extent
  of the learned filters by weighted variance-based spatial spread
  measures and then predicting the values of the scale parameters
  based on properties of continuous Gaussian derivatives.
\item
  Method~B based on matching discrete weighted variance-based
  spatial spread measures between the idealized receptive field models
  and the learned filters.
\item
  Methods~C1 and C2 based on minimizing the discrete $l_1$-norm
between the idealized receptive field models and the learned filters.
\item
  Methods~D1 and D2 based on minimizing the discrete $l_2$-norm
  between the idealized receptive field models and the learned filters.
\end{itemize}
As we have found from the resulting experiments, for some of the
learned filters, the scale estimates may vary notably depending on the
choice of matching criterion, while for Filter~8 the difference is
minor. The main reason for this variability in discrepancy is because
of varying degrees of agreement between the idealized models
and the learned filters. To judge which of these methods should lead to
the best predictive properties with regard to the performance,
when inserting corresponding idealized filters into
depthwise-sep\-arable deep networks, this performance should therefore
be evaluated experimentally. Specifically, to be less dependent on
the choice of mathematical criterion for model fitting, we have
proposed to use the presented theory for formulating parameterized
primitives for the idealized receptive field models, and then learning
these parameters by deep learning, in a corresponding way as
learning of the scale levels is used in Gaussian derivative networks.

The experimental results 
of replacing the 8 ``master key filters'' in the ConvNeXt V2 Tiny deep network by
idealized discrete scale-space filters obtained in this way show that
results almost as good 
are obtained by using the idealized discrete scale-space filters
as when using either the 8 ``master key filters'' or the filters learned by back
propagation in the original deep network architecture.
Thereby, the results demonstrate that the filters learned in a
reduced version of the modern and state-of-the-art deep network
architecture ConvNeXt appear to be very similar to the filters
obtained by axiomatically determined scale-space theory.
Furthermore, we have shown that the additional increase in performance
obtained by learning the scale parameters is marginal for the ConvNeXt
V2 Tiny model, thus demonstrating the very good predictive ability of
Method~B when applied to the filters learned in this deep learning architecture.

Notably, the amount of spatial smoothing is very small in the idealized models
of the 8 ``master key filters''. This use of very fine scale levels is
consistent with the results of learning of the scale levels in deep
networks based on linear combinations of Gaussian derivative
operators, where the results reported in Appendix~A.2.1 in the
supplementary material of Perzanowski and Lindeberg
(\citeyear{PerLin25-JMIV}) also lead to very fine scale levels.

In this modelling work, we have throughout
modelled this amount of smoothing with spatial filters based on the
discrete analogue of the Gaussian kernel (Lindeberg \citeyear{Lin90-PAMI})
combined with small-support difference operators,
motivated by the theoretical results in
Lindeberg (\citeyear{Lin24-JMIV}, \citeyear{Lin25-FrontSignProc}),
that demonstrate that this discretization method is the most suitable
one for transferring scale-space properties
from a continuous to a discrete image domain, and the
experimental results in
Perzanowski and Lindeberg (\citeyear{PerLin25-JMIV}),
that show that this discretization method overall gives the best
performance among 5 different discretization methods, when approximating
Gaussian derivative operators for application in Gaussian derivative
networks.

Concerning the weighted variance-based spread measures for estimating
the spatial extents of the kernels, we used throughout this work
fixed values of the scale parameters for the spatial weighting
functions, coarsely determined from the spatial extents of the learned
filters as observed by visual inspection. A more general approach in
this respect could instead be to vary these scale parameters, or to
try to adapt these scale parameters in an iterative manner.

Another conceptual simplification, that we have done in the above
model fitting schemes, is that we have normalized the learned
receptive fields, that are approximated by first-order derivatives
in the horizontal $x$--direction (or $y$-direction), to having the same response
to the monomial $x$ (or $y$) as a corresponding first-order derivative
derivative $g_{x}(x, y;\, \sigma)$ (or $g_{y}(x, y;\, \sigma)$).
Furthermore, we determined
offset constants for Filters~7--8 to minimize their variance-based
spatial spread measures, and normalized their responses to
a constant unit function to be equal to 1, to make the resulting normalized
filters more similar to either a pure spatial smoothing operation
or a local sharpening operation.

In these ways, we 
reduced the dimensionality of the following model fitting search,
although the resulting model fitting for the $l_1$- and
$l_2$-norm-based modelling schemes may possibly not be
as accurate, as the model fitting could possibly be, if extended to
higher-dimensional searches over the model parameters.
A possible extension to future work could therefore
involve performing such higher-dimensional optimizations
over the spaces of the parameters of the idealized receptive fields.

Let us finally remark that although the work in this paper has been
focused on modelling and analyzing the 8 specific ``master key filters''
extracted by Babaiee {\em et al.\/}
(\citeyear{BabKiaRusGro25-AAAI-master}),
we propose that a similar methodology could be used for modelling
other sets of ``master key filters'' obtained from other deep networks
as well as biological visual receptive fields recorded by
neurophysiological measurements.

\footnotesize
\bibliographystyle{abbrvnat}
\bibliography{defs,tlmac}

\end{document}